\documentclass{article}

\usepackage{microtype}
\usepackage{graphicx}
\usepackage{subfigure}
\usepackage{booktabs} 

\usepackage{hyperref}

\usepackage[accepted]{icml2024}

\usepackage{amsmath}
\usepackage{amssymb}
\usepackage{mathtools}
\usepackage{amsthm}
\usepackage{multirow}
\definecolor{NVgreen}{RGB}{118,185,0}
\definecolor{NVblack}{RGB}{0,0,0}
\definecolor{NVlgrey}{RGB}{205,205,205}
\definecolor{NVmgrey}{RGB}{140,140,140}
\definecolor{NVdgrey}{RGB}{94,94,94}

\definecolor{NVemerald}{RGB}{0,133,100}
\definecolor{NVamethyst}{RGB}{93,22,130}
\definecolor{NVintel}{RGB}{0,113,197}
\definecolor{NVgarnet}{RGB}{137,12,88}
\definecolor{NVfluorite}{RGB}{250,194,0}
\usepackage[capitalize,noabbrev]{cleveref}
\graphicspath{{./figs/}}
\theoremstyle{plain}
\newtheorem{theorem}{Theorem}[section]

\theoremstyle{definition}
\newtheorem{definition}[theorem]{Definition}

\theoremstyle{remark}

\usepackage{algorithm}
\usepackage[]{algpseudocode}
\renewcommand{\vec}[1]{\boldsymbol{#1}}
\newcommand{\revise}[1]{\textcolor{black}{#1}}
\usepackage[textsize=tiny]{todonotes}

\newcommand\round[1]{\left[#1\right]}
\icmltitlerunning{ILILT: Implicit Learning of Inverse Lithography Technologies}

\begin{document}

\twocolumn[
\icmltitle{ILILT: \underline{I}mplicit \underline{L}earning of \underline{I}nverse \underline{L}ithography \underline{T}echnologies}




\begin{icmlauthorlist}
\icmlauthor{Haoyu Yang}{1}
\icmlauthor{Haoxing Ren}{1}
\end{icmlauthorlist}

\icmlaffiliation{1}{Design Automation Research Group, NVIDIA, Austin, TX}
\icmlcorrespondingauthor{Haoyu Yang}{haoyuy@nvidia.com}

\icmlkeywords{Machine Learning, ICML}

\vskip 0.3in
]



\printAffiliationsAndNotice{} 

\begin{abstract}

Lithography, transferring chip design masks to the silicon wafer, is the most important phase in modern semiconductor manufacturing flow.
Due to the limitations of lithography systems, 
Extensive design optimizations are required to tackle the design and silicon mismatch.
Inverse lithography technology (ILT) is one of the promising solutions to perform pre-fabrication optimization, termed mask optimization.
Because of mask optimization problems' constrained non-convexity, numerical ILT solvers rely heavily on good initialization to avoid getting stuck on sub-optimal solutions.
Machine learning (ML) techniques are hence proposed to generate mask initialization for ILT solvers with one-shot inference, targeting faster and better convergence during ILT. This paper addresses the question of \textit{whether ML models can directly generate high-quality optimized masks without engaging ILT solvers in the loop}. We propose an implicit learning ILT framework: ILILT, which leverages the implicit layer learning method and lithography-conditioned inputs to ground the model. Trained to understand the ILT optimization procedure, ILILT can outperform the state-of-the-art machine learning solutions, significantly improving efficiency and quality.

\end{abstract}

\section{Introduction}
\label{sec:intro}
\begin{figure}
	\centering
	\includegraphics[width=.4\textwidth]{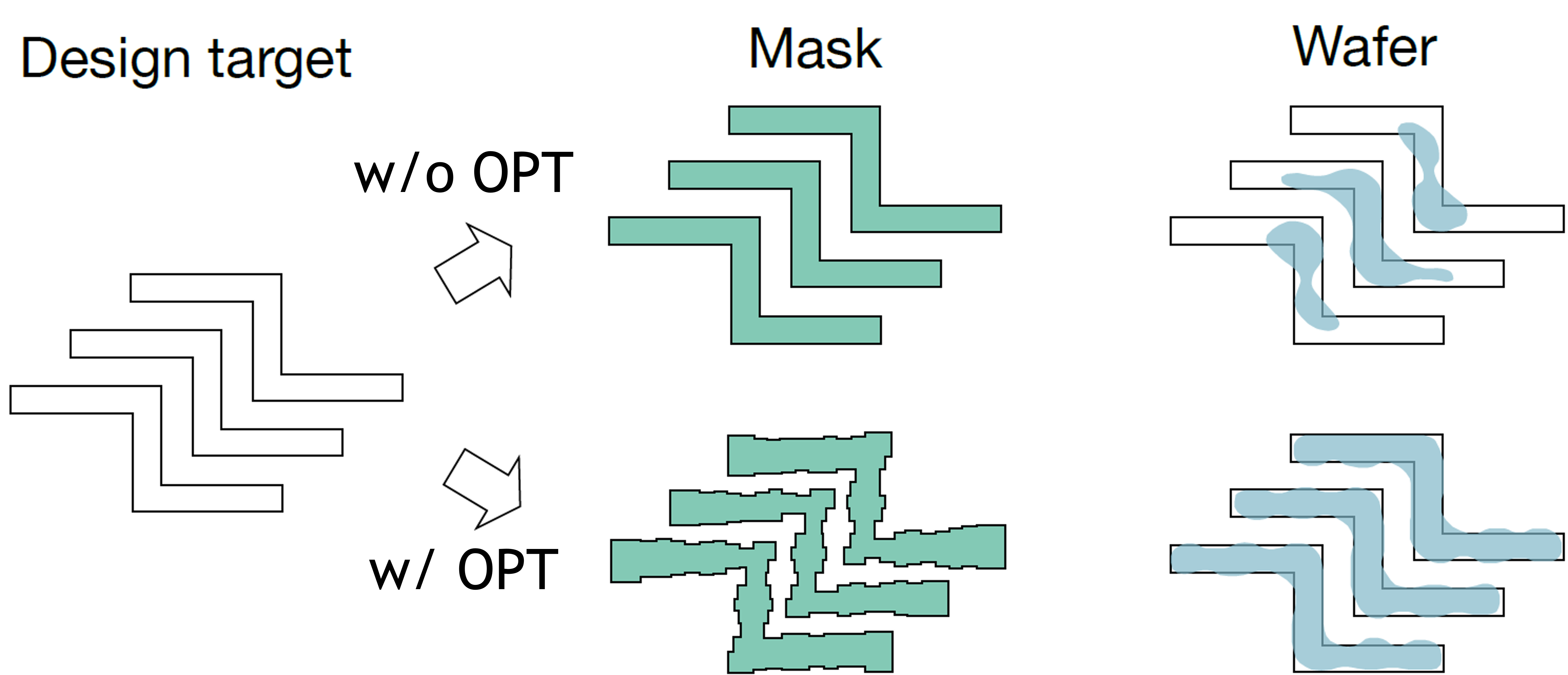}
	\caption{Limitations of lithography system requires design optimization to be correctly fabricated on the silicon wafer.}
	\label{fig:litho-prox}
\end{figure}

Lithography is the most important phase in semiconductor manufacturing flow, which transfers chip design patterns onto the silicon wafer.
However, due to the limitations of lithography systems, there are usually fetal manufacturing gaps between the design and the patterns on the wafer. 
Extensive prior fabrication design optimizations, termed \textit{mask optimization}, are required to compensate for the manufacturing loss (\Cref{fig:litho-prox}). 

\begin{figure}
    \centering
    \subfigure[ILT]{\includegraphics[height=.25\textwidth]{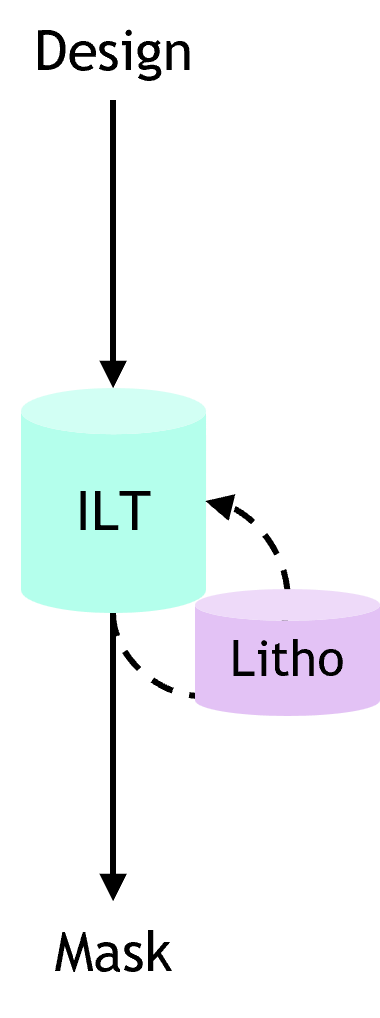} \label{fig:iltscheme-1}}
    \hspace{1cm}
    \subfigure[ML-Aided]{\includegraphics[height=.25\textwidth]{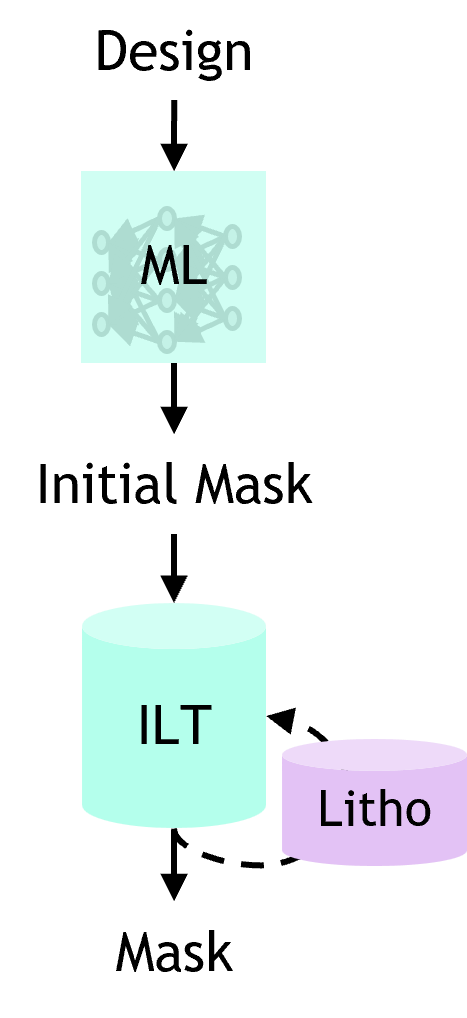} \label{fig:iltscheme-2}}
    \hspace{1cm}
    \subfigure[ILILT]{\includegraphics[height=.25\textwidth]{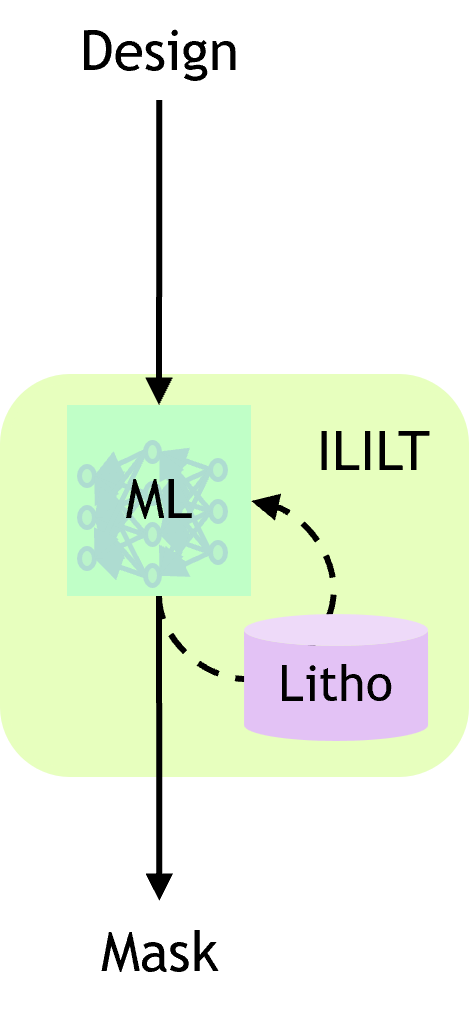} \label{fig:iltscheme-3}}
    \caption{Working scheme of ILT solvers. (a) Standard ILT solver that iteratively updates the mask from design till convergence. (b) Common ML approach that uses ML to produce an initial mask followed by standard ILT procedure. (c) The ILILT that uses ML to imitate the standard ILT procedure. }
    \label{fig:iltscheme}
\end{figure}

Inverse lithography technology (ILT), as a numerical solution, has shown great promise for mask optimization on advanced-node lithography \cite{OPC-DAC2014-Gao,OPC-DATE2021-Yu,OPC-SPIE2019-Kyle,OPC-SPIE2019-Pang}. 
As shown in \Cref{fig:iltscheme-1}, an ILT solver requires frequent calls of lithography simulator, which mathematically approximate the lithography fabrication behavior.
The forward path computes the wafer image corresponding to the current mask and we measure the error between the wafer image and the original design.
In the backward path, the error will be passed back through gradient method to update the mask.
However, the non-convexity of mask optimization problems poses a great challenge to numerical ILT solvers, that 
they rely heavily on good mask initialization and can easily get stuck at sub-optimal solutions.

Recent research into machine learning (ML) techniques for lithography have made ILT solvers more efficient.
As shown in \Cref{fig:iltscheme-2}, ML can provide better optimization starting points that significantly reduce ILT iterations and yield better convergence \cite{OPC-DAC2018-Yang,OPC-ICCAD2020-DAMO,OPC-ICCAD2020-NeuralILT}.
GAN-OPC \cite{OPC-DAC2018-Yang} is the earliest deep learning solution for inverse lithography, where a conditional generative adversarial network is proposed for initial mask generation with litho-guided pre-training.
DAMO \cite{OPC-ICCAD2020-DAMO} improves on this work by replacing the standard generator in GAN-OPC with nested UNet and ResNet bottleneck layers \cite{UNet++,pix2pix,Resnet}, yielding a direct output of high-resolution masks.
Jiang et~al.~\cite{OPC-ICCAD2020-NeuralILT} also integrates the explicit gradient calculation of litho-guided pre-training and achieves better training convergence and mask quality.
Very recently, Zhao et~al.~\cite{OPC-ICCAD2022-Zhao} develops the AdaOPC framework that is able to learn from layout geometry and identifies whether a pattern can be optimized by the machine learning engine.
This further increase the applicability of ML for mask optimization solutions.

However, existing solutions only exploit limited ML in the ILT solver and still require significant intervention by traditional numerical solvers.
For example, GAN-OPC and its variations can only learn a design-to-mask mapping, which cannot be used to fix poor-quality masks;
AdaOPC \cite{OPC-ICCAD2022-Zhao} has to conduct full numerical optimizations on complicated patterns. In this paper, we try to address the question of \textbf{whether an ML model can directly generate
high quality optimized masks without engaging ILT solvers in the loop}.  We believe that the key limitation of existing methods is that the trained ML models are not grounded to any actual lithography condition during inference, therefore training-inference distribution shift can not be recognized and corrected. 
We argue that \textbf{ML models grounded on lithography estimation of intermediate masks can make an ML-based ILT framework more effective.}



We propose the ILILT (\Cref{fig:iltscheme-3}) framework to formulate mask optimization as an implicit layer learning problem, trained to optimize the mask towards an equilibrium state \cite{DEQ}. Like a standard ILT solver that iteratively updates masks from a design,
ILILT iteratively modifies masks from a design but with much fewer steps and each step runs significantly faster because no gradient computing is required.
This process produces a number of intermediate masks which ground the ML model with intermediate wafer images computed by lithography models. 

\textbf{ILT As An Implicit Layer}.
Unlike common neural network layers that explicitly define specific computation graphs supporting forward and backward propagation, 
an implicit layer models a joint condition of its input and output \cite{implicit}.
The nature of ILT makes it a representative example of an implicit layer, where the input is a chip design and the output is the corresponding optimized mask that results in a similar wafer pattern compared to the input after the lithography process.
Inspired by the implicit layer learning approach, we design ILILT with a weight-tied sequence model,
making it possible to learn an optimization trajectory from training design-mask pairs. 
To make the trajectory learning grounded with physics, we also integrate ILILT with fast lithography estimator that can implicitly apply lithography conditions on top of the optimization procedure. 
After training, ILILT is able to perform mask optimization on a given design and reaches a stable state in a few iterations.
We will later show how ILILT outperforms a state-of-the-art academic ILT engine with technical details and supporting experiments on public chip design benchmarks.

The reminders of the paper are organized as follows:
\Cref{sec:prelim} covers basic terminologies and related works of mask optimization;
\Cref{sec:alg} discusses and analyzes the technique details of the ILILT framework;
\Cref{sec:result} presents the experimental results supporting the effectiveness of ILILT
and \Cref{sec:conclu} concludes the paper.

\section{Preliminaries}
\label{sec:prelim}


\subsection{ILT Solver}
Mask optimization tries to find a mask design such that the remaining pattern on the silicon wafer after the lithography process is as close as possible to the desired shapes.
This can be formulated as a constrained optimization problem:
\begin{align}
	\min_{\vec{M}} l=d(\vec{Z},\vec{Z}^\ast),  \label{eq:mo}
\end{align}
where,
\begin{align}
	 \vec{Z} = f_l (\vec{M}), 
\end{align}
where $\vec{Z}$ and $\vec{Z}^\ast$ represent the wafer image and the chip design, respectively, $\vec{M}$ corresponds to the mask to be optimized, 
$f_l$ represents the lithography process which is highly non-convex (see appendix for more details)
and $d$ is a measurement indicating the differences between $\vec{Z}$ and $\vec{Z}^\ast$.
Inverse lithography techniques \cite{OPC-DAC2014-Gao} try to solve Problem~(\ref{eq:mo}) in a numerical way by repeating the following steps until convergence:
\begin{enumerate}
	\item Forward: $ \vec{Z}_t=f_l(\vec{M}_t). $
	\item Backward: $ \vec{M}_{t+1}=\vec{M}_{t}-\lambda \dfrac{\partial l}{\partial \vec{Z}_t} \dfrac{\partial \vec{Z}_t}{\partial \vec{M}_t}.$
\end{enumerate}
where $\lambda$ is some hyper-parameter that controls how the mask image is updated during the optimization procedure.
There are also variations of ILTs \cite{OPC-DATE2021-Yu,OPC-ICCAD2020-NeuralILT,OPC-DAC2014-Gao}, but they mostly share similar mask-update concepts.

\subsection{Implicit Layers}
Most neural network layers or computational graphs explicitly define a function $\vec{y}=f(\vec{x})$, which is differentiable and can be back-propagated via the chain rule.
An implicit layer, instead of specifying how to compute the layer’s output from the input, specifies the conditions that we want the layer’s output to satisfy, separating the procedure to compute the layer from the layer definition itself.  It has been used in deep equilibrium models, Neural ODEs, and differentiable optimizations. 

\begin{definition}[Implicit Layer]
Implicit layers try to find $\vec{y}$ such that $f(\vec{x}, \vec{y}) \in \mathcal{C}$,
which defines joint conditions between $\vec{x}$ and $\vec{y}$ \cite{implicit}.
\end{definition}

A fixed-point layer is an example of the implicit layer method, which directly computes the fixed point of a nonlinear transformation, i.e., the solution to the nonlinear system which models the convergent state of a recurrent sequence.
\begin{definition}[Fixed-Point Layer]
	A fixed-point layer can be mathematically given by
	\begin{align}
		\vec{y} = g(\vec{x},\vec{y}).
		\label{eq:fplayer}
	\end{align}
\end{definition}
A fixed-point layer can be considered to be an infinite number of layers parameterized by the same weight $\vec{w}$ (weight-tied), where each layer can be represented as 
\begin{align}
	\vec{y}_{t+1} = g(\vec{x},\vec{y}_t, \vec{w}),
\end{align}

To train a fixed-point layer, \cite{DEQ} proposes an unrolling-free deep equilibrium (DEQ) method, which trains the fixed-point layer in two alternative steps: (1) solve for the fixed-point of the layer and (2) update network parameters. 
This process consists of repeated forward and backward computations as follows.
The forward path is related to the finding of the fixed point of \Cref{eq:fplayer} and can be solved with Newton's method.
Let
\begin{align}
	h(\vec{x}, \vec{y}^\ast,  \vec{w}) = g(\vec{x}, \vec{y}^\ast, \vec{w}) - \vec{y}^\ast,
\end{align}
and $\vec{y}^\ast$ can be found through the following step until convergence:
\begin{align}
	\vec{y}_{t+1} = \vec{y}_{t} - \lambda \vec{J}_t^{-1} h(\vec{x}, \vec{y}_{t},  \vec{w}),
	\label{eq:deqforward}
\end{align}
where $\vec{J}_t$ is the Jacobian matrix of $h$ at $\vec{y}_{t}$ and $\lambda$ is the update rate.

The backward process covers the update of $\vec{w}$, which can be written as 
\begin{align}
	\vec{w} = \vec{w} - \lambda \dfrac{\partial l}{\partial \vec{w}}=\vec{w} + \lambda  \dfrac{\partial l}{\partial \vec{y}^\ast} \vec{J}^{\ast -1} \dfrac{\partial g}{\partial \vec{w}},
	\label{eq:deqback}
\end{align}
where $\vec{J}^\ast$ is the Jacobian matrix of $h$ at $\vec{y}^\ast$, $l$ is a loss function that measures whether $f(x,y) \in \mathcal{C}$ satisfies the implicit layer definition, e.g. the ILT loss \Cref{eq:mo}.
For a detailed proof of \Cref{eq:deqback}, we refer readers to \cite{DEQ}.

\subsection{Neural Networks for ML-based Computational Lithography}
There are two major groups of network architectures used in lithography applications:
(1) The UNet family \cite{UNet,UNet++} and (2) The FNO family \cite{fno}. UNets consist of various bypass connections between convolution layer outputs. Representative works include TEMPO \cite{DFM-ISPD2020-Ye}, LithoGAN \cite{DFM-DAC2019-Ye}, DAMO \cite{OPC-ICCAD2020-DAMO} and GAN-OPC \cite{OPC-DAC2018-Yang}.
 FNOs contain one or more Fourier Neural Operators along with auxiliary convolution layers. Representative works are DOINN \cite{DFM-DAC2022-Yang} and CFNO \cite{CFNO}.
UNet-based models usually rely on large model capacity to learn training data distribution while FNO-based models focus on global information learning through frequency domain embedding.

\section{The ILILT Framework}
\label{sec:alg}

\subsection{ILT Layer}
Following the definition of implicit layer learning, we can solve the ILT equations with
an implicit ILT layer defined as finding $\vec{M}$ such that 
\begin{align}
	f(\vec{M},\vec{Z}^\ast)=d(f_l(\vec{M}),\vec{Z}^\ast)=\epsilon,
	\label{eq:iltlayer}
\end{align}
where $\epsilon$ is some minimum value achievable for \Cref{eq:mo}.

Because \Cref{eq:iltlayer} is a composite of a group of non-linear functions, it is difficult to solve explicitly.
Therefore, we assume \Cref{eq:iltlayer} has a solution $\vec{M}^\ast$ that follows the form of
\begin{align}
	\vec{M}^\ast = g(\vec{M}^\ast, \vec{Z}^\ast),
	\label{eq:fpilt}
\end{align}
which yields a fixed-point layer representation of the ILT solution.

We start solving \Cref{eq:fpilt} from the weight-tied approach, where we define
\begin{align}
	\vec{M}_{t+1} = g(\vec{M}_t, \vec{Z}^\ast, \vec{w}), ~~t=1,2,...,
	\label{eq:wtilt}
\end{align}  
and 
\begin{align}
	\lim\limits_{t\rightarrow \infty} \vec{M}_t = \lim\limits_{t\rightarrow \infty} g(\vec{M}_t, \vec{Z}^\ast, \vec{w}) = g(\vec{M}^\ast, \vec{Z}^\ast, \vec{w}) =\vec{M}^\ast.
\end{align}

\subsection{Lithography Estimation}
As we discussed previously, due to the lack of prior lithography knowledge, state-of-art AI approaches have been limited to only generate initialization for numerical ILT solvers.
This motivates us to build lithography estimation inside $g$ to ground the AI model.

We rewrite \Cref{eq:wtilt} as :
\begin{align}
	\vec{M}_{t+1} = g(\vec{M}_t, \vec{Z}_t, \vec{Z}^\ast, \vec{w}), ~~t=1,2,...,
	\label{eq:ililt}
\end{align}
where $\vec{Z}_t = f_l(\vec{M}_t)$ is the wafer image of $\vec{M}_t$ at time stamp $t$.
\Cref{eq:ililt} formulates the core of the ILILT framework.
A virtue of \Cref{eq:ililt} is the embedded condition of the lithography estimator that implicitly tells $g$ the physical meaning of each step.

\subsection{Training}

\Cref{eq:ililt} defines a weight-tied fixed-point model that can be solved with the DEQ method. Although DEQ \cite{DEQ} reduces the computing overhead of sequence unrolling and back-propagation through time, there are several concerns when it's used for solving ILT problems:
\begin{itemize}
	\item Solving $\vec{M}^\ast$ through Newton's method or other Root-Finding algorithms is still computationally costly considering that $\vec{M}^\ast$ usually has millions of entries.
	\item The solution error caused by the non-optimality in Newton's method will further propagate to the gradient calculation in backward DEQ, inducing additional errors.
	\item The DEQ backward pass still requires the computation of the gradient over lithography modeling in the term $\frac{\partial d}{\partial \vec{M}^\ast}$, showing no advantage over a traditional ILT solver.
\end{itemize} 
To address these concerns, we developed the following training architecture that solves the ILT layer efficiently and effectively.
We unroll ILILT to a fixed depth $T$ and allow the final output $\vec{M}_T$ to be trained towards a ground truth mask $\vec{M}^\ast$,
\begin{align}
	\min_{\vec{w}}~~& l= \sum_{t=\round{\frac{T}{2}}}^{T} \exp[\frac{t}{T} -1] \cdot ||\vec{M}_t-\vec{M}^\ast||_F^2, 	\label{eq:trainililt}\\
	\text{s.t.~~}& \vec{M}_{t+1} = g(\vec{M}_t, \vec{Z}_t, \vec{Z}^\ast, \vec{w}), ~~t=0,1,...,T-1, \nonumber \\
	& \vec{Z}_t = f_l(\vec{M}_t),  ~~t=0,1,...,T, \nonumber
\end{align}
where $t$ indicates each unrolling step.

\revise{
It should be noted that to encourage the mask to grow towards and stop at the golden solution, we did not adopt the cost function that directly minimizes 
the difference between $\vec{M}_T$ and $\vec{M}^\ast$.
Instead, we ask intermediate masks in later unrolling stages to have a trend growing toward the ground truth, which is reflected in \Cref{eq:trainililt} having masks at each time step $\vec{M}_t$ trained towards the ground truth $\vec{M}^\ast$. 
The coefficient $\exp[\frac{t}{T} -1]$ applied at each loss term is to let the training process be aware of the progressive growth of masks at each time step.}
This trick allows the mask to grow smoothly in these steps while stopping at the fixed point.
The exponential term is empirically chosen to regularize the mask-growing trends which are consistent with the behavior of the numerical ILT solver that only minor changes are observed in later optimization iterations.
Solving \Cref{eq:trainililt} is straightforward through back-propagation through time (BPTT). 

\subsection{Discussion}
\subsubsection{Relations to ML-Aided Solutions}
For the training procedure, we observe that ILILT resembles GAN-OPC \cite{OPC-DAC2018-Yang} in terms of the training objectives, 
where we also force the model to learn a golden mask generated by some conventional methods.
However, ILILT advances beyond GAN-OPC works \cite{OPC-DAC2018-Yang,OPC-ICCAD2020-NeuralILT,OPC-ICCAD2022-Zhao,OPC-ICCAD2020-DAMO} because it not only learns what a good mask should look like, but also \textbf{why and how a good mask is grown} thanks to the implicit lithography condition introduced by $f_l (\cdot)$.

\subsubsection{Relations to Implicit Layer Learning}
The whole ILILT method is built upon the definition of the ILT layer and how it is solved. 
Representative solutions, e.g.~DEQ \cite{DEQ}, still requires costly computation of gradients of the objectives over physical models.
ILILT makes it \textbf{even more implicit} by feeding the wafer image, target design and the mask image simultaneously into $g(\cdot, \vec{w})$ at each time stamp.
Finally $g(\cdot, \vec{w})$ will automatically figure out the inherent relations between masks, designs and resist patterns.

\subsubsection{Relations to Learning-to-Optimize}
The ILILT system is a similar scheme to learning-to-optimize (L2O) solutions \cite{L2O}, particularly the algorithm unrolling approach,
which unrolls a truncated algorithm into multi-stage neural networks with independent parameters.
In contrast, ILILT benefits from the \textbf{weight-tied structure} that (1) significantly reduces model size for memory efficiency
and (2) generalizes better \cite{DEQ}.

\subsubsection{Practical Usage Scenario}

Lastly, ILILT imitates a real ILT solver during the inference runtime but with many benefits: 
(1) ILILT requires to query a lithography simulator in each step but updates the mask in a derivative-free manner. 
(2) ILILT requires significant less steps than traditional ILT solver.
(3) ILILT has the opportunity to outperform numerical ILT solvers due to strong mask priors learned from broad data points. 
(4) ILILT can serve as a real solver during chip design optimization.
One representative case is when a design is not well-optimized by an existing mask optimizer.
Because most AI solutions only learn a design-to-mask mapping, it is not possible to use them to fix partially optimized masks.
ILILT, however, does not have this limitation, and can take over from anywhere in the optimization trajectory.

\section{Experiments}
\label{sec:result}

\subsection{The Dataset and Configurations}
To demonstrate the effectiveness of the ILILT framework, we adopt the latest AI for computational lithography benchmark suite \texttt{LithoBench} \cite{lithobench},
where state-of-the-art mask optimization solutions are evaluated and will be used as comparison baselines.
LithoBench contains over 100K design clips 
and each covers an area of $2\mu m \times 2\mu m$.
Each data point contains a pair of chip design patterns and the ground truth masks generated by a numerical ILT solver. 
These design patterns and masks are rasterized with a $1nm^2$ pixel size, yielding $2048 \times 2048$ images.
Example designs are displayed in \Cref{fig:dataset}.
We can observe that in real world, mask optimization does not simply finetune the design like in \Cref{fig:litho-prox}. 
The process also produce \textit{sub-resolution assist features} (SRAFs) surrounding the main design pattern to improve its robustness against process variations.
This poses great challenge of ML models on optimization tasks.
\begin{figure}[tb!]
	\centering 
	\subfigure[Design]{\includegraphics[width=.13\textwidth,trim={10cm 10cm 10cm 10cm},clip]{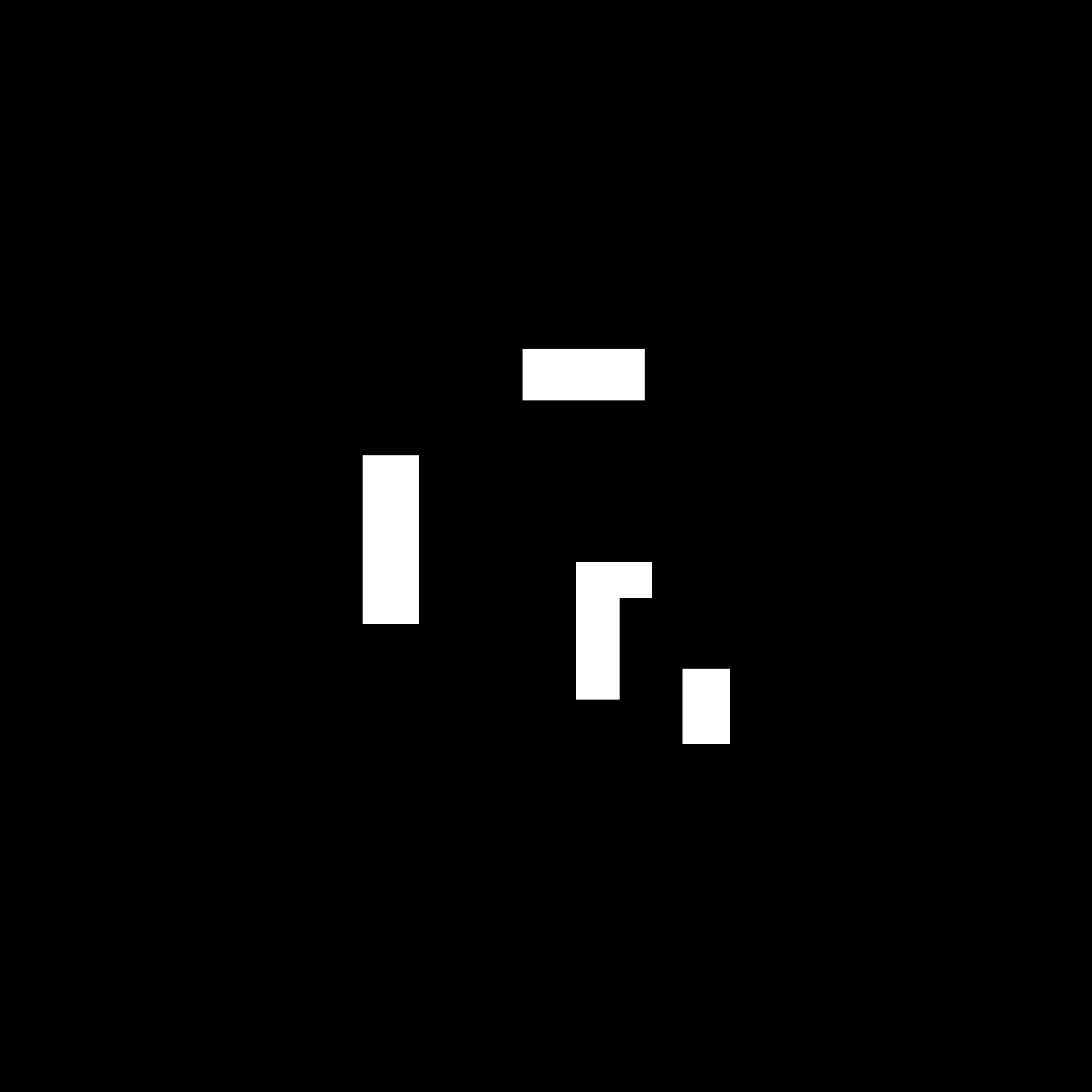}}
	\hspace{.3cm}
	\subfigure[Mask]{\includegraphics[width=.13\textwidth,trim={10cm 10cm 10cm 10cm},clip]{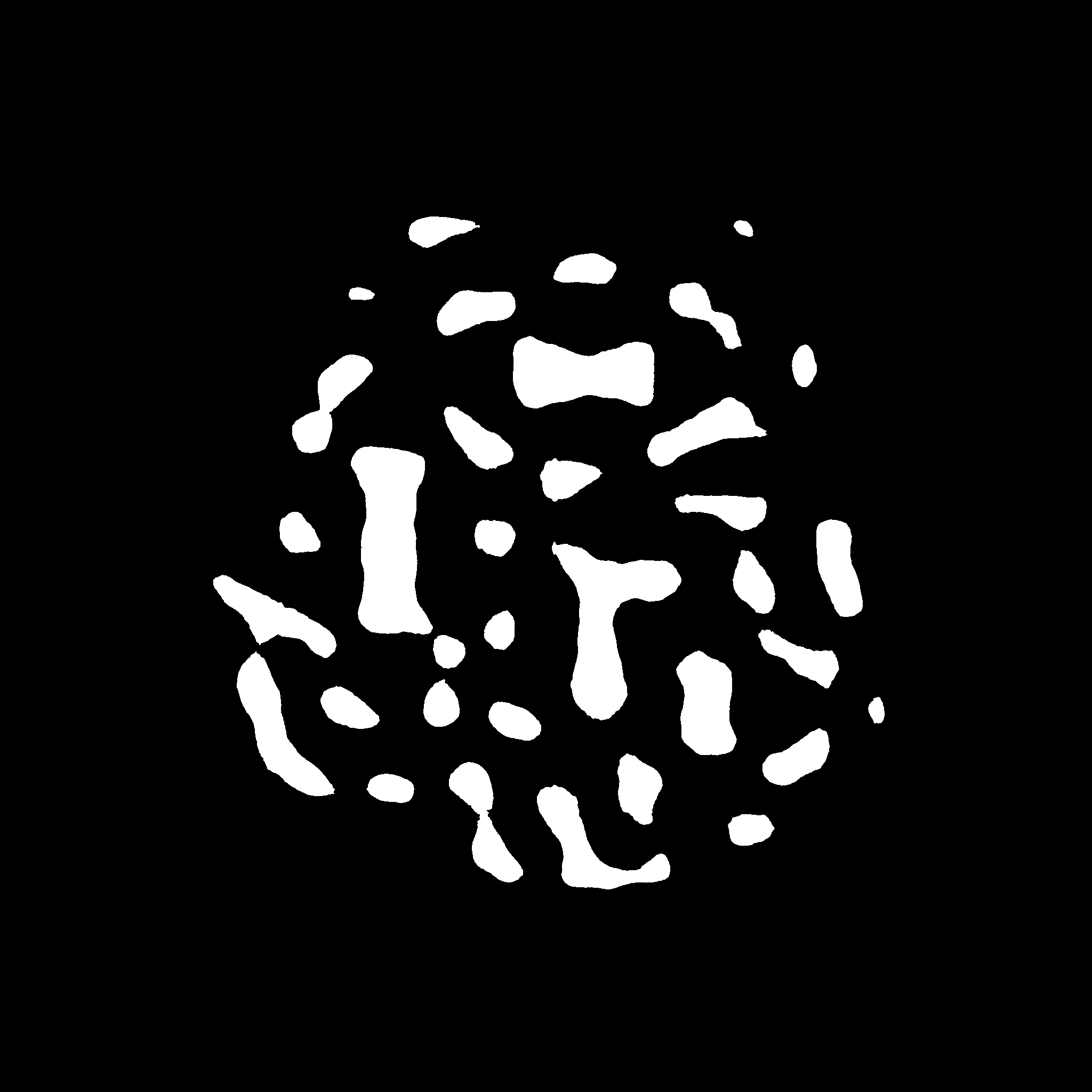}}
	\hspace{.3cm}
	\subfigure[Wafer]{\includegraphics[width=.13\textwidth,trim={10cm 10cm 10cm 10cm},clip]{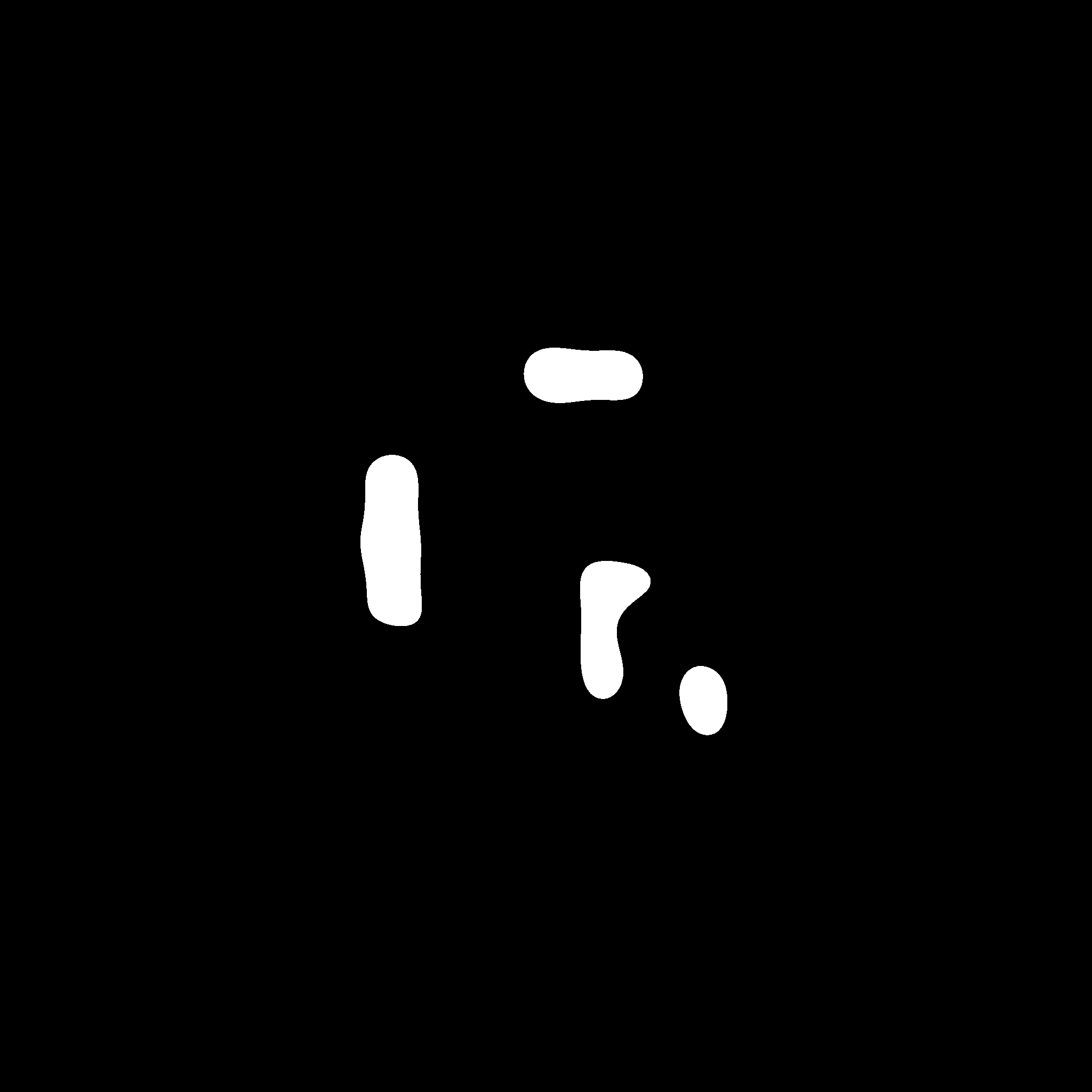}}
	\caption{Data example from LithoBench \cite{lithobench}. (a) Design image contains the patterns of original circuit devices or connectivity; (b) Mask image is the optimized design that is manufacturing-friendly; (c) Wafer image is the patterns on silicon given the mask image.}
	\label{fig:dataset}
\end{figure}

To deploy ILILT, we re-implement the lithography simulator as a pytorch layer that can be integrated in the data pipeline. 
Detailed configurations of the generator used to reproduce our results are listed in \Cref{tab:config}.
Here we adopt CFNO \cite{CFNO} as the backbone network for $g$ to evaluate ILILT.

\iftrue
\begin{table}[tb!]
	\centering
	\caption{Model Configuration.}
	\label{tab:config}
	\begin{tabular}{c|c}
		\toprule
		Item  & $g$ \\ \midrule
		\revise{Backbone} & \revise{CFNO/Pix2PixHD} \\
	Max Epoch &5    \\
	Initial Learning Rate & 0.004    \\
	Learning Rate Decay Policy	& step \\
	Optimizer & Adam\\
	Weight Decay & {0.0001}\\
	Loss &  \Cref{eq:trainililt}\\
	Sequence Unrolling Depth ($T$) & 4-8\\
	Batch Size & 4   \\ \bottomrule
	\end{tabular}
\end{table}
\fi 

\subsection{Evaluation Metrics}
\begin{figure}
	\centering
	\includegraphics[width=.2\textwidth]{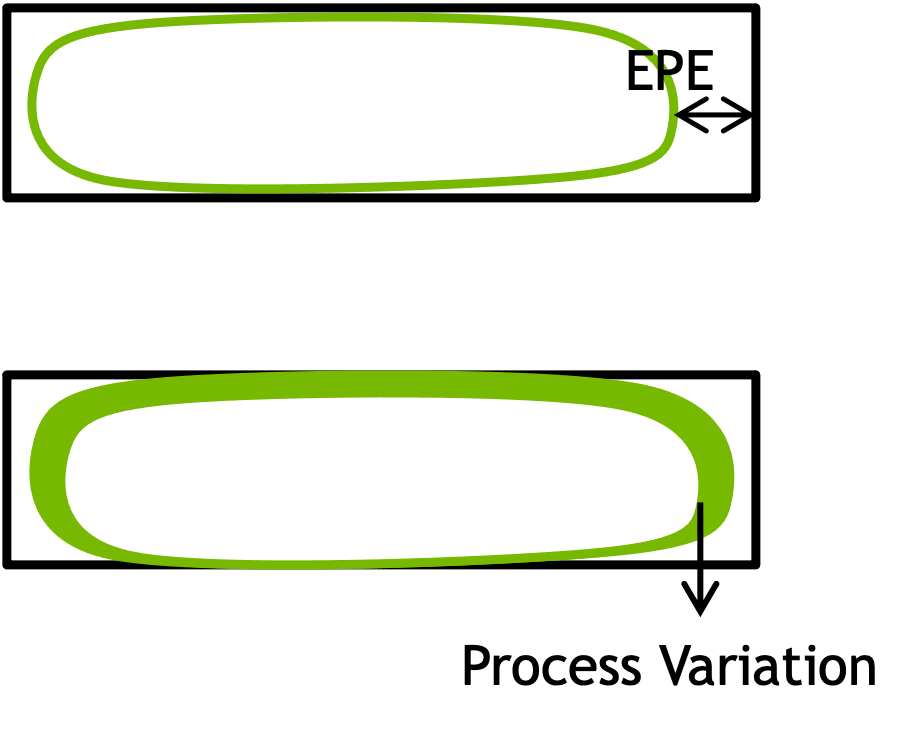}
	\caption{Lithography edge placement error and process variations. Smaller EPE and PVB area indicates better optimization QoR.}
	\label{fig:metric}
\end{figure}

\revise{As is typical for mask optimization tasks, we evaluate the final performance through the metrics that are widely adopted in the chip manufacturing industry}: edge placement error (EPE) violations and the process variation band (PVB) area.
In semiconductor manufacturing processes, the lithography conditions are not always as we expected due to systematic error and mechanical perturbations, which is the root cause of process variations. 
Therefore, we expect the design to be robust under different manufacturing conditions.
Typically, we call the expected lithography settings and configurations \textit{nominal condition},
and non-nominal condition will result in larger or smaller patterns on the silicon wafer.
The difference between the smallest pattern and the largest pattern will form a process variation band (PVB) as in \Cref{fig:metric}. 
A good mask optimization result should have a smaller number of EPE violations and a smaller PVB area.

\begin{definition}[EPE Violation\cite{OPC-ICCAD2013-Banerjee}]
	EPE is measured as the geometric distance between the target edge and the lithographic contour printed at the nominal condition. 
	If the EPE measured at a point is greater than a certain tolerance value, we call it an EPE violation.
        In our experiments, the EPE measuring points are sampled at 40$nm$ interval.
\end{definition}

\begin{definition}[PVB Area\cite{OPC-ICCAD2013-Banerjee}]
	This is evaluated by running lithography simulations at different process conditions on the final mask solution.
	Once run, the total area of the process variation band will be defined as PVB Area.
\end{definition}

\subsection{Comparison with State-of-the-Art}
In the first experiment, we compare the performance of ILILT with state-of-the-art mask optimizers.
A quantitative comparison is listed in \Cref{tab:result}.
Columns ``EPE'', ``PVB'' and ``Throughput (s/tile)'' indicate the count of EPE violations, PVB Area in terms of $nm^2$ and optimization throughput, respectively;
``\revise{GPU-ILT}'' \cite{OPC-DAC2023-Sun} is the latest academic numerical ILT engine with GPU acceleration, which will also be the engine used to generate the training data;
``GAN-OPC'' \cite{OPC-DAC2018-Yang} is one of the early attempts using machine learning for mask optimization following the scheme in \Cref{fig:iltscheme-2}. GAN-OPC employs UNet as a generator backbone.
\revise{``Pix2PixHD'' \cite{pix2pixhd} is another representative CNN-based backbone that enjoys better generalization capability for high-resolution image generation, which fits design-mask translation task well.}
``Neural-ILT'' \cite{OPC-ICCAD2020-NeuralILT} is a follow-up of GAN-OPC, which adopts a true lithography simulator to produce the gradient for generator training. 
``CFNO'' \cite{CFNO} is an FNO-based design that preserves the global perception of the Fourier neural operator while saving computation with ViT's patch embedding.
CFNO and Pix2PixHD will also be the backbone frameworks of the ILILT generator, and we will show the advantage of the ILILT data pipeline over the standalone generator. 
ILILT-$X$ lists the results for $X$ unrolling steps.

\revise{Overall, the ILILT with Pix2PixHD backbone achieves the smallest EPE violation count among all other ML-based solutions with $0.08$ average EPE violations (achieved by $T=8$),
which outperforms the golden numerical solver.
ILILT-CFNO though behaves slightly worse than ILILT-P2PHD, it can still exhibit comparable results with ``GPU-ILT'' with a 400$\times$ smaller model size.} 
All methods do not show a significant advantage on PVB area, and this is because the optimization flow in these solutions do not explicitly target process variations.
Lastly, the ILILT already achieves better mask quality than the other three ML-based solutions without the refinement from the legacy ILT engine. 
We also show a 10$\times$ speedup over the numerical ILT solver.

\revise{
The ILILT framework also ensembles the beneath idea of learning-to-optimize (L2O), which unrolls optimization steps on multiple layers of neural networks whose weights are not shared.
As a comparison, we also implement two standard L2O flows with CFNO and Pix2PixHD as backbones respectively.
We choose the 8 unrolling steps that have similar computing costs to the best ILILT settings.
As shown in \Cref{tab:result}, L2O does not preserve the performance of standalone backbone models with a significant increase in EPE violations. 
L2O unrolls the ILT problem into a fixed number of optimization steps, which are parameterized by different neural network layers. 
Such a model, with 8$\times$ trainable parameters in our case, requires more converging training efforts.
Specifically, the error of each optimization step will propagate to the next level which causes performance degradation compared to standalone backbone models.}

\begin{table}[tb!]
\centering
\caption{Comparison with State-of-the-Arts.}
\label{tab:result}
\setlength{\tabcolsep}{2pt}
\renewcommand{\arraystretch}{1.3}
\revise{
\begin{tabular}{c|ccc}
\toprule
Method     & EPE & PVB  & Throughput\\ \midrule
GPU-ILT \cite{OPC-DAC2023-Sun}    & 0.21 & 4656 & 4        \\
GAN-OPC  \cite{OPC-DAC2018-Yang}  & 8.3 & 6686 & 0.008   \\
Neural-ILT \cite{OPC-ICCAD2020-NeuralILT}  & 6.2 & 8537 & 10       \\
Pix2PixHD \cite{pix2pixhd} &0.12 &4685 &0.1 \\
CFNO \cite{CFNO} & 0.51    & 4623     &  0.06       \\ \midrule
L2O-CFNO-8-Step  &1.59 &4436 &0.6 \\
L2O-P2PHD-8-Step  & 0.87 &4526 &0.5       \\ \midrule
ILILT-CFNO-2-130K (ours)   &  0.39   & 4704 & 0.1  \\
ILILT-CFNO-4-130K (ours)   & 0.39    & 4667 & 0.2  \\
ILILT-CFNO-6-130K (ours)   &  0.35   & 4657 & 0.3    \\
ILILT-CFNO-8-130K (ours)   &  0.25 & 4674  &  0.4    \\ 
ILILT-P2PHD-2-45M (ours)   &  0.13  & 5056 & 0.2  \\
ILILT-P2PHD-4-45M (ours)   & 0.13    & 4680 & 0.4  \\
ILILT-P2PHD-6-45M (ours)   &  0.11   & 4670 & 0.6    \\
ILILT-P2PHD-8-45M (ours)   &  \textbf{0.08} & 4695   & 0.8    \\ \bottomrule
\end{tabular}}
\end{table}

\subsection{Generalization of ILILT}
\revise{
ILILT also shows better generalization capability thanks to its weight-tied structure. 
This can be observed in the superior performance boosting over standalone CNN and Neural Operator-based backbone models.
We can observe that with the weight-tied sequence training on CFNO and Pix2PixHD, the mask quality has an improvement of 50\% and 33\% reduced EPE violation.
Although the unrolled sequence induces additional runtime, the model inference time can be ignored when compared to the entire physical implementation and manufacturing flow.
}

\revise{
\subsection{Self-learning Capability}
Usually, ML-based solutions rely highly on the dataset quality, and hence the performance of traditional ILT solvers. 
However, ILT is naturally non-convex and suffers from sub-optimal solutions.
Under such a scenario, ILILT can take advantage of its capability to outperform numerical solvers in certain chip designs, providing faster iteration for dataset updates and serving as a powerful auxiliary candidate for traditional numerical solvers.
}

\subsection{ILILT Optimization Flow}
The learned mask optimization process is depicted in \Cref{fig:trajectory},
where we visualize the growth of a mask along the unrolled sequence for a given design.
We bring the model trained with unrolling step T=4 but keep the loop running until T=8. 
We show that the mask reaches the best quality at the configured unrolling depth. 
Interestingly, the ILILT also learns a fixed-point iteration such that if we continuously feed the mask beyond the unrolling depth, the model will no longer make significant changes to the mask except minor adjustments.
At T=8, the ILILT engine tries to remove the rule-violating artifact from the middle-bottom region. 
This property brings more application scenarios of the ILILT framework that include producing initial solutions and performing fine-grained fixes at the post-optimization stage.
Another benefit of ILILT is its stable optimization trajectory,
because quality fluctuation is common during the optimization runtime of numerical ILT solvers \cite{OPC-DAC2014-Gao,OPC-ICCAD2017-Ma,OPC-DATE2021-Yu,OPC-DAC2023-Sun}, 
causing additional overhead to identify good solutions.

\iftrue
\begin{figure*}[tb!]
	\centering 
	\subfigure[T=1]{\includegraphics[width=.11\textwidth,trim={25cm 26cm 10cm 9cm},clip]{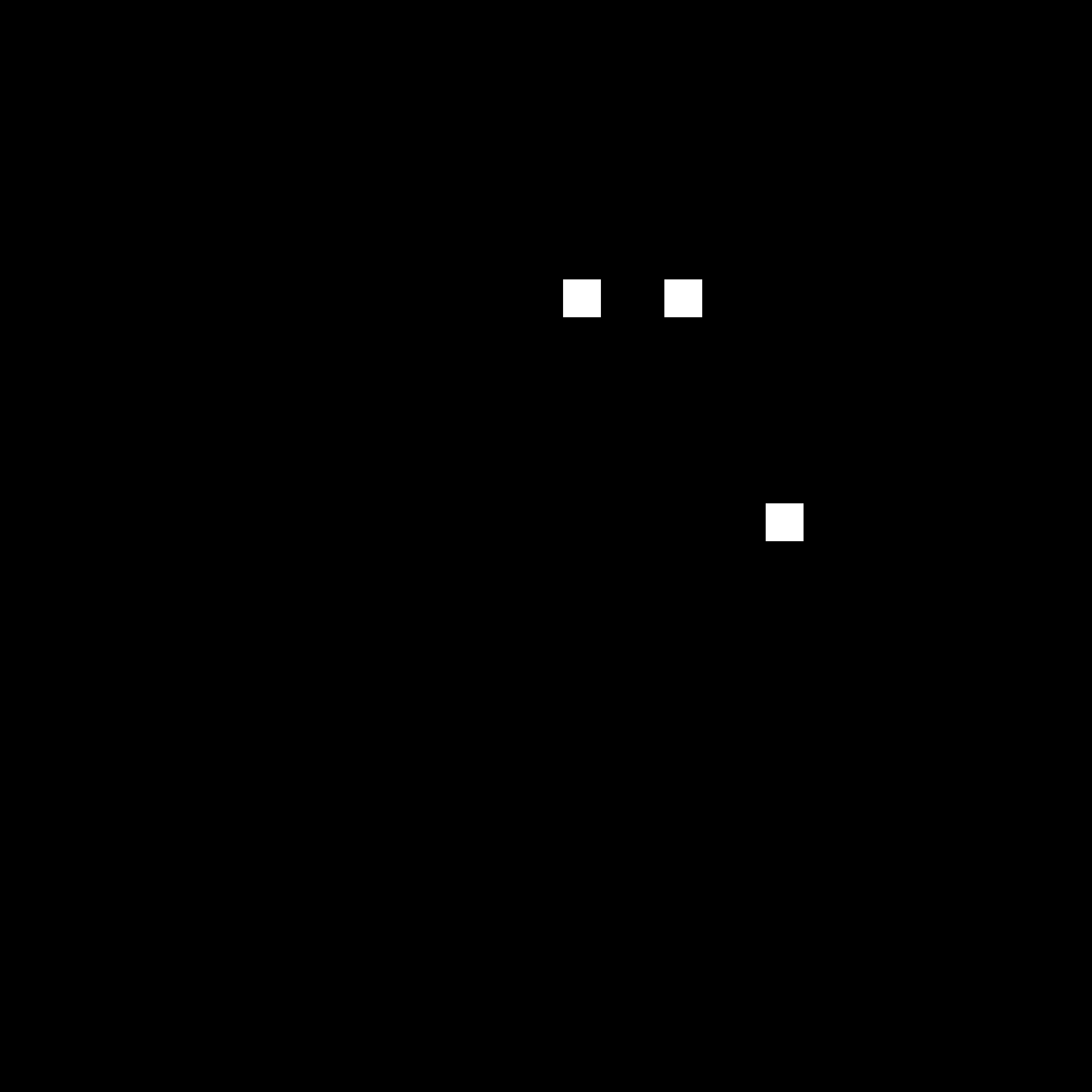}}
	\subfigure[T=2]{\includegraphics[width=.11\textwidth,trim={25cm 26cm 10cm 9cm},clip]{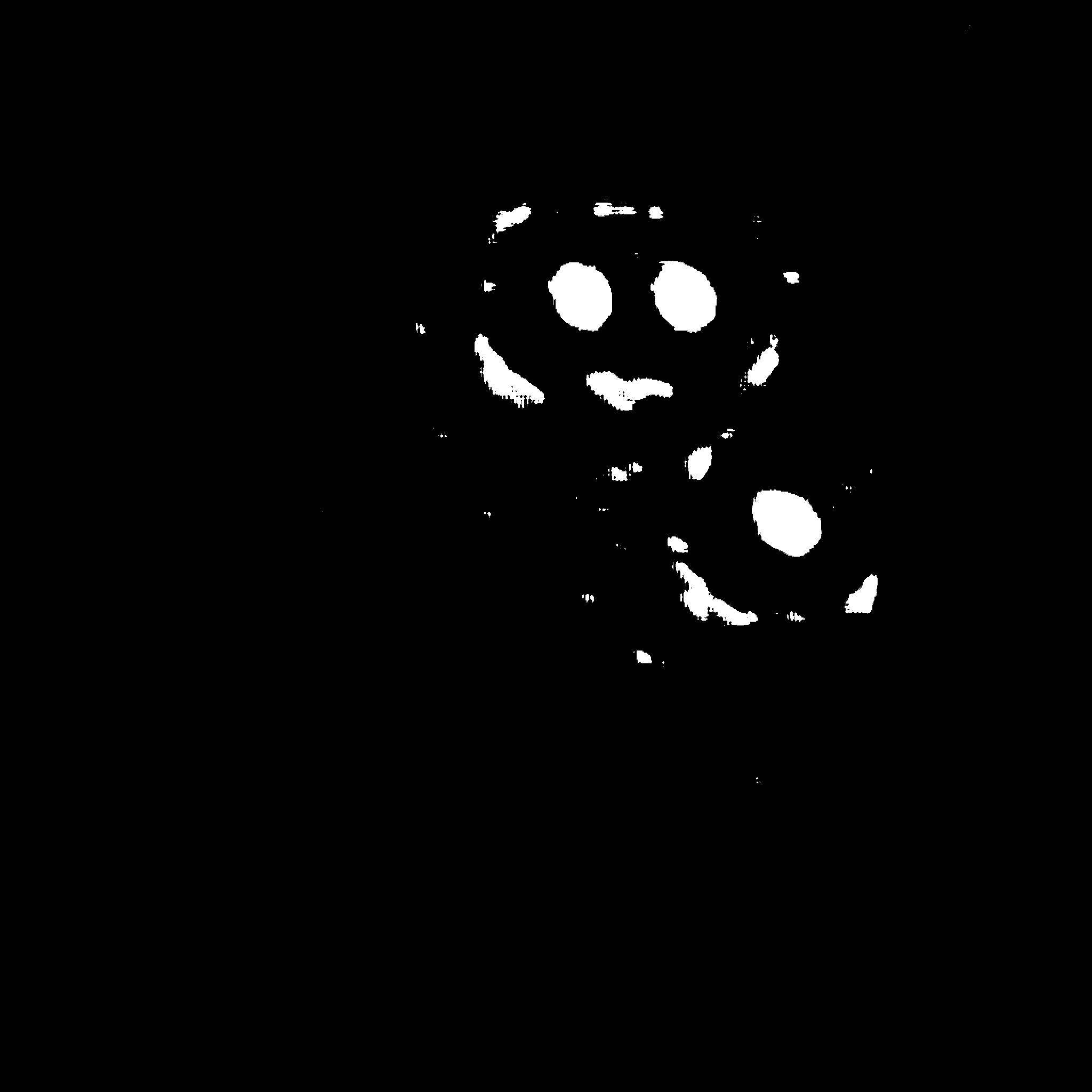}}
	\subfigure[T=3]{\includegraphics[width=.11\textwidth,trim={25cm 26cm 10cm 9cm},clip]{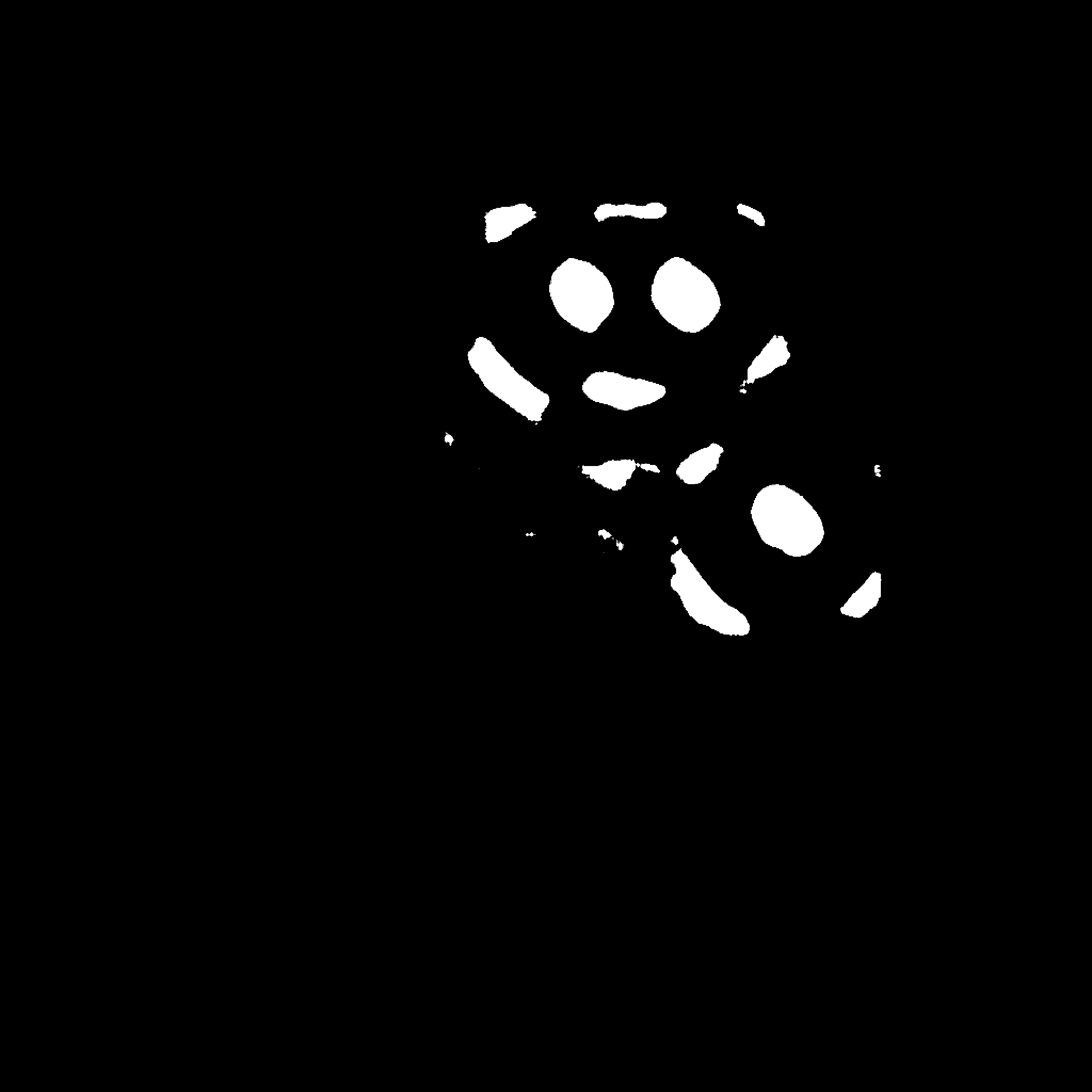}}
	\subfigure[T=4]{\includegraphics[width=.11\textwidth,trim={25cm 26cm 10cm 9cm},clip]{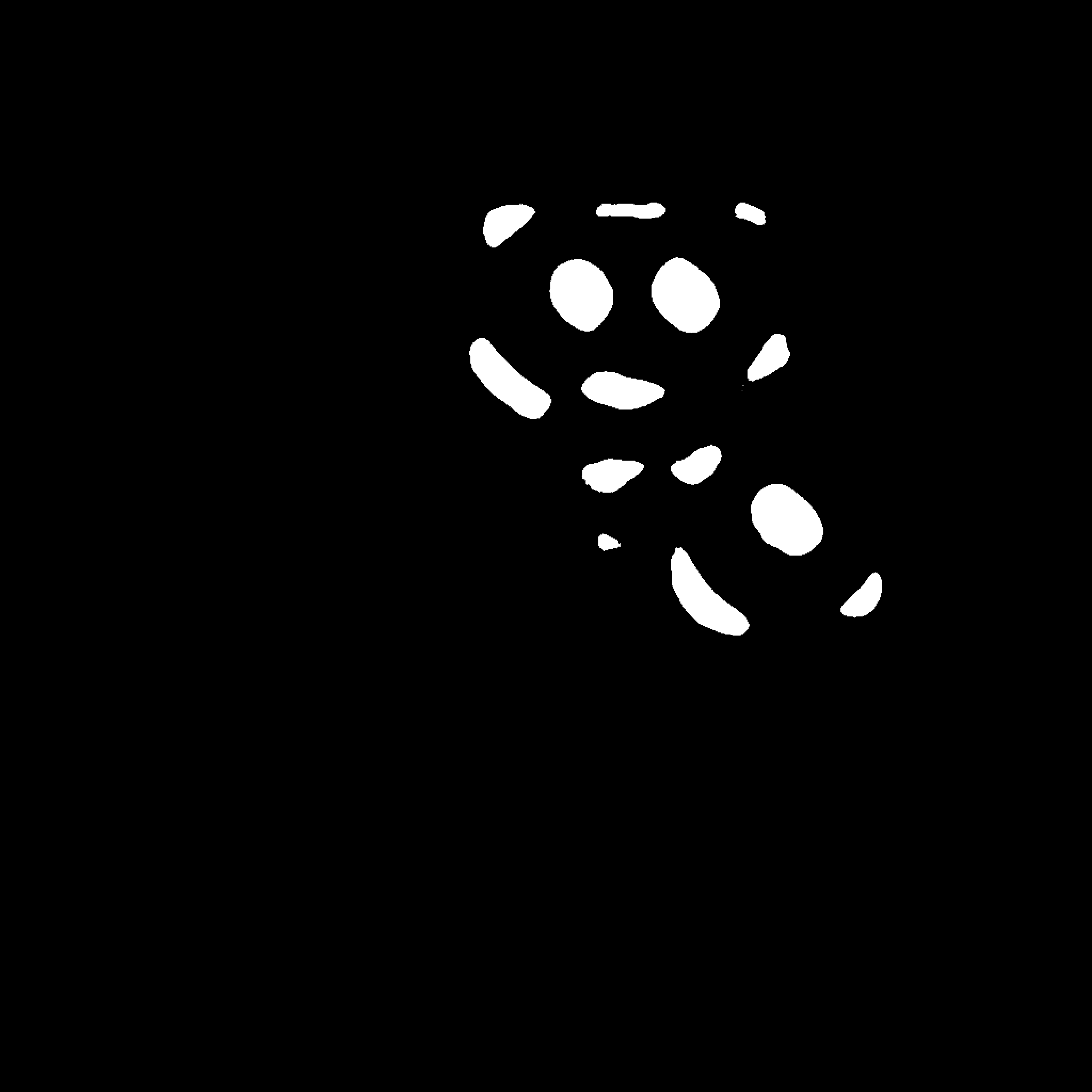}}
	\subfigure[T=5]{\includegraphics[width=.11\textwidth,trim={25cm 26cm 10cm 9cm},clip]{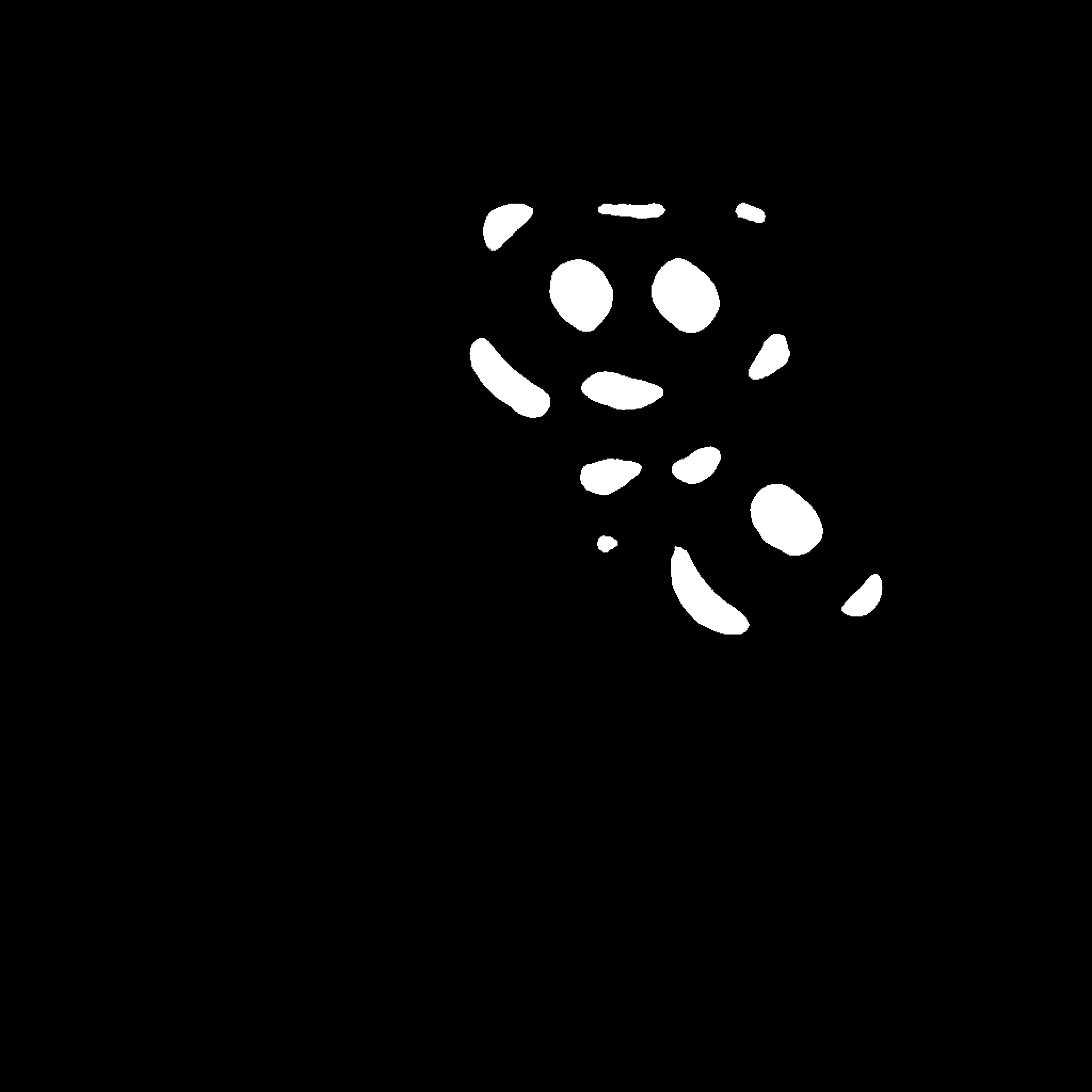}}
	\subfigure[T=6]{\includegraphics[width=.11\textwidth,trim={25cm 26cm 10cm 9cm},clip]{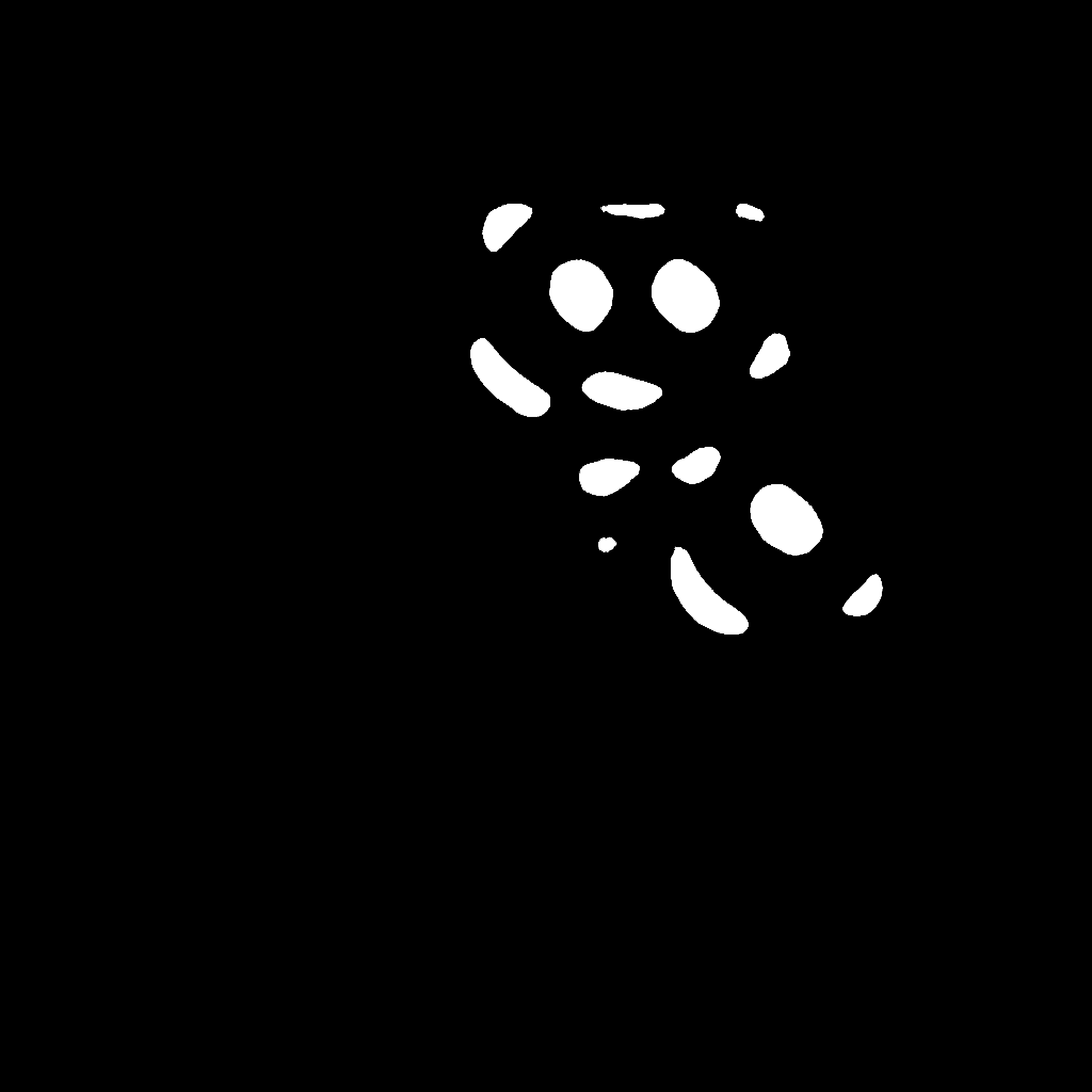}} 
	\subfigure[T=7]{\includegraphics[width=.11\textwidth,trim={25cm 26cm 10cm 9cm},clip]{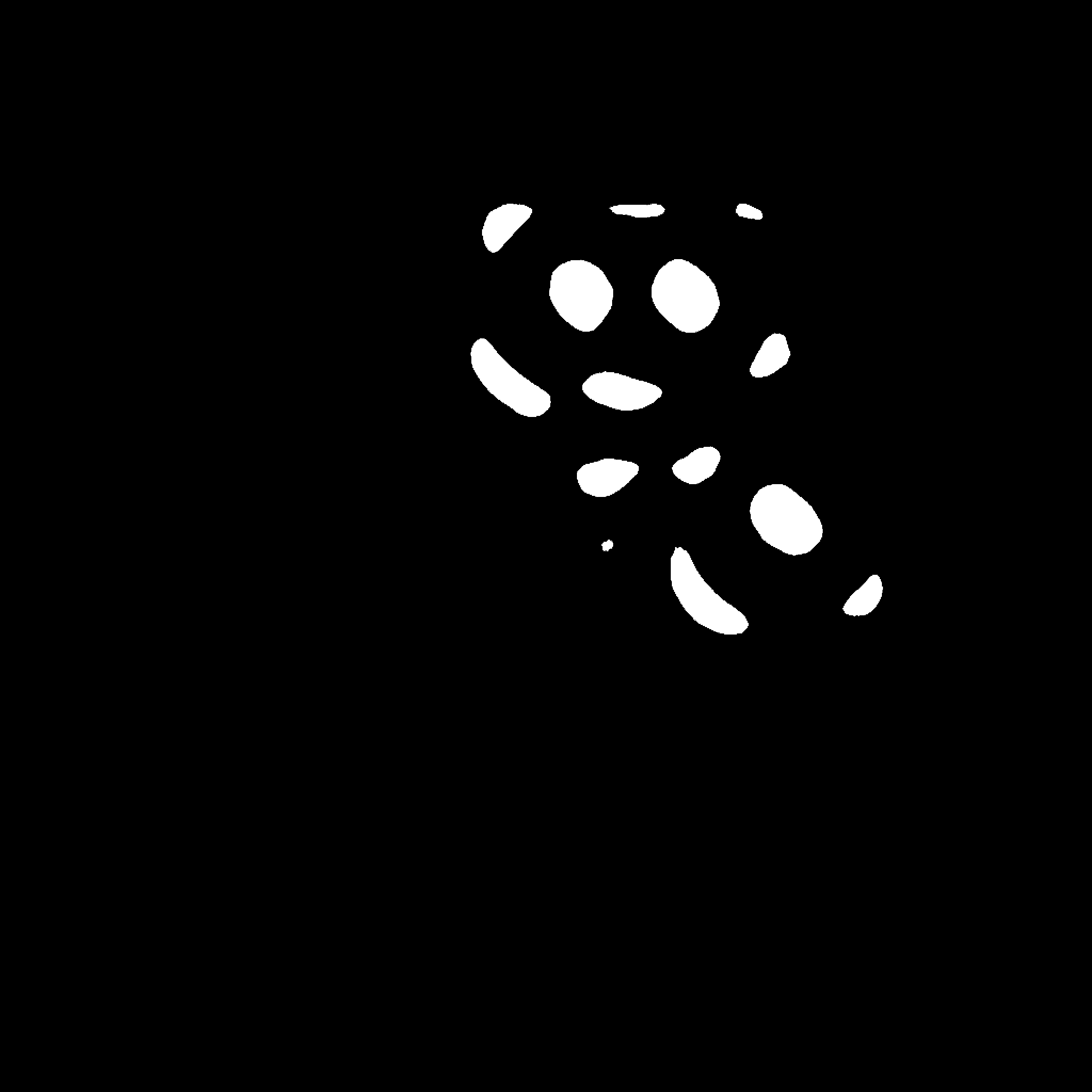}} 
	\subfigure[T=8]{\includegraphics[width=.11\textwidth,trim={25cm 26cm 10cm 9cm},clip]{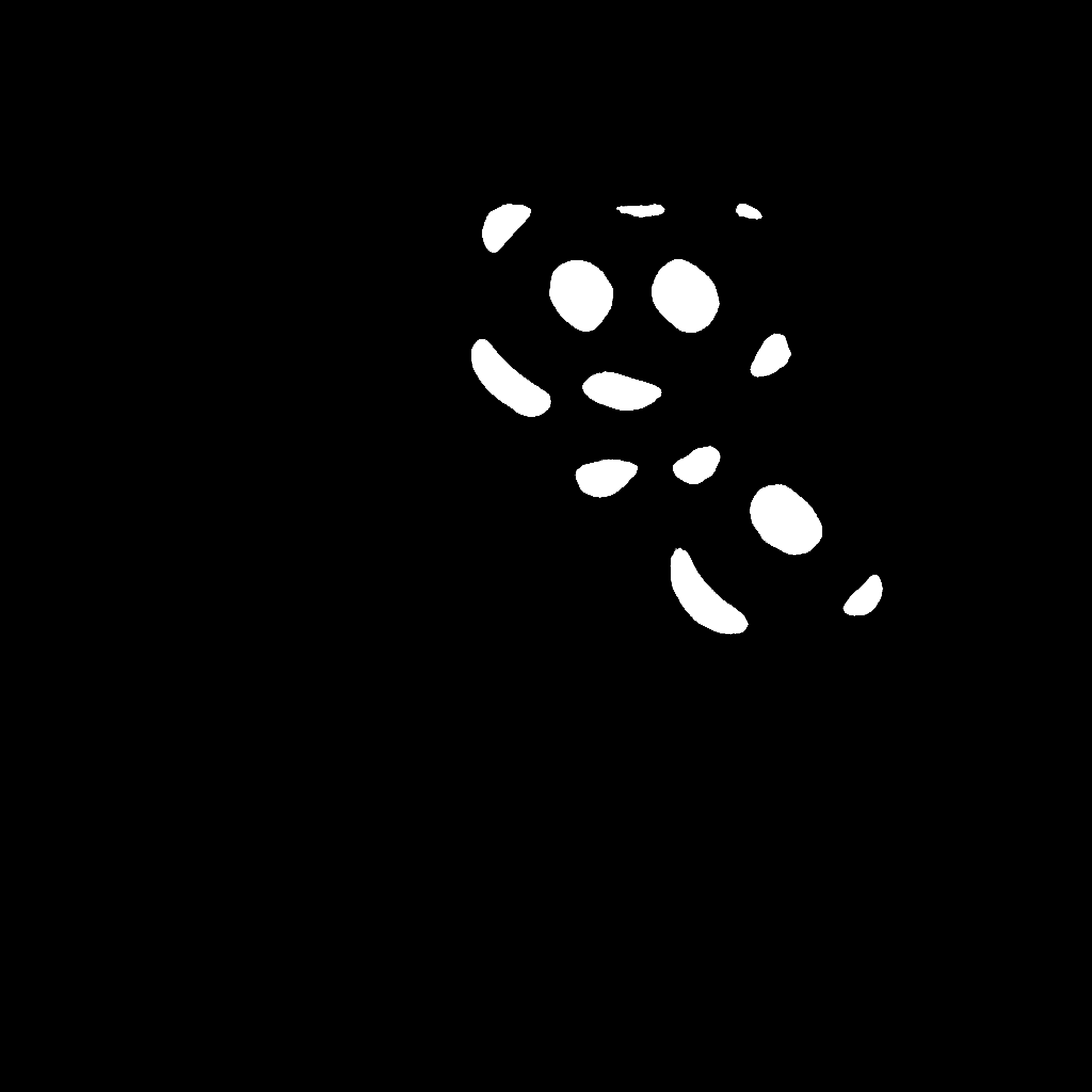}} 
 \\
	\subfigure[EPE=12]{\includegraphics[width=.11\textwidth,trim={25cm 26cm 10cm 9cm},clip]{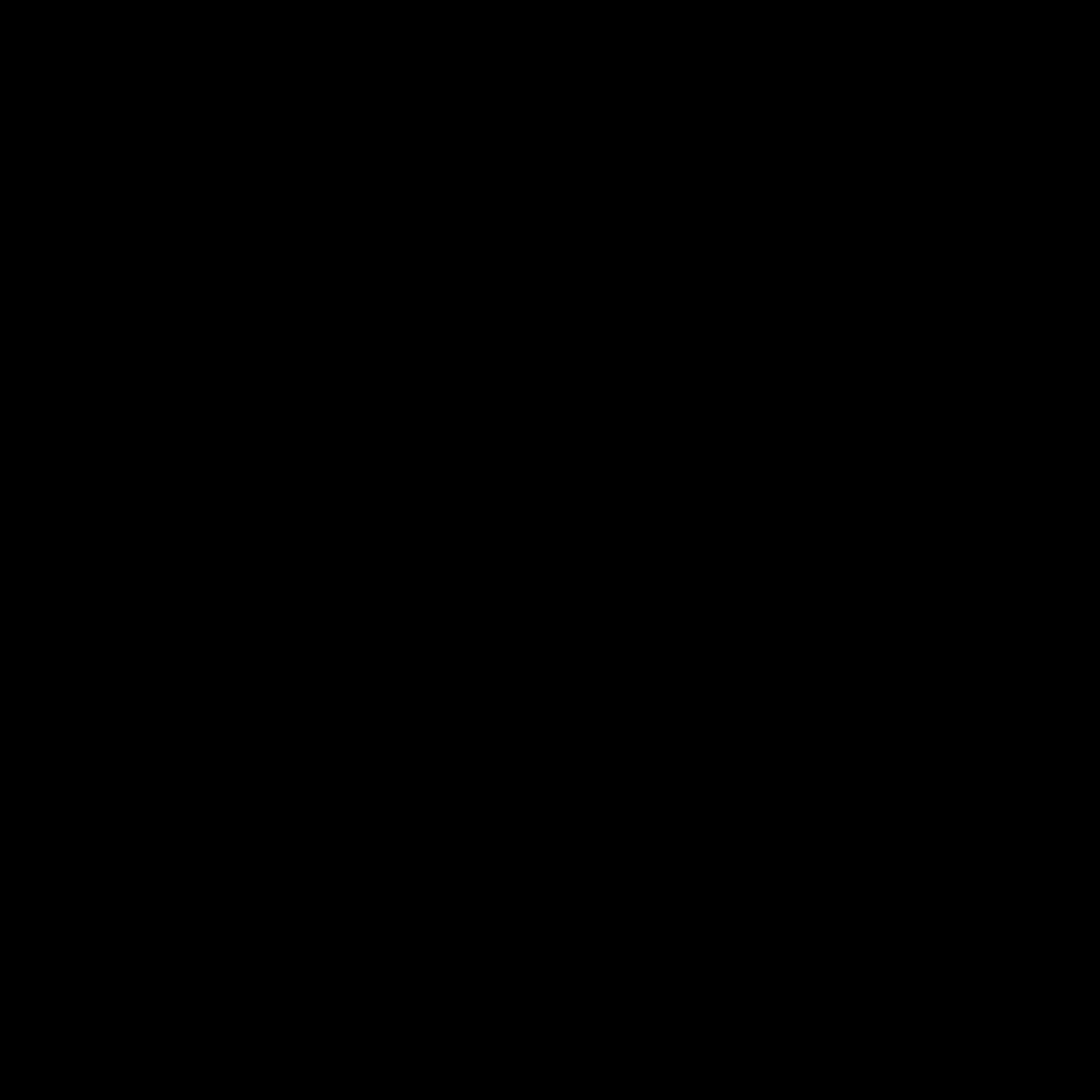}}
	\subfigure[EPE=12]{\includegraphics[width=.11\textwidth,trim={25cm 26cm 10cm 9cm},clip]{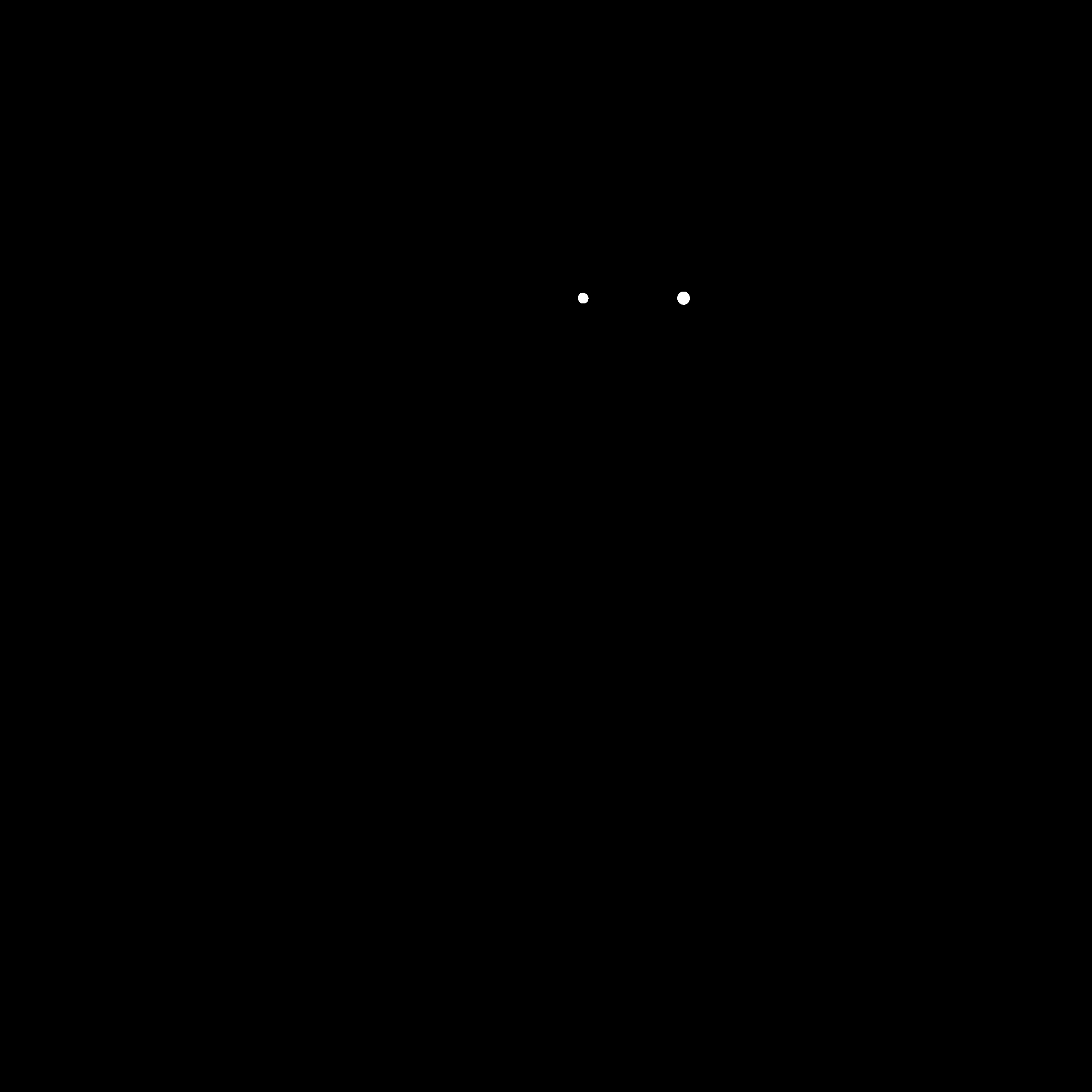}}
	\subfigure[EPE=0]{\includegraphics[width=.11\textwidth,trim={25cm 26cm 10cm 9cm},clip]{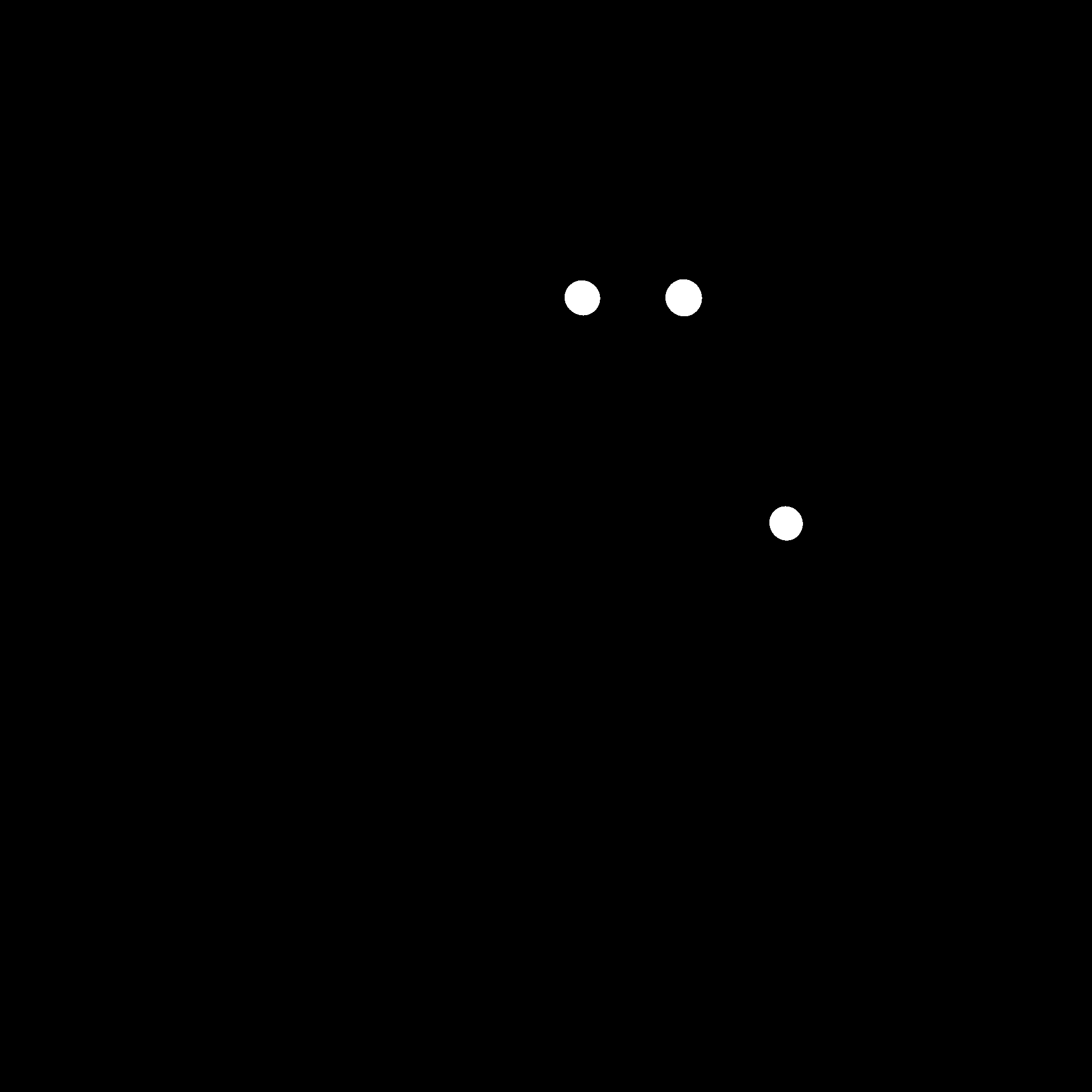}}
	\subfigure[EPE=0]{\includegraphics[width=.11\textwidth,trim={25cm 26cm 10cm 9cm},clip]{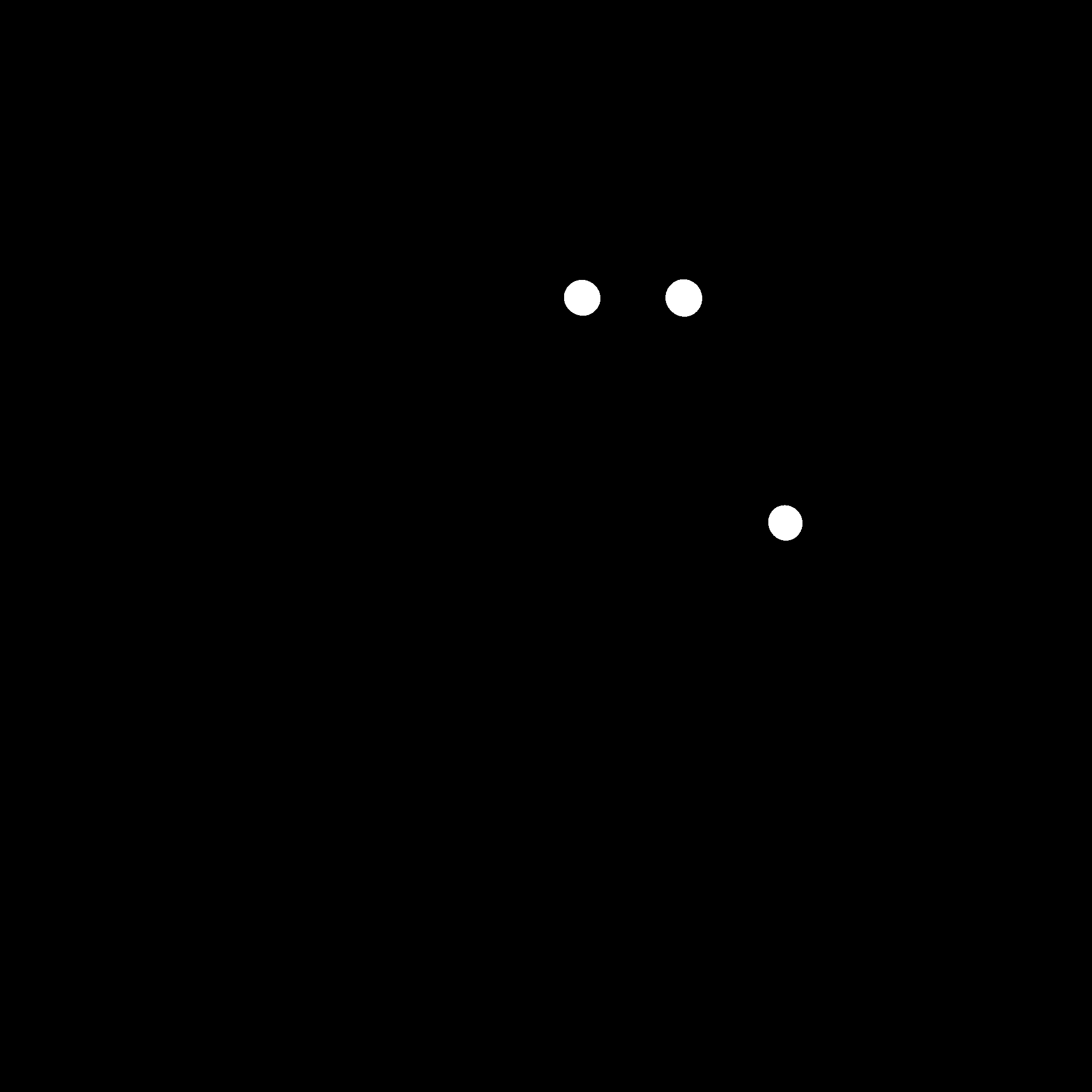}}
	\subfigure[EPE=0]{\includegraphics[width=.11\textwidth,trim={25cm 26cm 10cm 9cm},clip]{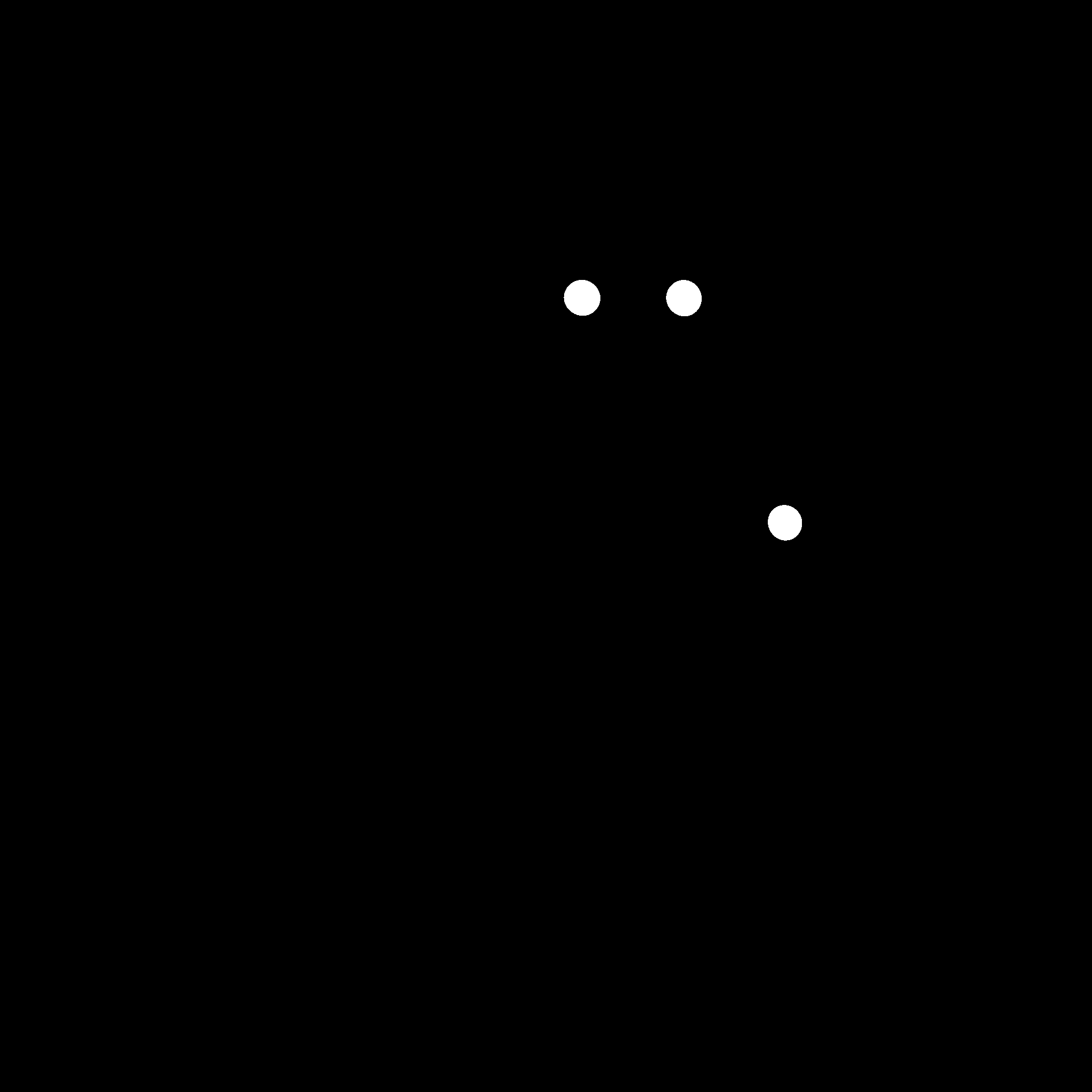}}
	\subfigure[EPE=0]{\includegraphics[width=.11\textwidth,trim={25cm 26cm 10cm 9cm},clip]{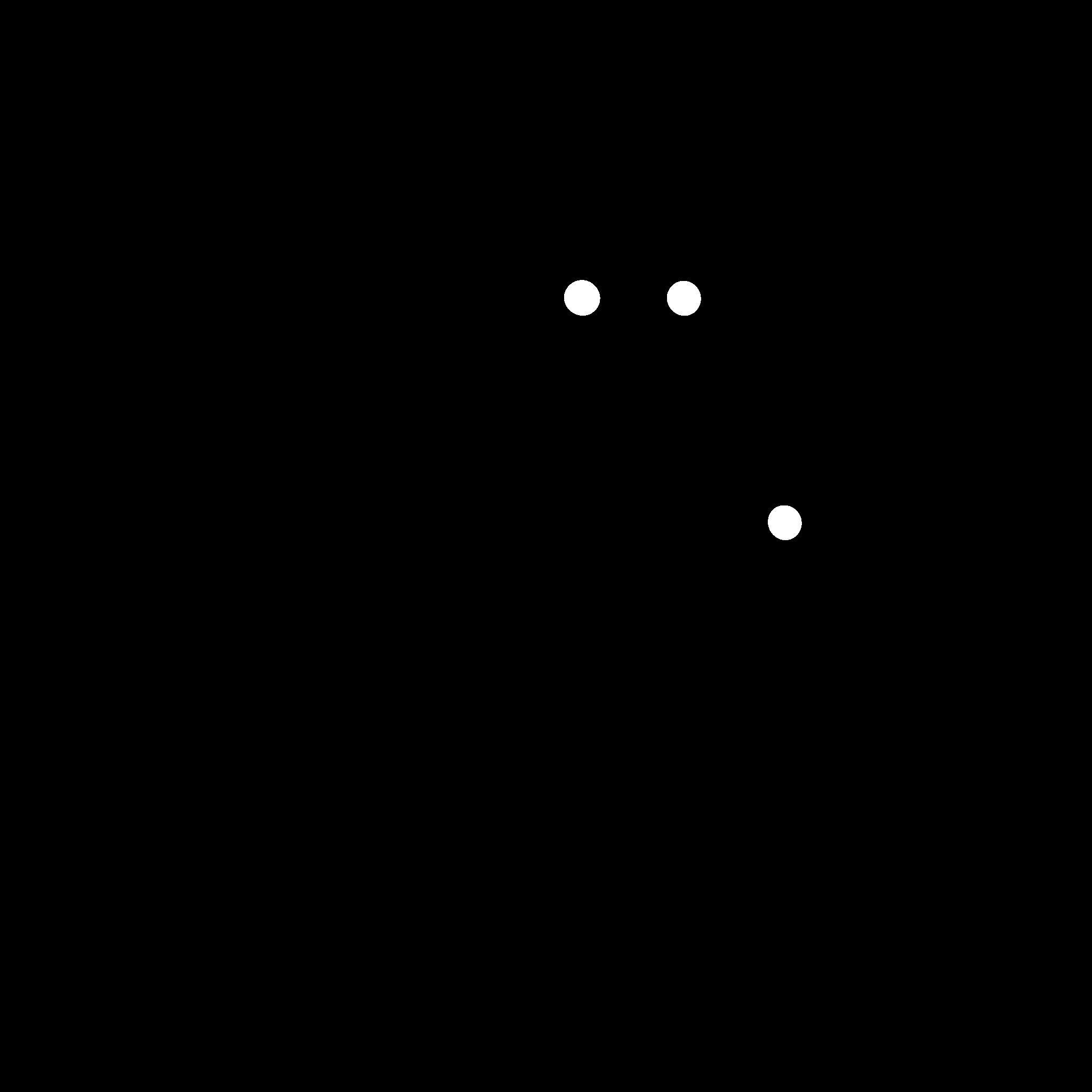}} 
	\subfigure[EPE=0]{\includegraphics[width=.11\textwidth,trim={25cm 26cm 10cm 9cm},clip]{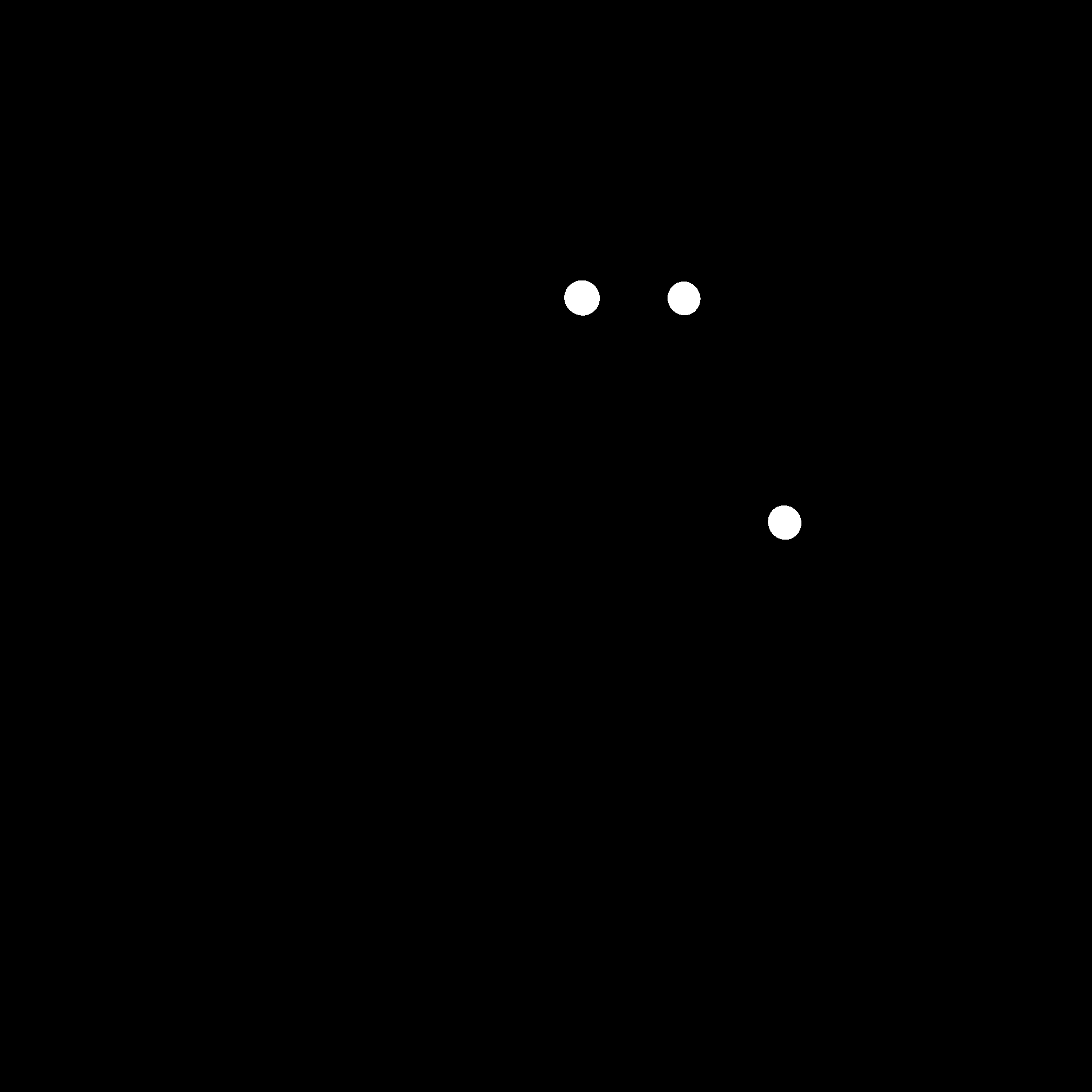}}
	\subfigure[EPE=0]{\includegraphics[width=.11\textwidth,trim={25cm 26cm 10cm 9cm},clip]{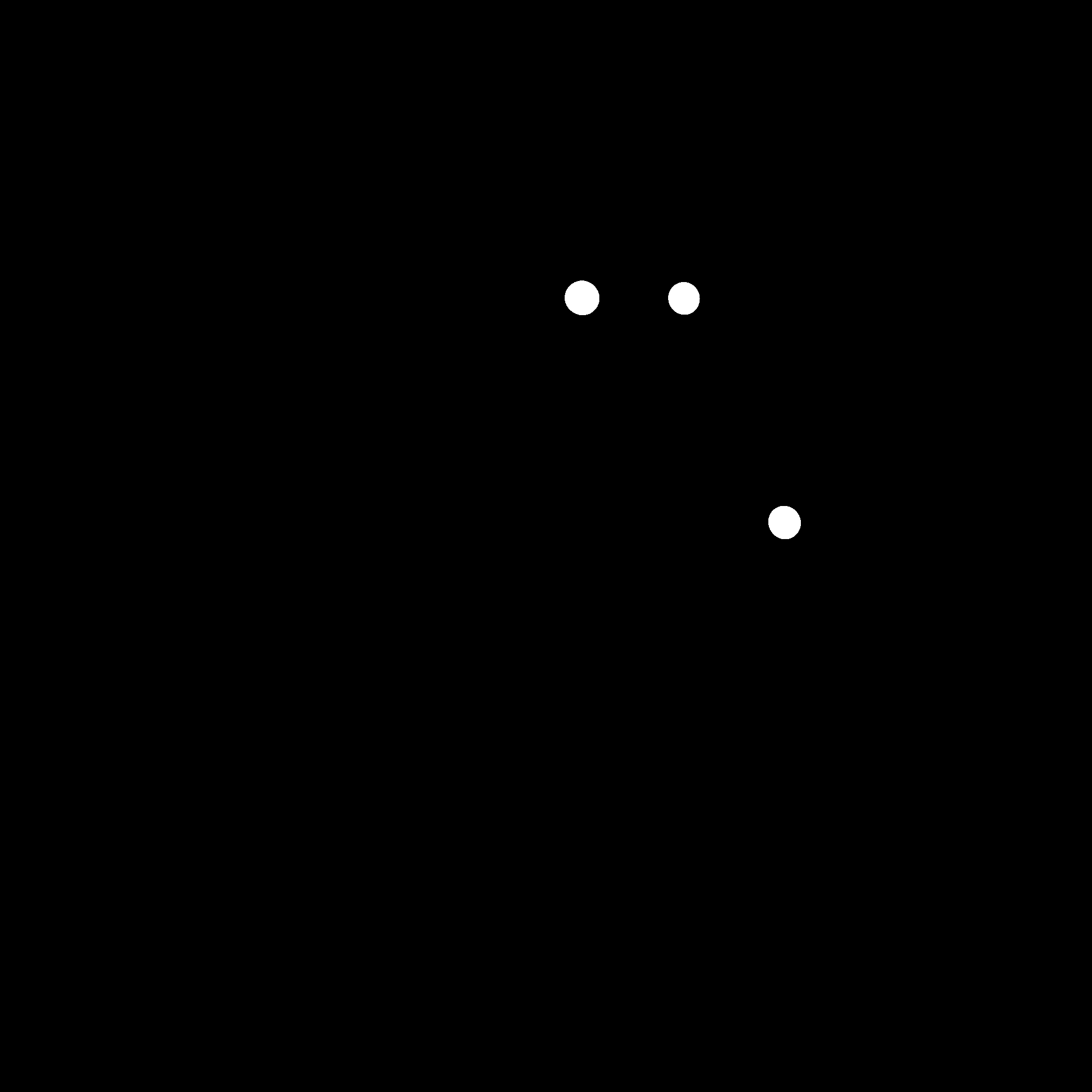}}

	\caption{Visualization of the optimization trajectory beyond the maximum unrolling depth. We use the model trained with unrolling step T=4, and mask stays stable after the 4th step with only minor tuning. 
	(a)--(h) correspond to the generated masks generated from time stamps 1--8. 
	(i)--(p) are the lithography simulated wafer image with their corresponding EPE violation count.}
	\label{fig:trajectory}
\end{figure*}
\fi

\subsection{Investigation of Unrolling Depth}
Here we discuss the effects of the maximum unrolling depth.
The ILILT is trained using different max unrolling depths ($T$) with $T \in \{2,4,6,8\}$.
With PVB area being similar, we observe better results happen in larger unrolling steps in the configuration.
When $T$ is small, ILILT does not benefit too much from the weight-tied structure and exhibits larger EPE violation counts.
Instead, a larger $T$ will put stronger regularization on the model and enable temporal feature learning to increase the model generality.
\revise{However, a longer unrolling sequence will linearly increase the computing cost as well as training efforts.
In this paper, we experimentally show the performance boosting saturates at $T=8$ for a fixed number of training epochs.}

\subsection{Mask Prior Learning}

\begin{figure*}[tb!]
	\centering 
	\subfigure[Design]{\includegraphics[width=.13\textwidth,trim={15cm 20cm 5cm 0cm},clip]{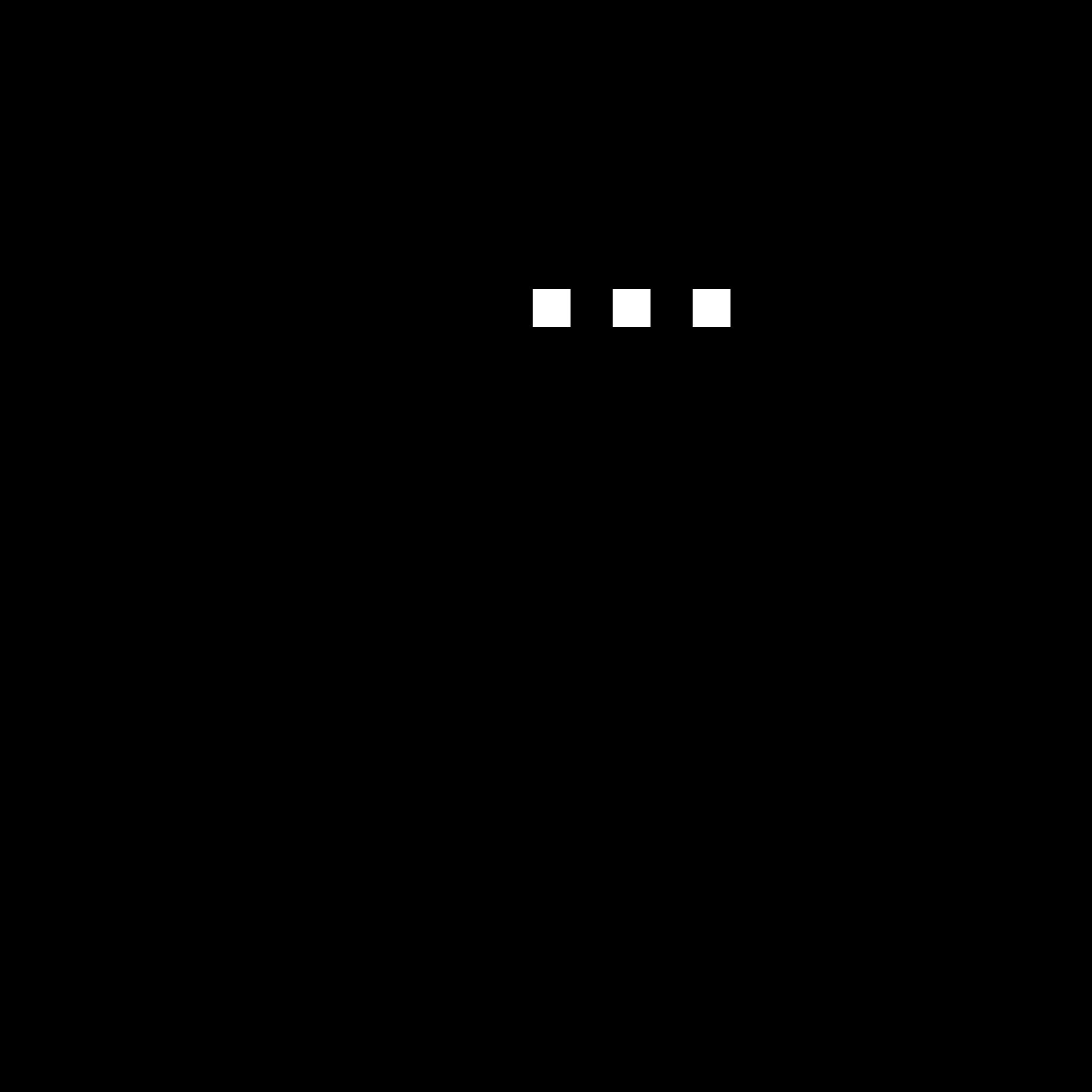}}
	\hspace{.3cm}
	\subfigure[ILT-Mask]{\includegraphics[width=.13\textwidth,trim={15cm 20cm 5cm 0cm},clip]{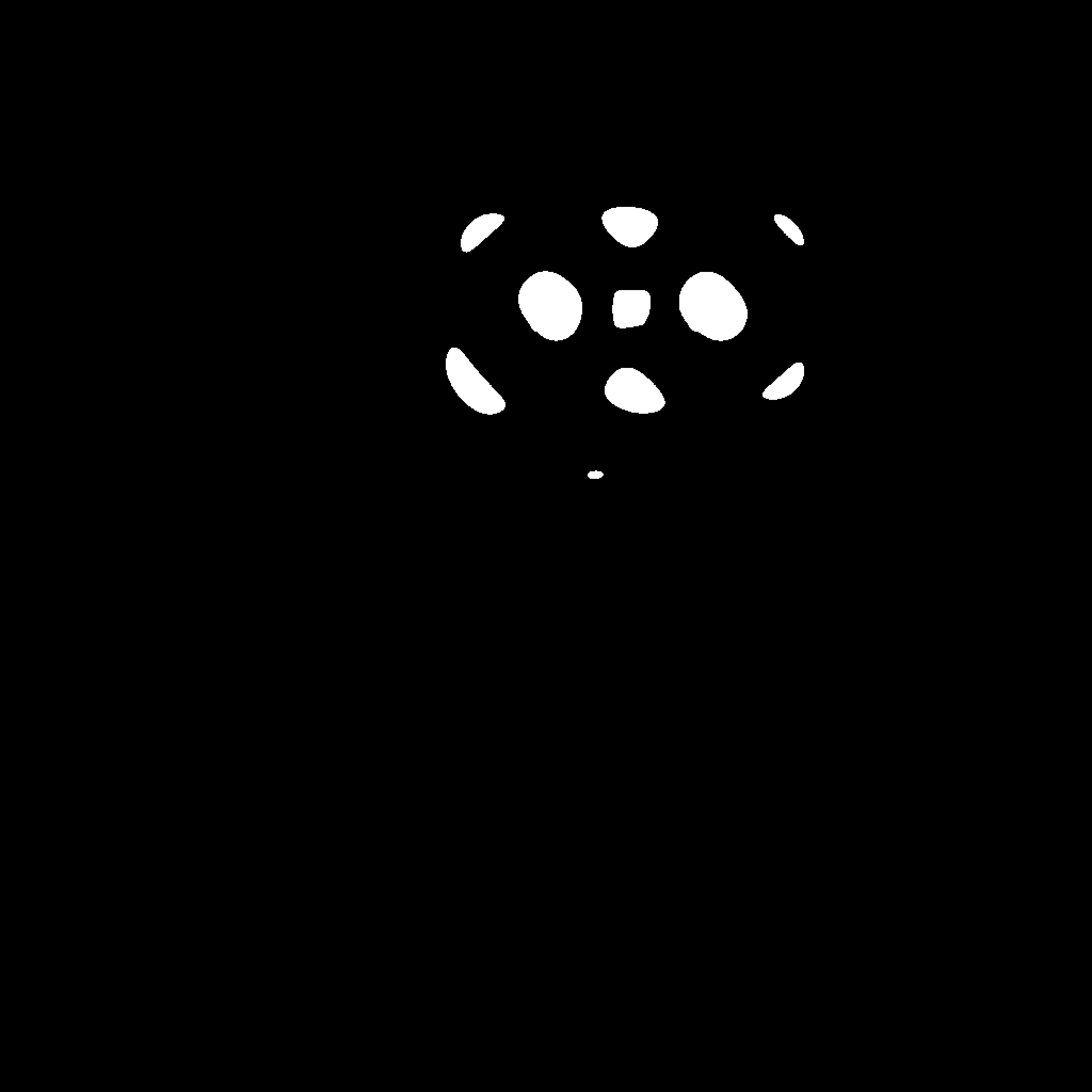} \label{fig:mprior-realb}}
	\hspace{.3cm}
	\subfigure[ILT-Wafer]{\includegraphics[width=.13\textwidth,trim={15cm 20cm 5cm 0cm},clip]{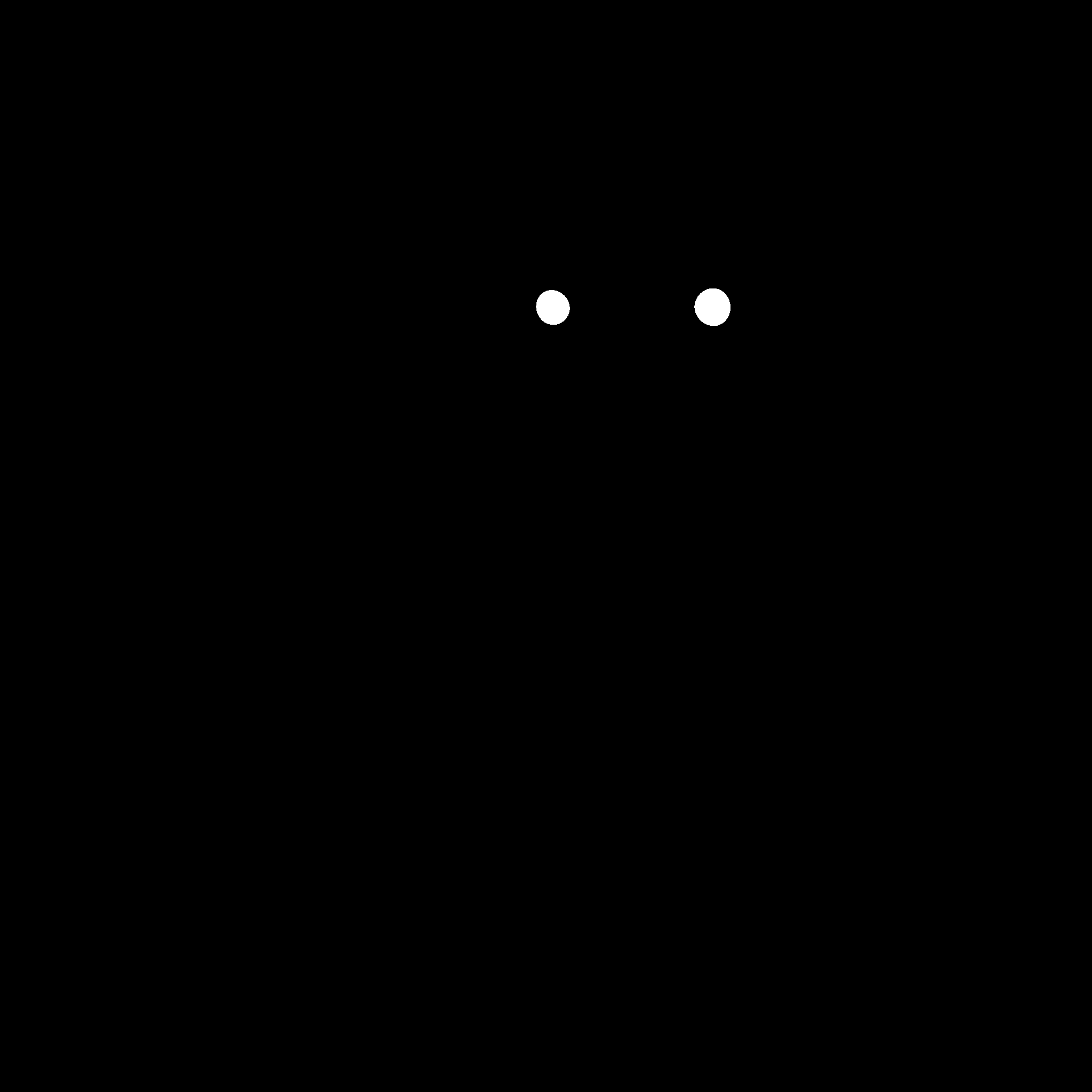}}
	\hspace{.3cm}
	\subfigure[ILILT-Mask]{\includegraphics[width=.13\textwidth,trim={16cm 12cm 4cm 8cm},clip]{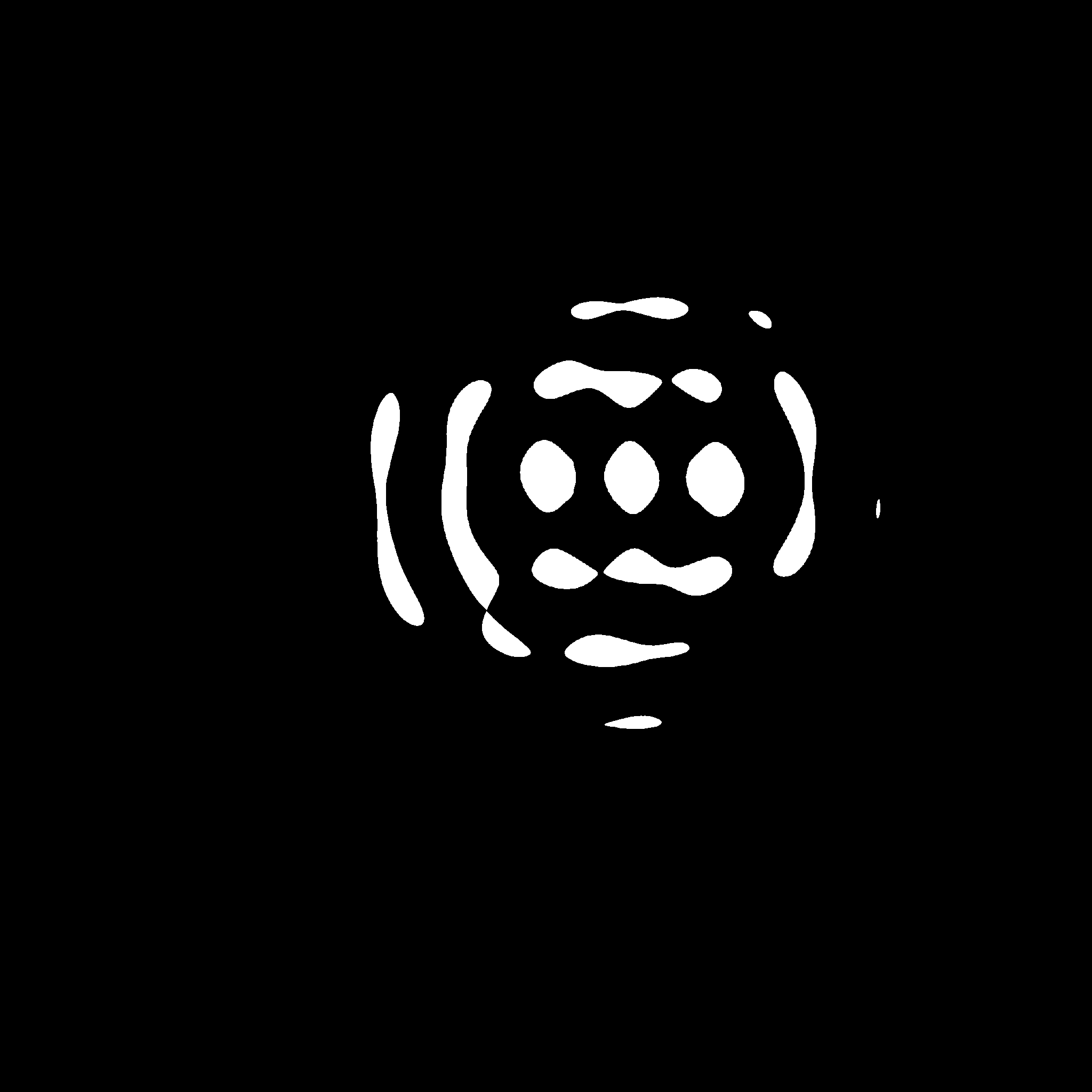}}
	\hspace{.3cm}
	\subfigure[ILILT-Wafer]{\includegraphics[width=.13\textwidth,trim={16cm 12cm 4cm 8cm},clip]{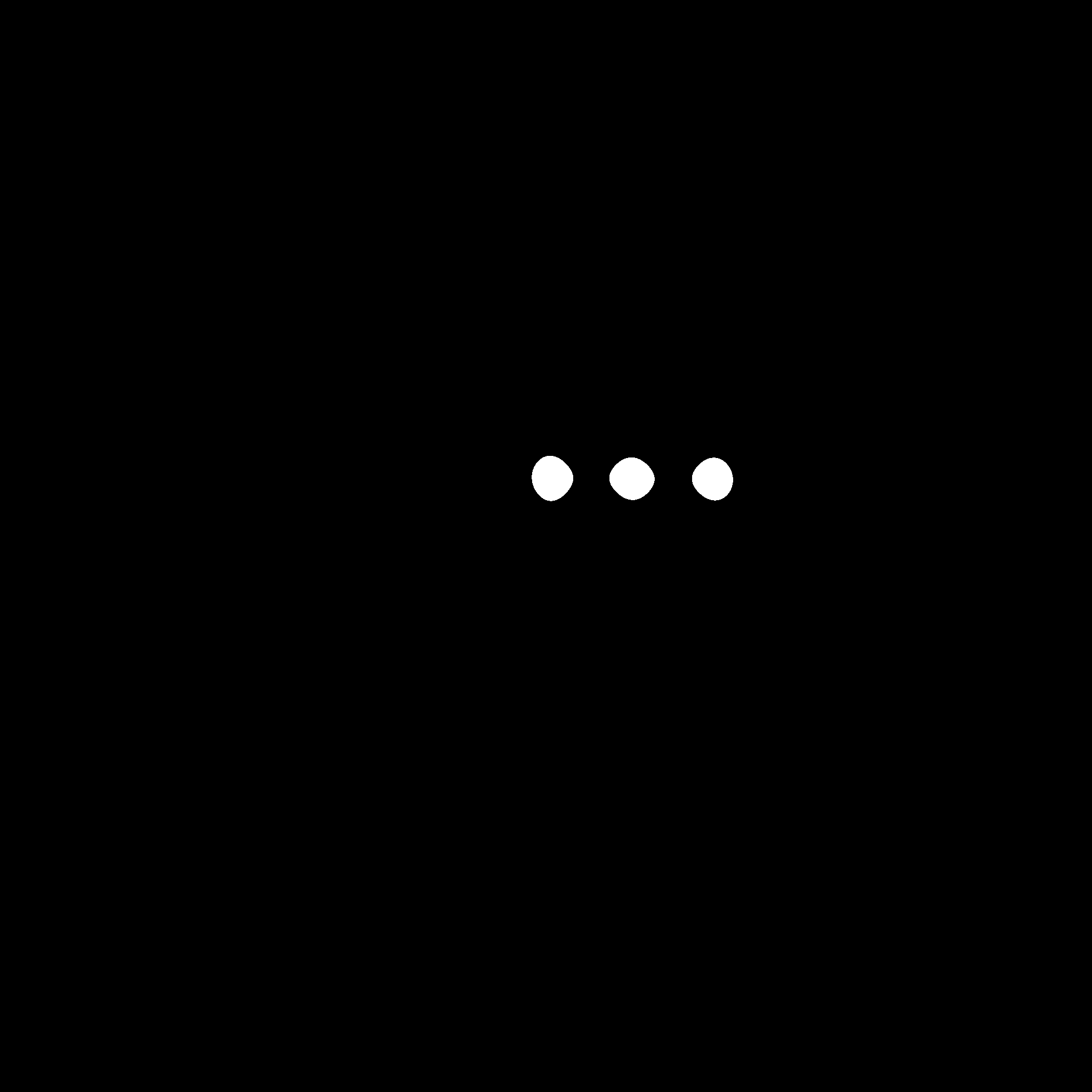}} \\
	\subfigure[Design]{\includegraphics[width=.13\textwidth,trim={10cm 15cm 10cm 5cm},clip]{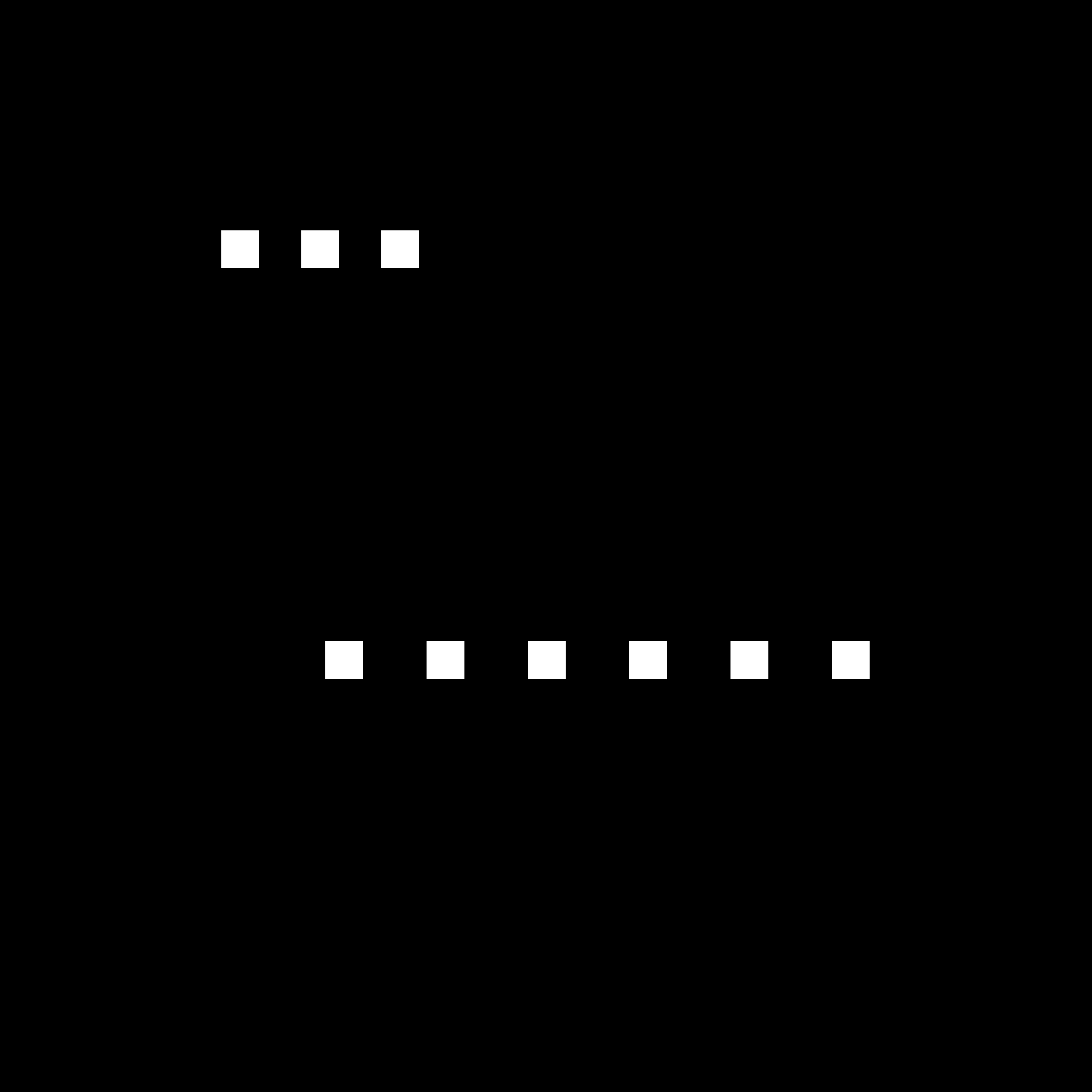}}
	\hspace{.3cm}
	\subfigure[ILT-Mask]{\includegraphics[width=.13\textwidth,trim={10cm 15cm 10cm 5cm},clip]{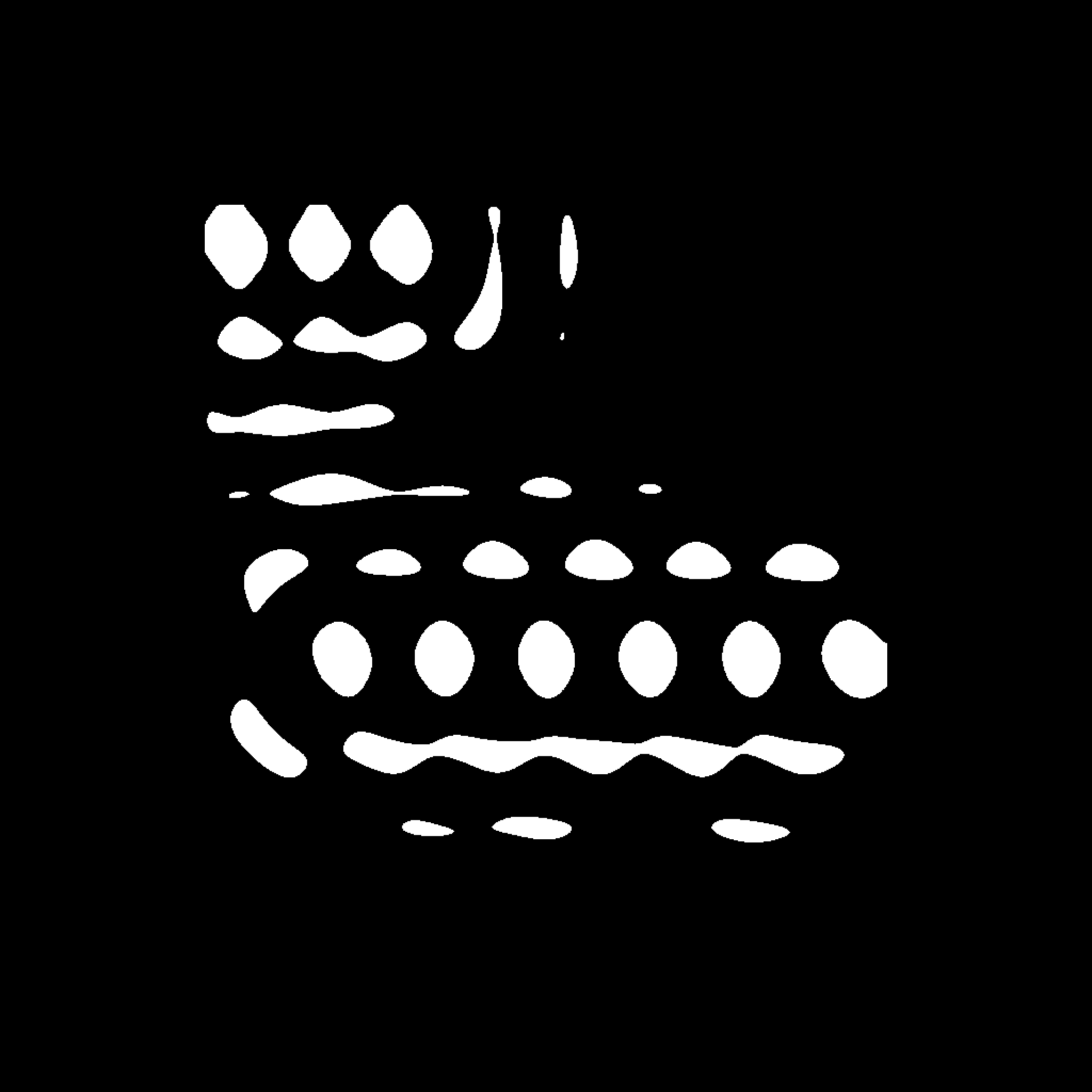} \label{fig:mprior-realb}}
	\hspace{.3cm}
	\subfigure[ILT-Wafer]{\includegraphics[width=.13\textwidth,trim={10cm 15cm 10cm 5cm},clip]{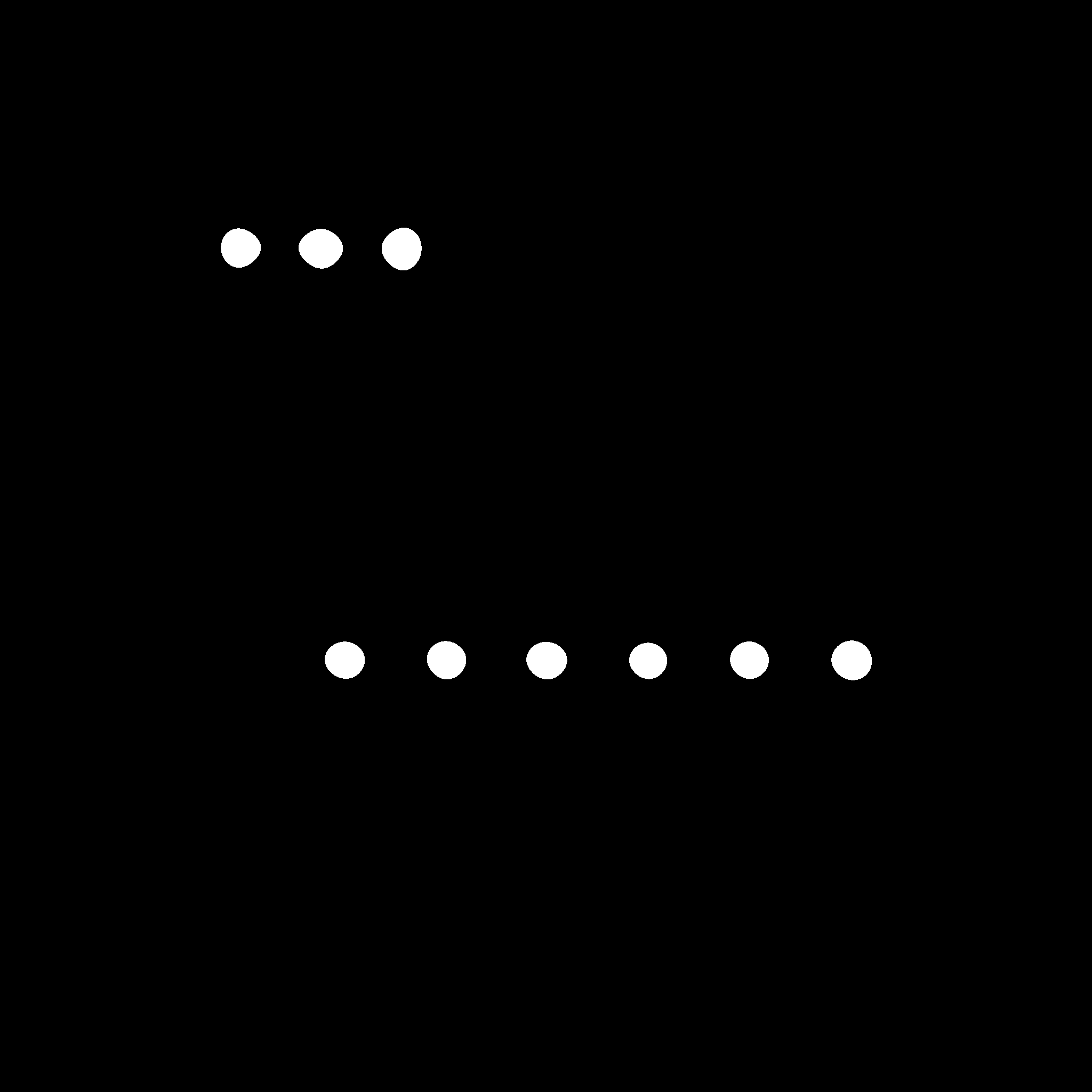}}
	\hspace{.3cm}
	\subfigure[ILILT-Mask]{\includegraphics[width=.13\textwidth,trim={10cm 15cm 10cm 5cm},clip]{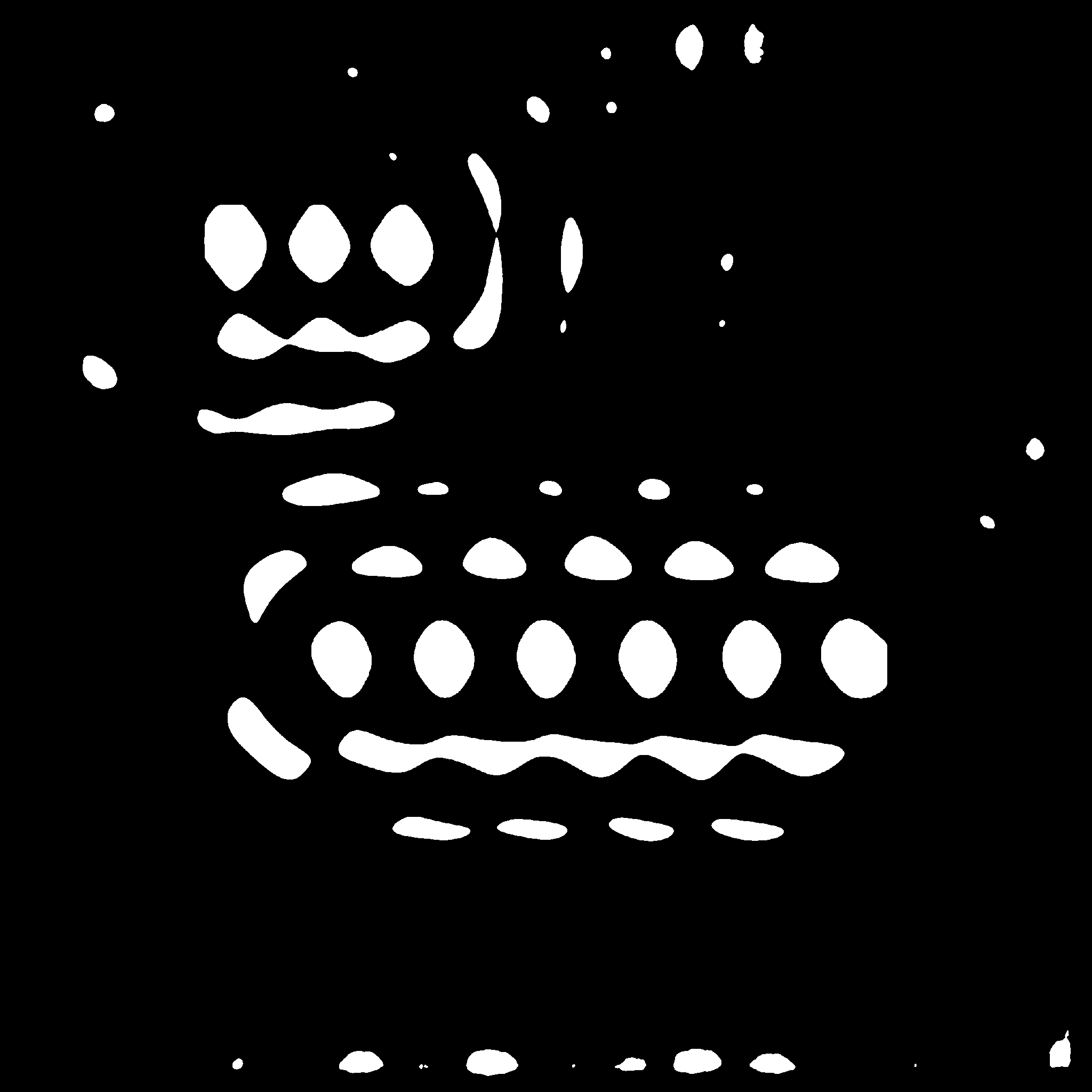}}
	\hspace{.3cm}
	\subfigure[ILILT-Wafer]{\includegraphics[width=.13\textwidth,trim={10cm 15cm 10cm 5cm},clip]{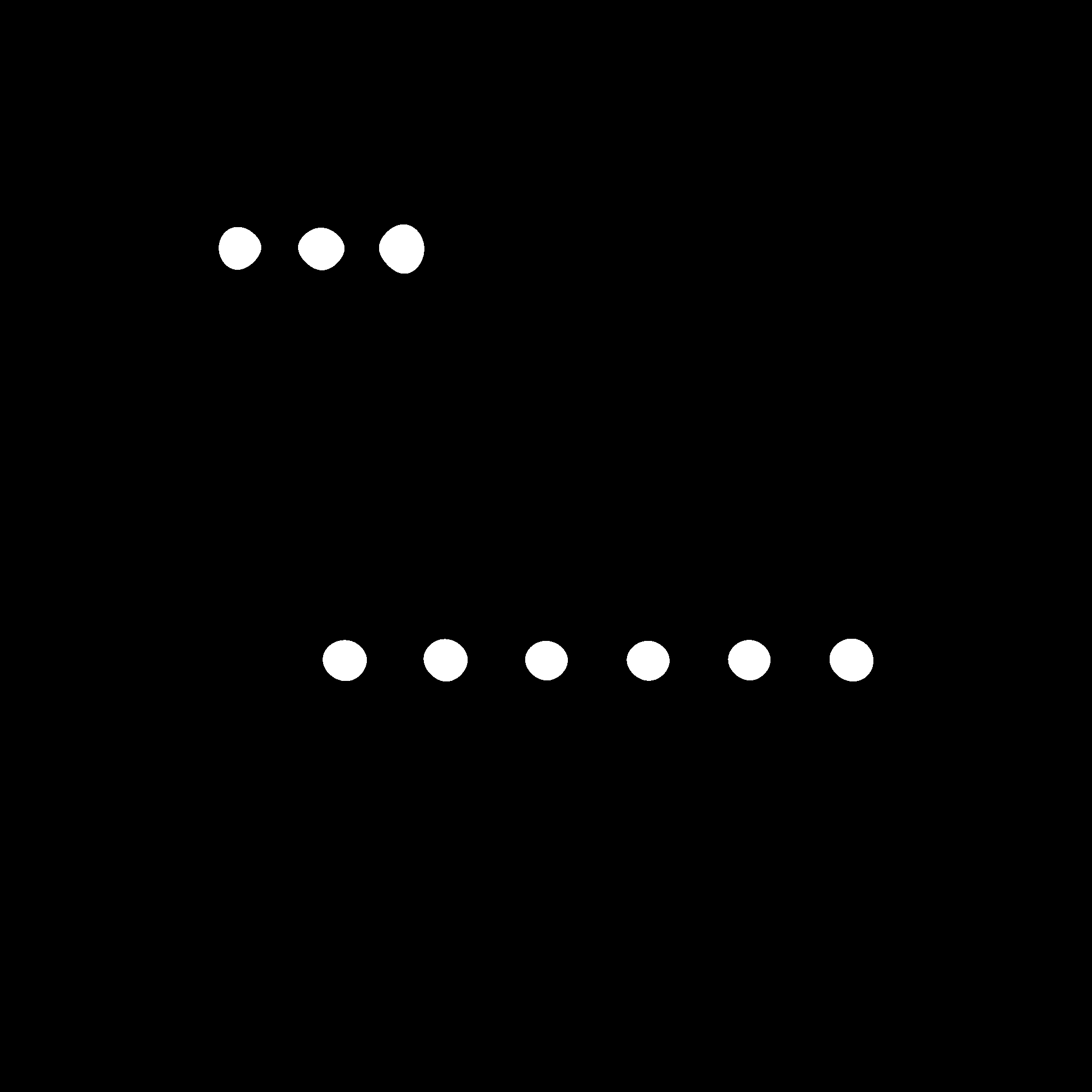}} 
	\caption{Mask prior learning from ILILT (P2PHD). Due to design pattern variations, the numerical solver sometimes fails to produce the correct solution. In this example, ILT generates rule-violating artifacts and performs feature shrinking excessively, causing the missing of design patterns as in (c). While ILILT can learn from the broader dataset and make correct predictions. \revise{For a more complicated case (f)-(j), ILILT can successfully produce optimized mask that can pass lithography printability check while numerical ILT still exhibits EPE violations in the printed wafer image.}}
	\label{fig:mprior}
\end{figure*}

In our experiments, we also observe an interesting behavior of ILILT. 
For some cases, ILILT can even produce better QoR than numerical solution.
An example is shown in \Cref{fig:mprior}, where the ILILT produces mask solution with zero EPE violations while numerical solver comes with four. 
This is because ILILT not only learns the mask optimization trajectory from the golden numerical solutions but also captures the mask solution from the broader dataset,
which makes the trained network equip deep mask priors. 
This can also be observed in the generated mask images that ILILT avoids generating rule-violation isolated artifacts while numerical ILT does (see small white dots at the bottom of \Cref{fig:mprior-realb}).
The result can also be explained through deep image prior \cite{deep-image-prior}, that during training neural networks capture the structural knowledge with priority. ILILT strengthens this prior through weigh-tied unrolling and lithography guidance.
\section{Conclusion}
\label{sec:conclu}

In this paper, we propose a novel solution to the mask optimization problem with inverse lithography technology.
We argue that existing AI-based approaches are overly reliant on conventional numerical solvers and lack forward lithography conditions, 
limiting the applicability of AI-based approaches to mask optimization scenarios.
To further exploit the potential of AI and machine learning, we carefully analyze the characteristics of the ILT procedure and reformulate it as an implicit layer learning problem. 
Instead of applying the end-to-end DEQ method, we solve the ILT layer through ILILT, a weight-tied model, where a forward lithography estimator provides lithography information in the optimization procedure.
Experiments demonstrate the effectiveness of ILILT learning an ILT layer with limited intervention of a traditional numerical solver.
To the best of our knowledge, this is the first time an AI solution has been proposed to imitate the working scheme of a traditional ILT system.
We hope this work can further motivate research into expanding the potential of AI for solving complex design automation problems. 
Additional investigations on production-level designs are necessary to prototype the framework to study whether QoR meets manufacturing standards. 

\section*{Broader Impact}
Lithography is the very first topic in chip design community brought into applied deep learning. 
Because design layouts, masks and wafer patterns are naturally in the format of images which can benefit from many backbones and algorithms, though domain adaptions are necessary. 
Early efforts only fall into forward predictive tasks like hotspot detection or lithography simulation. 
There is limited research dealing with chip optimization problems that can directly benefit semiconductor manufacturing.
This work, ILILT, we believe is the first effort to solve inverse lithography/mask optimization directly using machine learning without the intervention of traditional numerical solvers. 
Hopefully, this can motivate new research strategies on how is AI included in chip design and manufacturing flow,
and the advancement of chip design can feedback and benefit AI in the long run.

{
	\bibliographystyle{icml2024}
	\bibliography{ref/Top,ref/DFM,ref/Additional,ref/HSD}
}

\newpage
\appendix
\onecolumn
\section{Lithography Simulation Model}
Throughout the manuscript, we use $f_l$ to represent the entire lithography process.
In detail, this can be broken into two stages: (1) optical modeling and (2) resist modeling.
The optical modeling tries to find the illumination intensity after the mask and is projected on the silicon wafer.
This process can be mathematically written as,
\begin{align}
    \label{eq:optical}
    \vec{I} = \sum_{k=1}^{N} \alpha_k ||\vec{h}_k \otimes \vec{M}||^2,
\end{align}
where $\vec{I}$ is the light intensity projected on the silicon wafer, $\vec{M}$ denotes the mask image, and $\vec{h}_k$'s and $\alpha_k$'s
are lithography system-related parameters that are calibrated in foundries.
$\vec{h}_k$'s and $\alpha_k$'s are different under different process conditions.

During resist modeling, $\vec{I}$ will have a chemical reaction with some "resist materials" and etch chip design on the top of the silicon wafer, which is given by
  \begin{equation}
  \label{eq:resist}
    \vec{Z}(i,j)=
    \begin{cases}
      1, & \text{if}~~\vec{I}(i,j)>I_\text{th}, \\
      0, & \text{otherwise}.
    \end{cases}
  \end{equation}
In this paper, $I_\text{th}=0.225$, which is determined by the technology profile.

\begin{figure}[h]
	\centering 
	\subfigure[Mask]{\includegraphics[width=.13\textwidth,trim={10cm 10cm 10cm 10cm},clip]{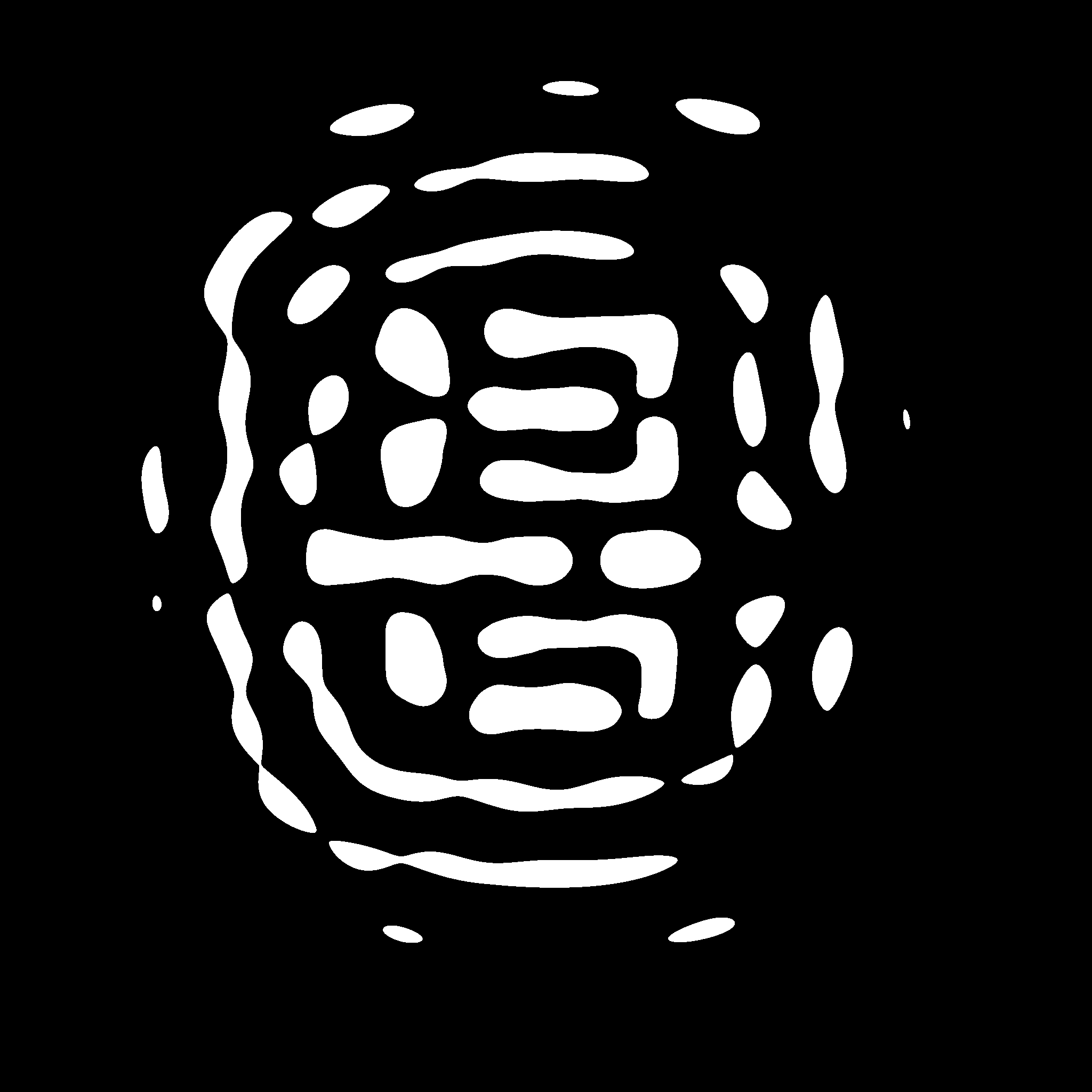}}
	\hspace{.3cm}
	\subfigure[Intensity]{\includegraphics[width=.13\textwidth,trim={10cm 10cm 10cm 10cm},clip]{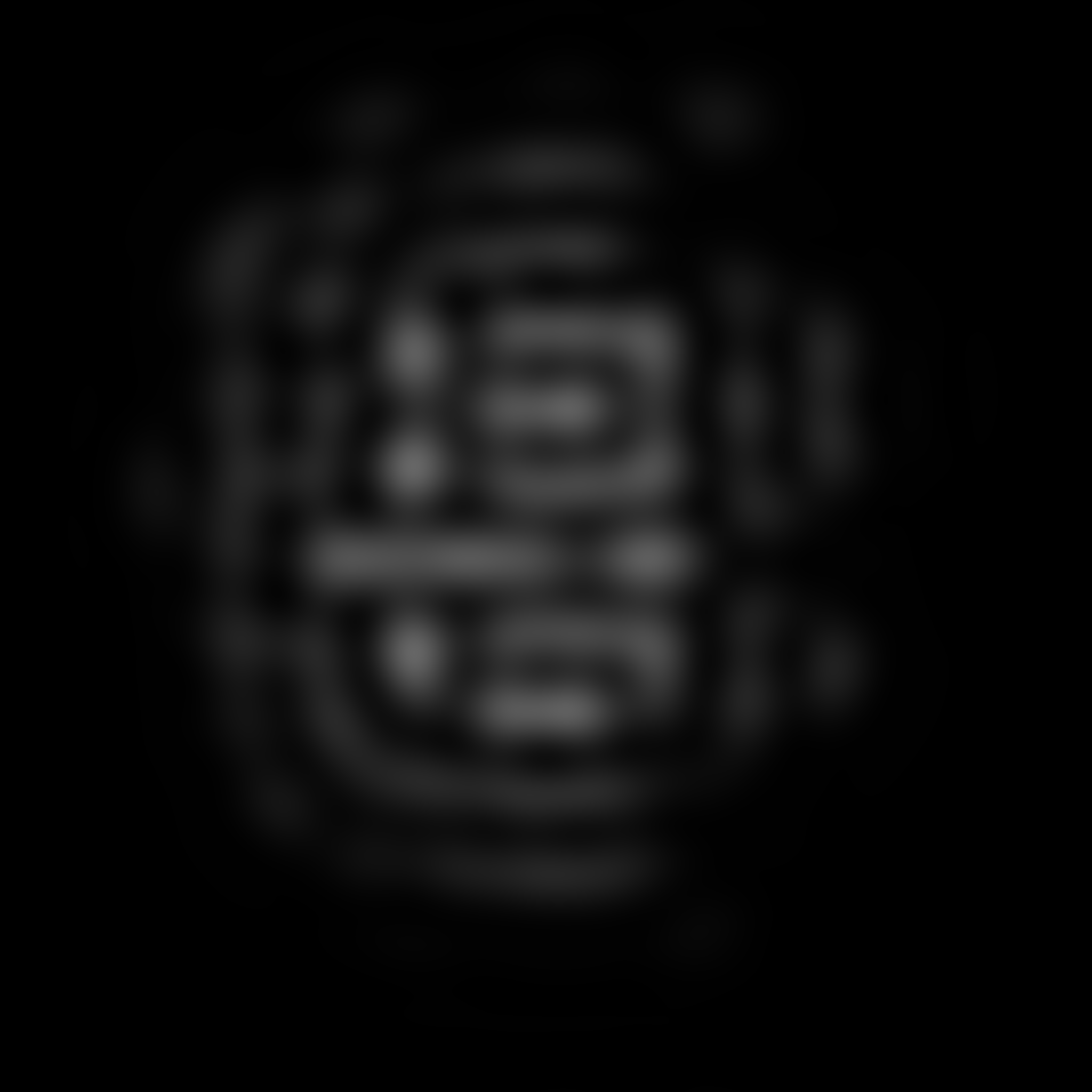}}
	\hspace{.3cm}
	\subfigure[Wafer]{\includegraphics[width=.13\textwidth,trim={10cm 10cm 10cm 10cm},clip]{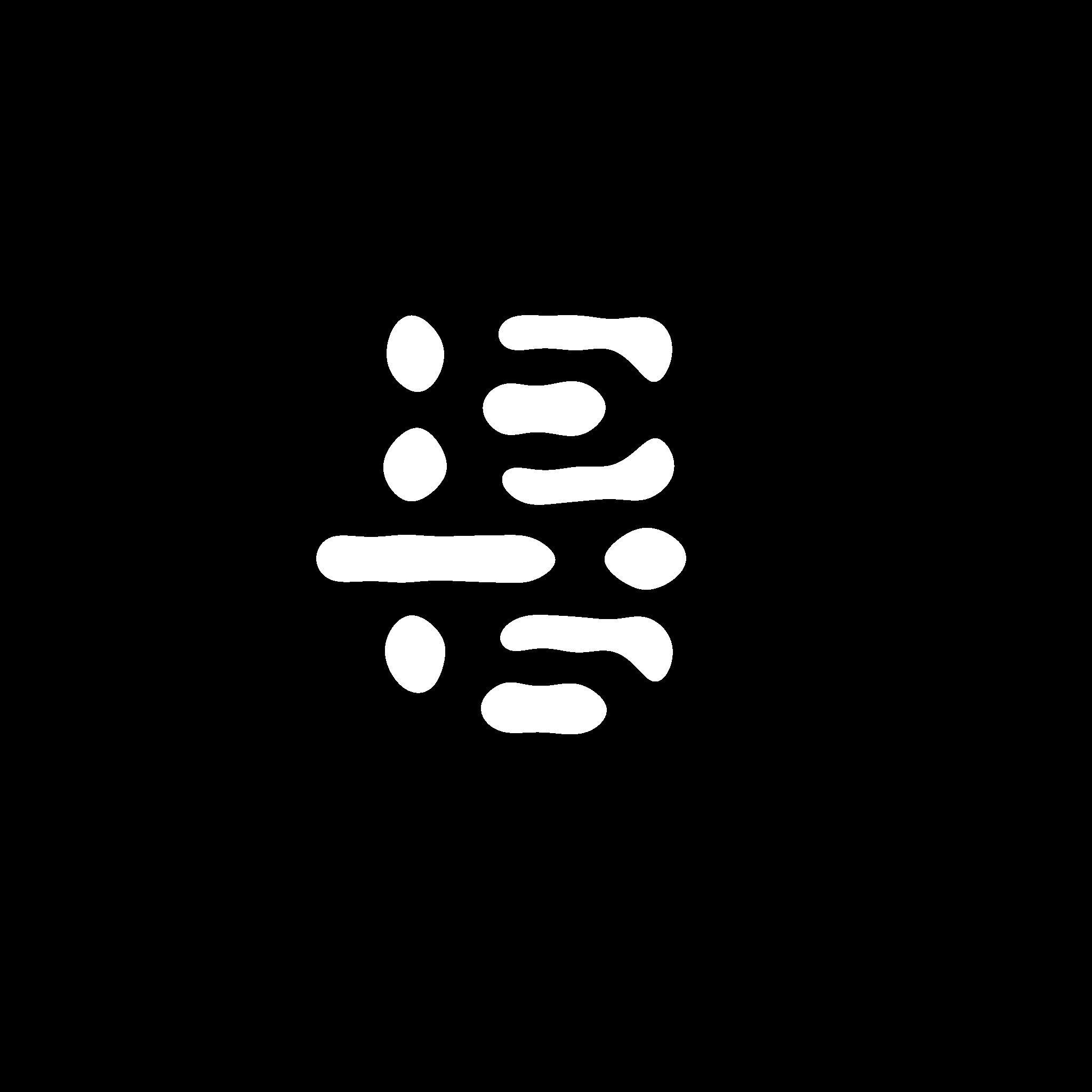}}
	\caption{Lithography process. (a) Mask image contains the patterns of optimized design; (b) Intensity image is the UV lights that are projected on the silicon wafer given by \Cref{eq:optical}; (c) Wafer image is the patterns on silicon given \Cref{eq:resist}.}
	\label{fig:dataset}
\end{figure}

\section{Numerical ILT}
Because numerical solver usually requires differentiable objectives. 
In practical, we relax $\vec{M} \in \{0,1\}$ and $\vec{Z} \in \{0,1\}$ through sigmoid. 
\begin{align}
\label{eq:mask_relax}
    \vec{M} = \dfrac{1}{1+ \exp(-\beta_m (\vec{M}^\prime-0.5))},
\end{align}
where $\vec{M}^\prime$ is some auxilary variable that is usually initialized as the original chip design $\vec{Z}^\ast$.
\begin{align}
\label{eq:resist_relax}
    \vec{Z} = \dfrac{1}{1+ \exp(-\beta_z (\vec{I}-I_\text{th})}.
\end{align}
And now we can solve the following ILT problem through gradient descent.
\begin{align}
	\min_{\vec{M}, \vec{M}^\prime}~~& l=||\vec{Z}-\vec{Z}^\ast|| 	\label{eq:trainilt}\\
	\text{s.t.~~}& \text{\Cref{eq:optical,eq:mask_relax,eq:resist_relax}}. \nonumber 
\end{align}
It should be noted that numerical ILT requires frequent calls of time-consuming lithography simulation.
\section{CFNO}
\begin{figure}
    \centering
    \includegraphics[width=0.5\textwidth]{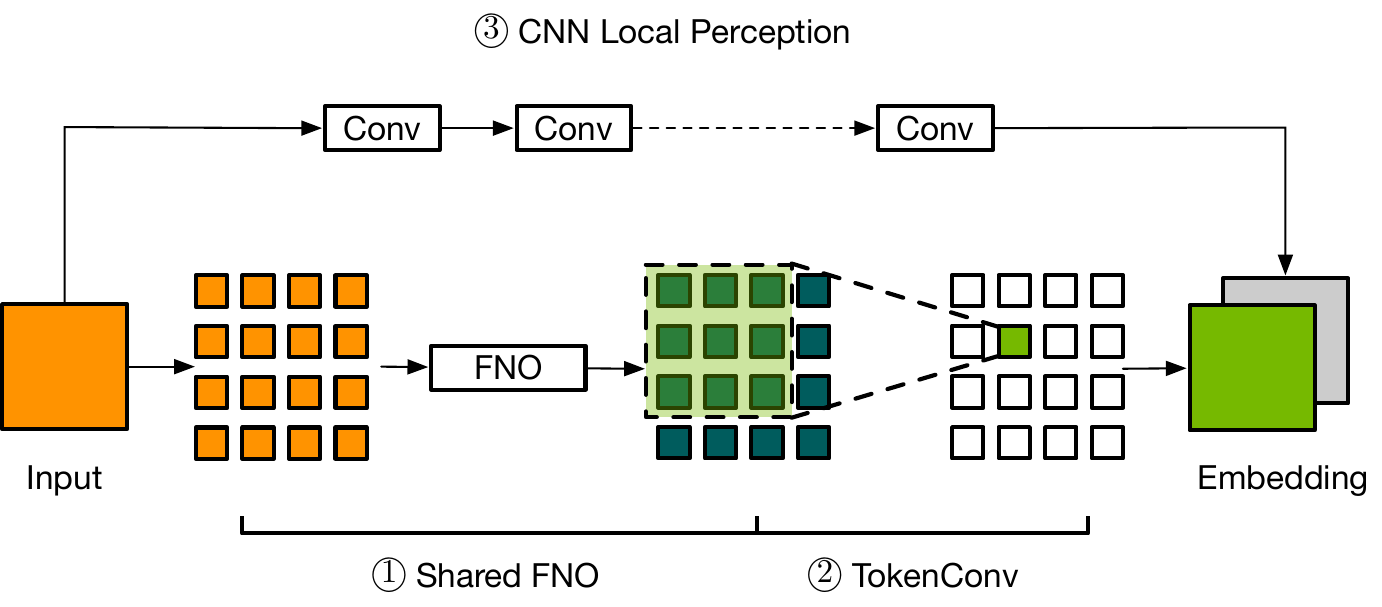}
    \caption{CFNO backbone design.}
    \label{fig:cfno}
\end{figure}

\begin{figure}
    \centering
    \includegraphics[width=0.3\textwidth]{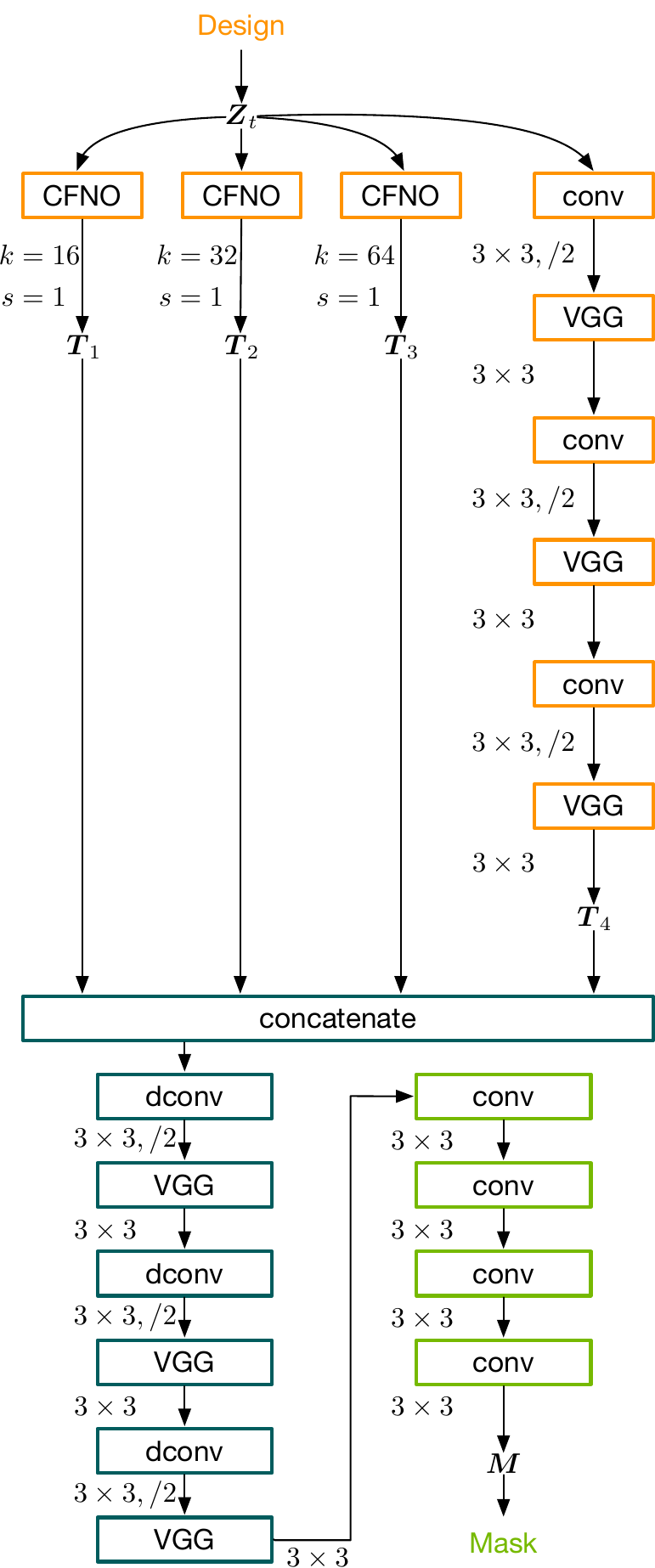}
    \caption{The data pipeline of $g$ used throughout experiments.}
    \label{fig:cfno-mo}
\end{figure}

CFNO, the convolutional Fourier Neural Operator \cite{CFNO}, was built upon the original FNO \cite{fno} design. 
The capability of FNO learning frequency domain features fits the lithography well. 
However, the original FNO design has a few challenges: (1) FNO requires computations of FFT of the entire input mask image which is computationally costly; (2) FNO, once trained, can only be applied on the same-sized image. 
CFNO addresses these problems by combining the patch embedding in ViT. 
The core idea is to separate the input into non-overlapped patches, and each patch will go through a shared FNO layer as in \Cref{fig:cfno}.
Then we perform a token-wise convolution to learn the relations among different patches and finally form the mask embedding. 
The overall architecture of $g$ is shown in \Cref{fig:cfno-mo}.

\revise{
\section{Pix2PixHD}
Pix2PixHD \cite{pix2pixhd} is an improved version of Pix2Pix which employs hierarchical local enhancers for high-resolution image generation.
To make the framework fit more on the mask optimization problem we choose $ngf=16, n\_downsample\_global=4, n\_blocks\_global=9$ as defined in \url{https://github.com/NVIDIA/pix2pixHD}.
We also make certain improvements by replacing deconvolution layers with bicubic interpolation and stride-1 convolution layers for smooth shape generation.
We also apply 4$\times$4 downsampling (achieved by average pooling) at the input directly to reduce computing overhead and perform interpolation to cast the image back to desired resolution.
This modification yields 45M model as listed in \Cref{sec:result}.}
\revise{
\section{Additional Visualization}
\begin{figure}[h]
	\subfigure[AES-Design]{\includegraphics[width=.2\textwidth,trim={5cm 5cm 7cm 7cm},clip]{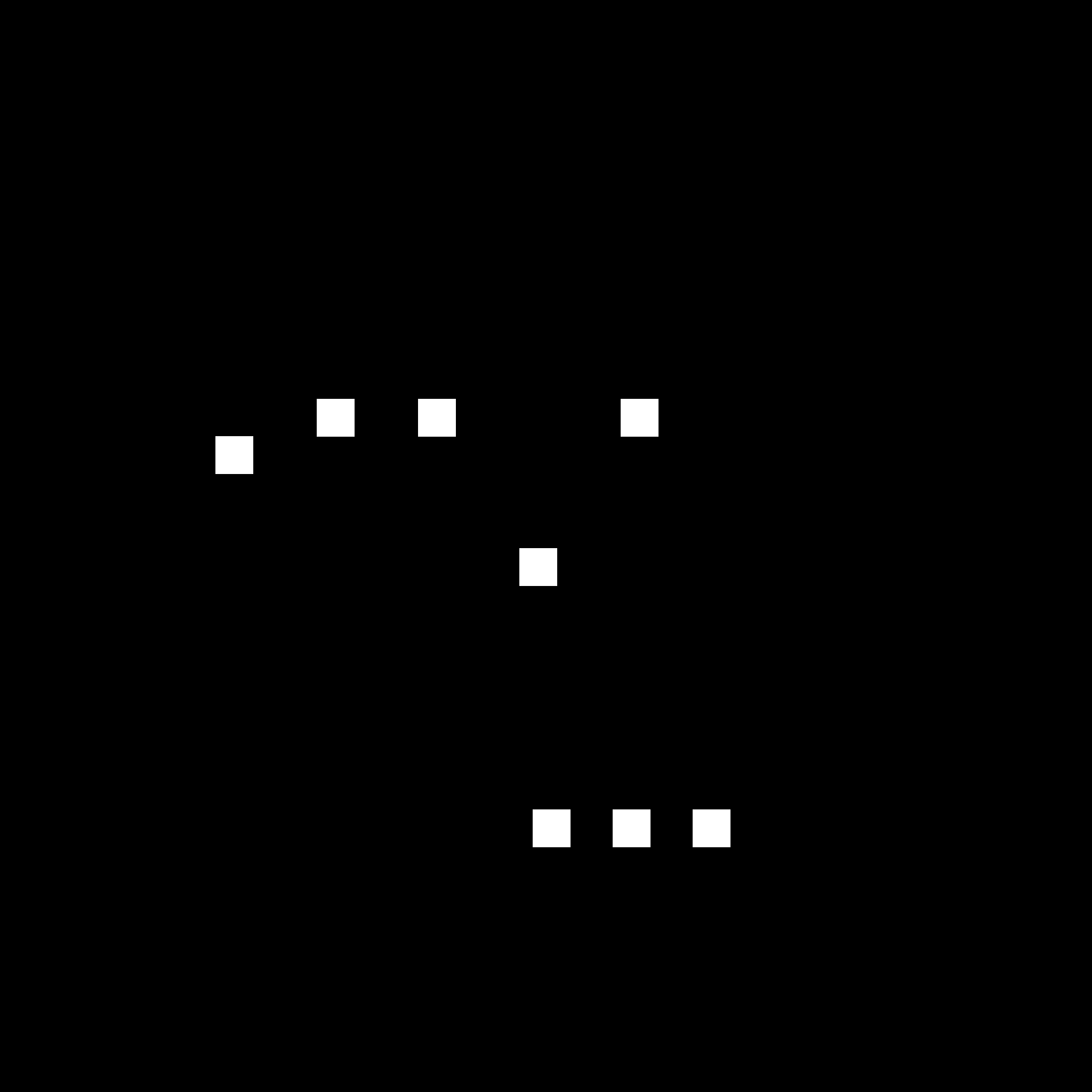}} \\
	\subfigure[ILT-Mask]{\includegraphics[width=.2\textwidth,trim={5cm 5cm 7cm 7cm},clip]{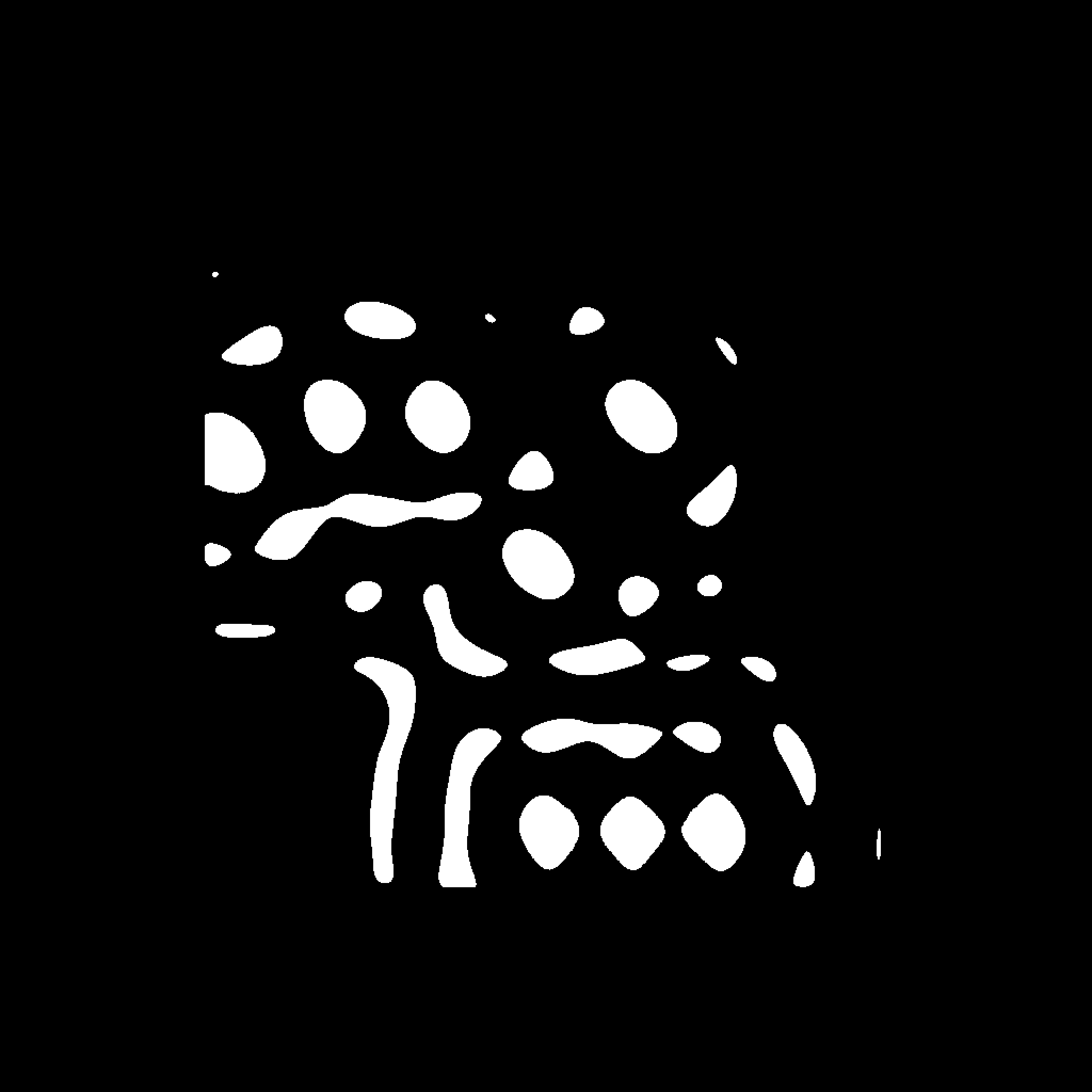}}
	\hspace{-.15cm}
	\subfigure[ILILT(P2P)-Mask]{\includegraphics[width=.2\textwidth,trim={5cm 5cm 7cm 7cm},clip]{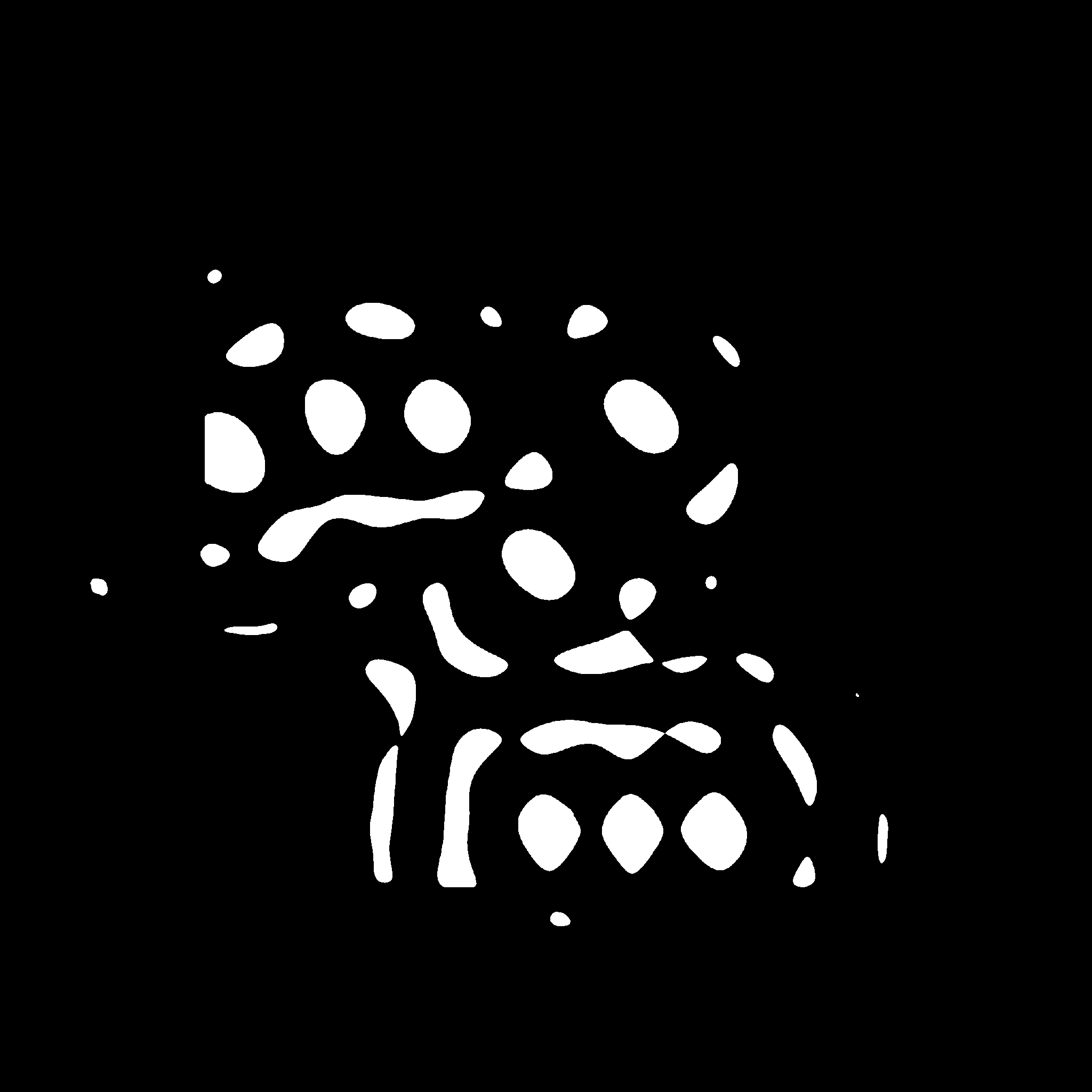}}
	\hspace{-.15cm}
	\subfigure[ILILT(CFNO)-Mask]{\includegraphics[width=.2\textwidth,trim={5cm 5cm 7cm 7cm},clip]{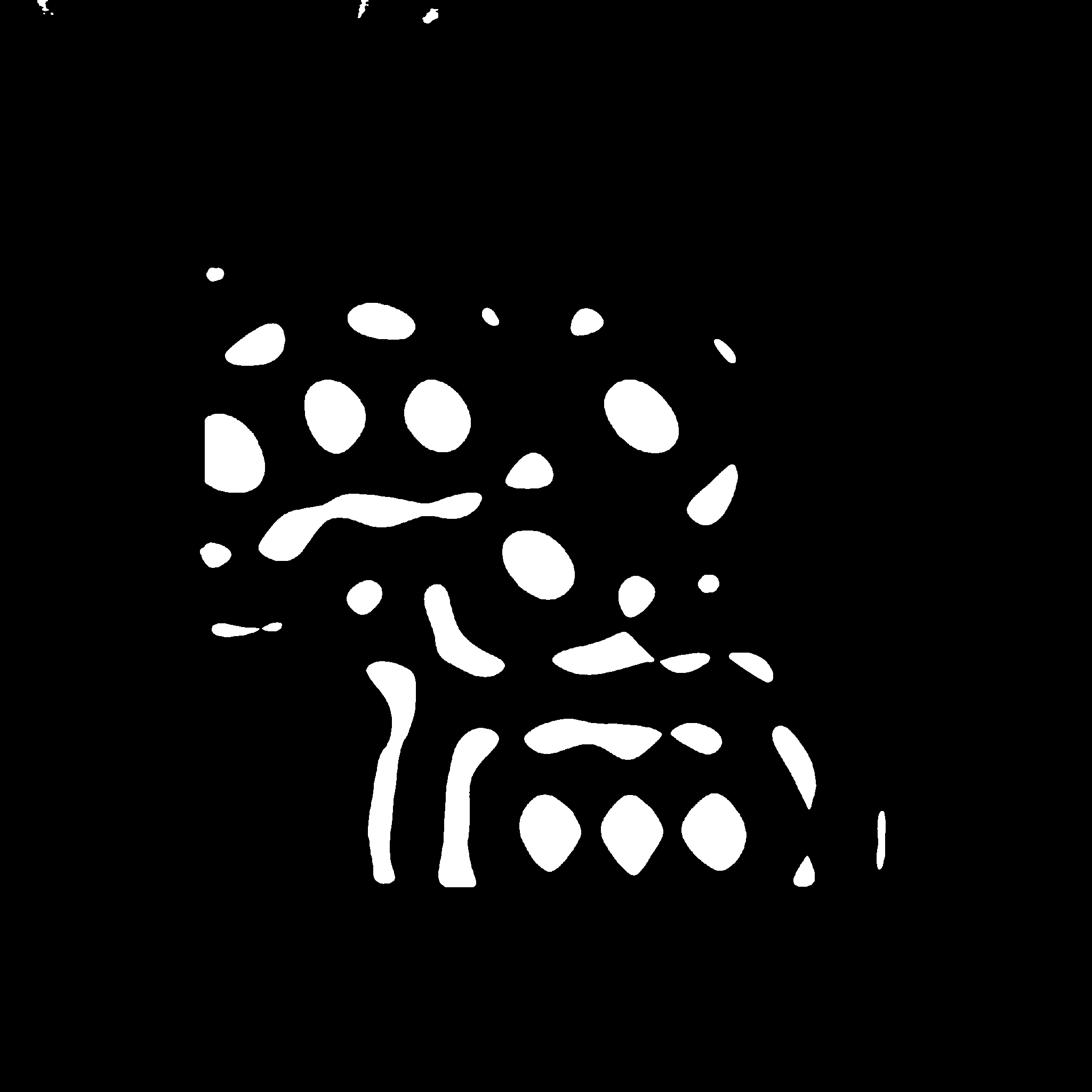}}
	\hspace{-.15cm}
	\subfigure[L2O(P2P)-Mask]{\includegraphics[width=.2\textwidth,trim={5cm 5cm 7cm 7cm},clip]{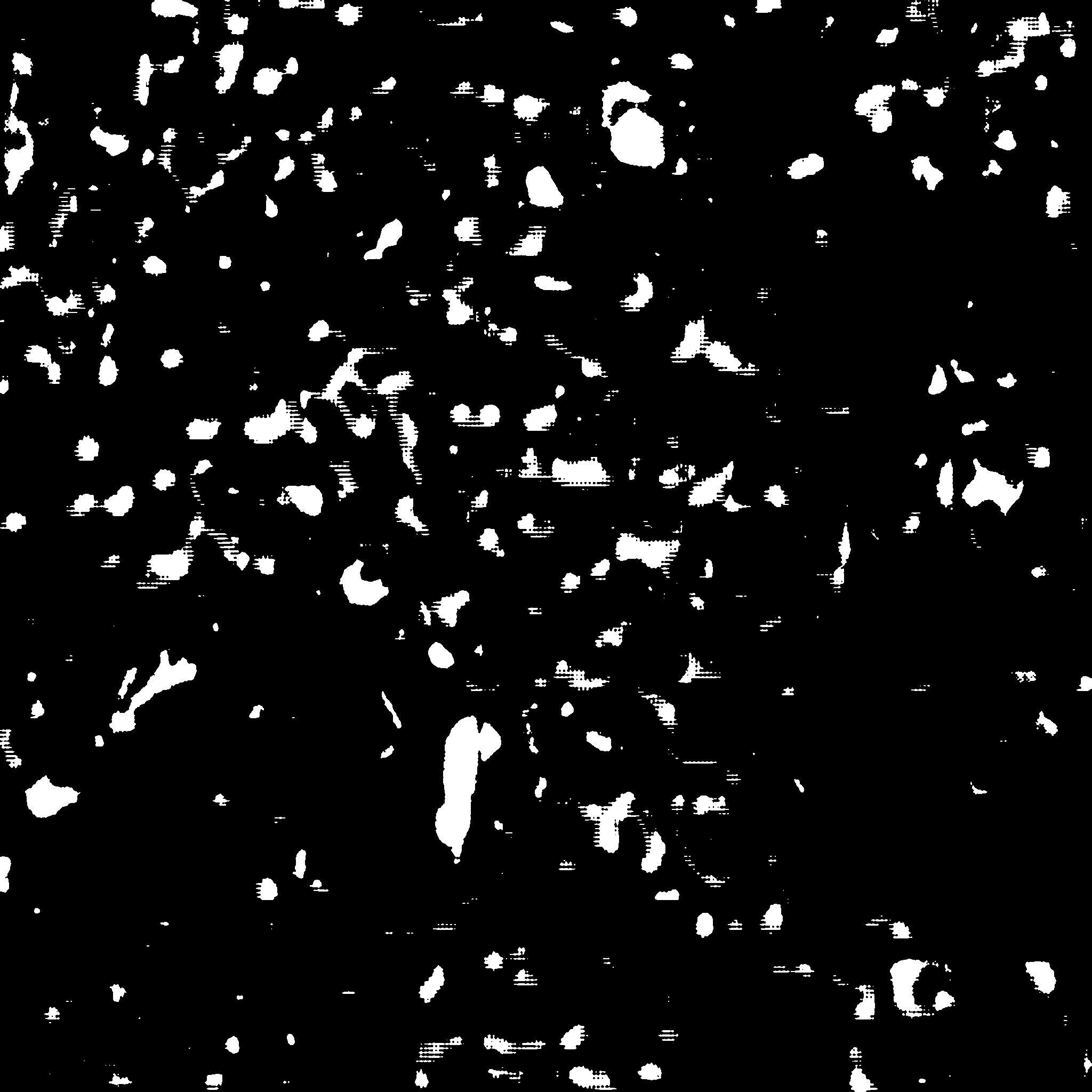}}
	\hspace{-.15cm}
	\subfigure[L2O(CFNO)-Mask]{\includegraphics[width=.2\textwidth,trim={5cm 5cm 7cm 7cm},clip]{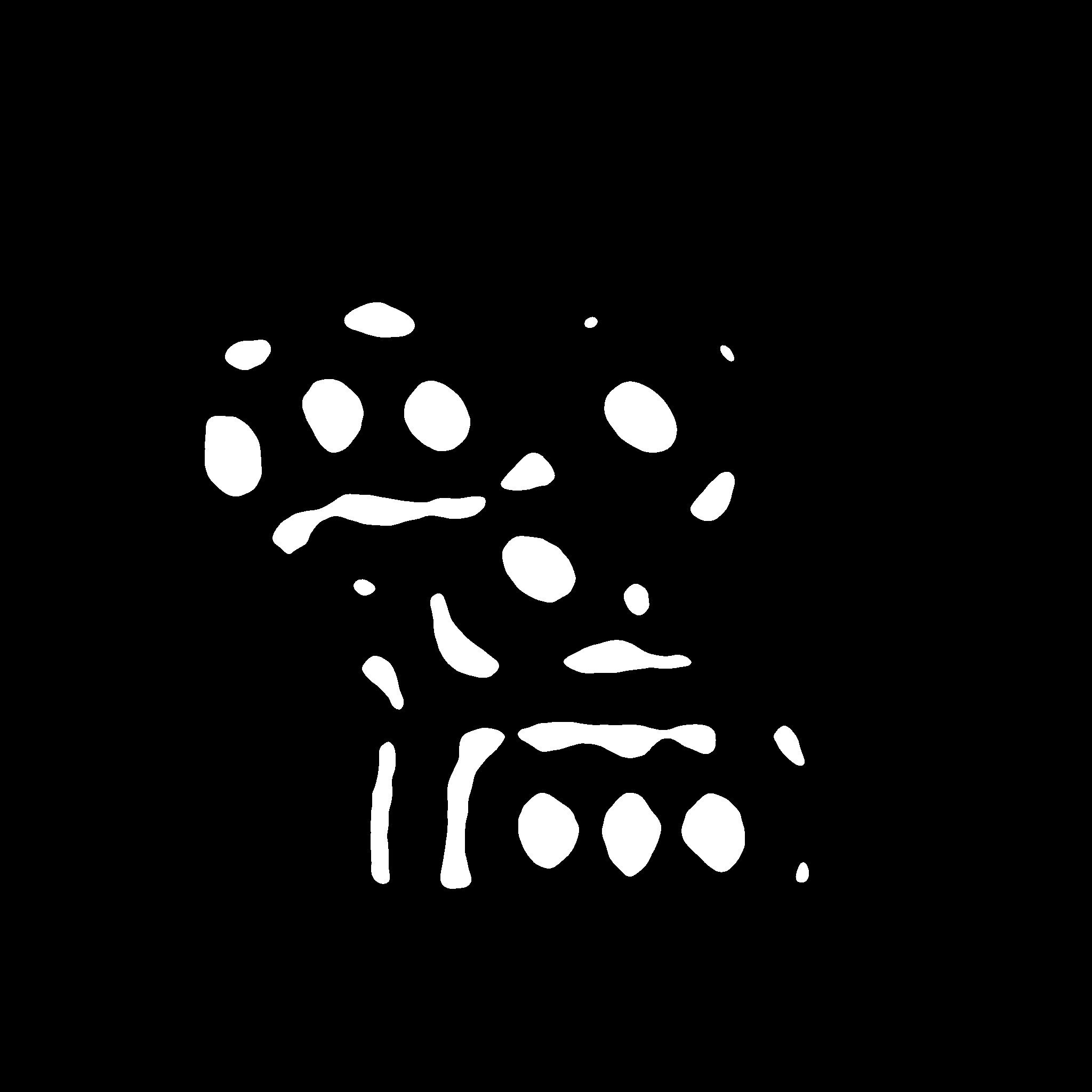}}\\
	\subfigure[ILT-Wafer]{\includegraphics[width=.2\textwidth,trim={5cm 5cm 7cm 7cm},clip]{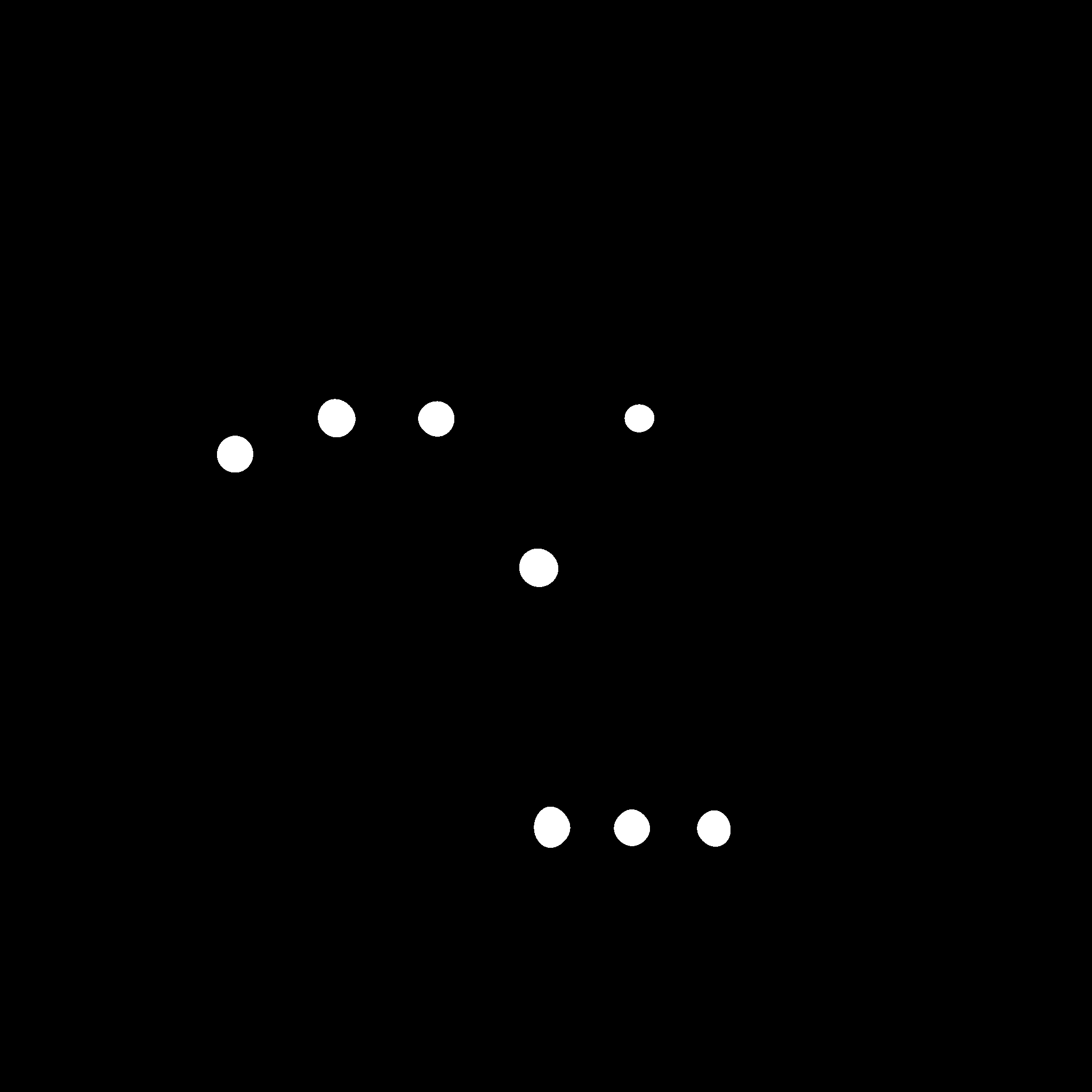}}
	\hspace{-.15cm}
	\subfigure[ILILT(P2P)-Wafer]{\includegraphics[width=.2\textwidth,trim={5cm 5cm 7cm 7cm},clip]{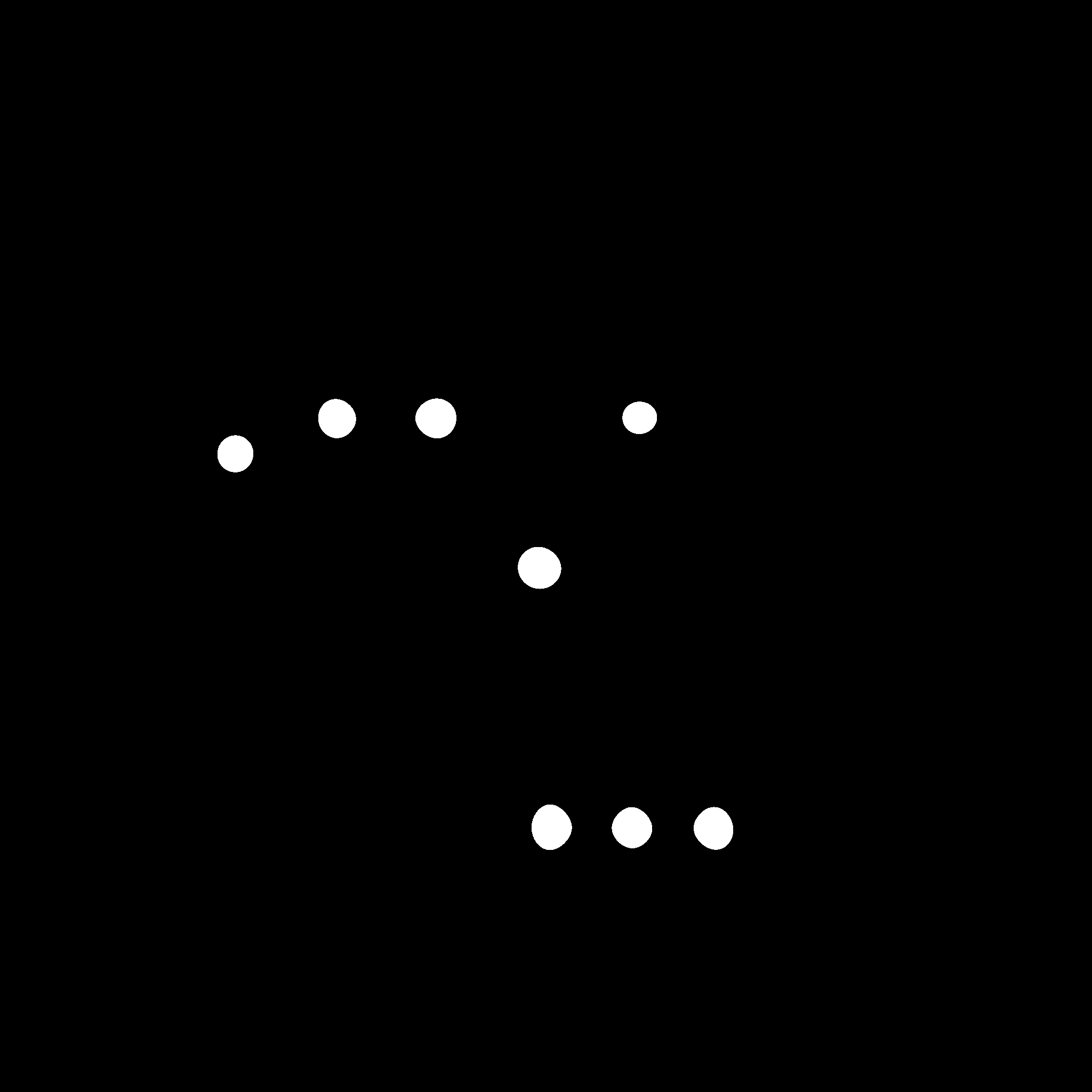}} 
	\hspace{-.15cm}
	\subfigure[ILILT(CFNO)-Wafer]{\includegraphics[width=.2\textwidth,trim={5cm 5cm 7cm 7cm},clip]{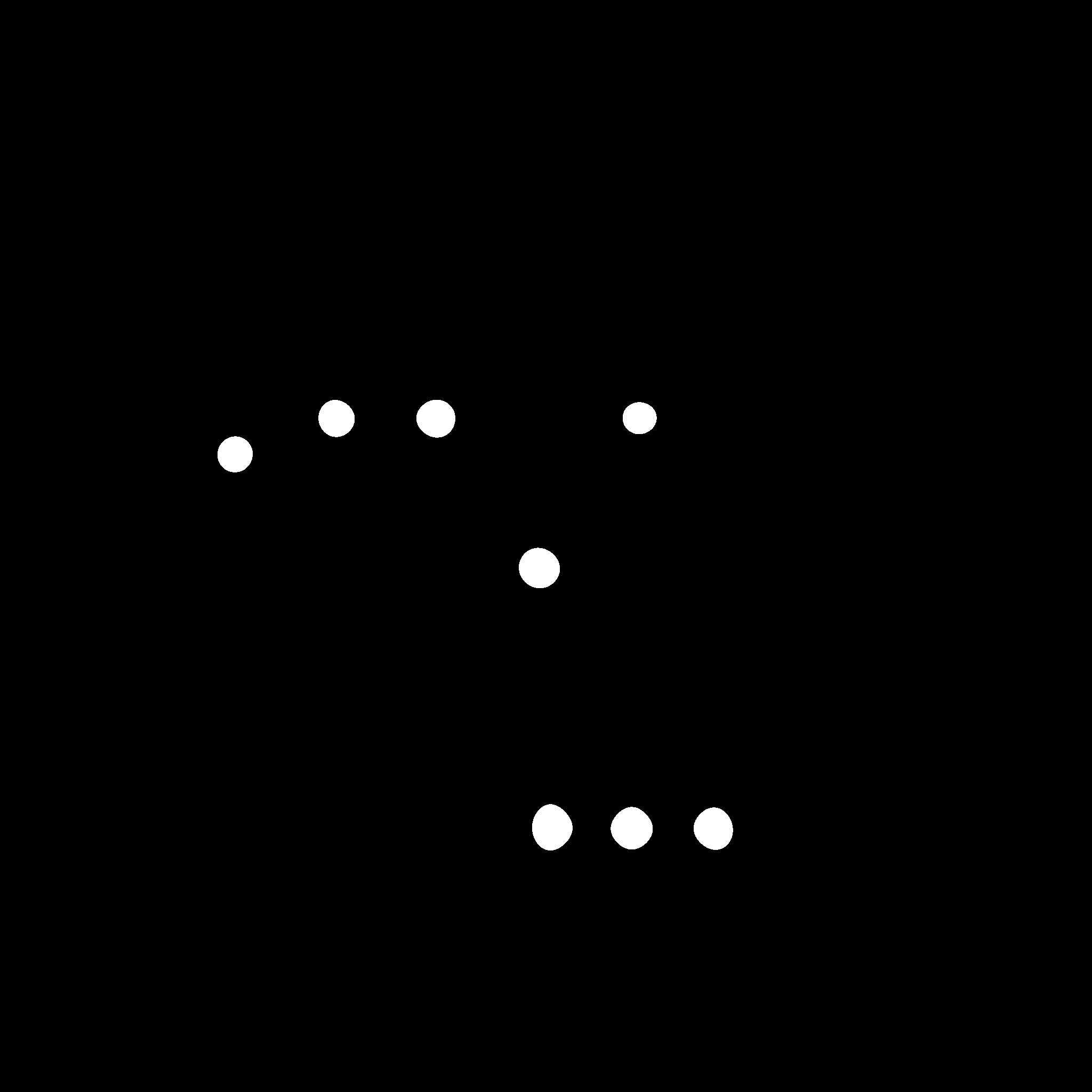}} 
	\hspace{-.15cm}
	\subfigure[L2O(P2P)-Wafer]{\includegraphics[width=.2\textwidth,trim={5cm 5cm 7cm 7cm},clip]{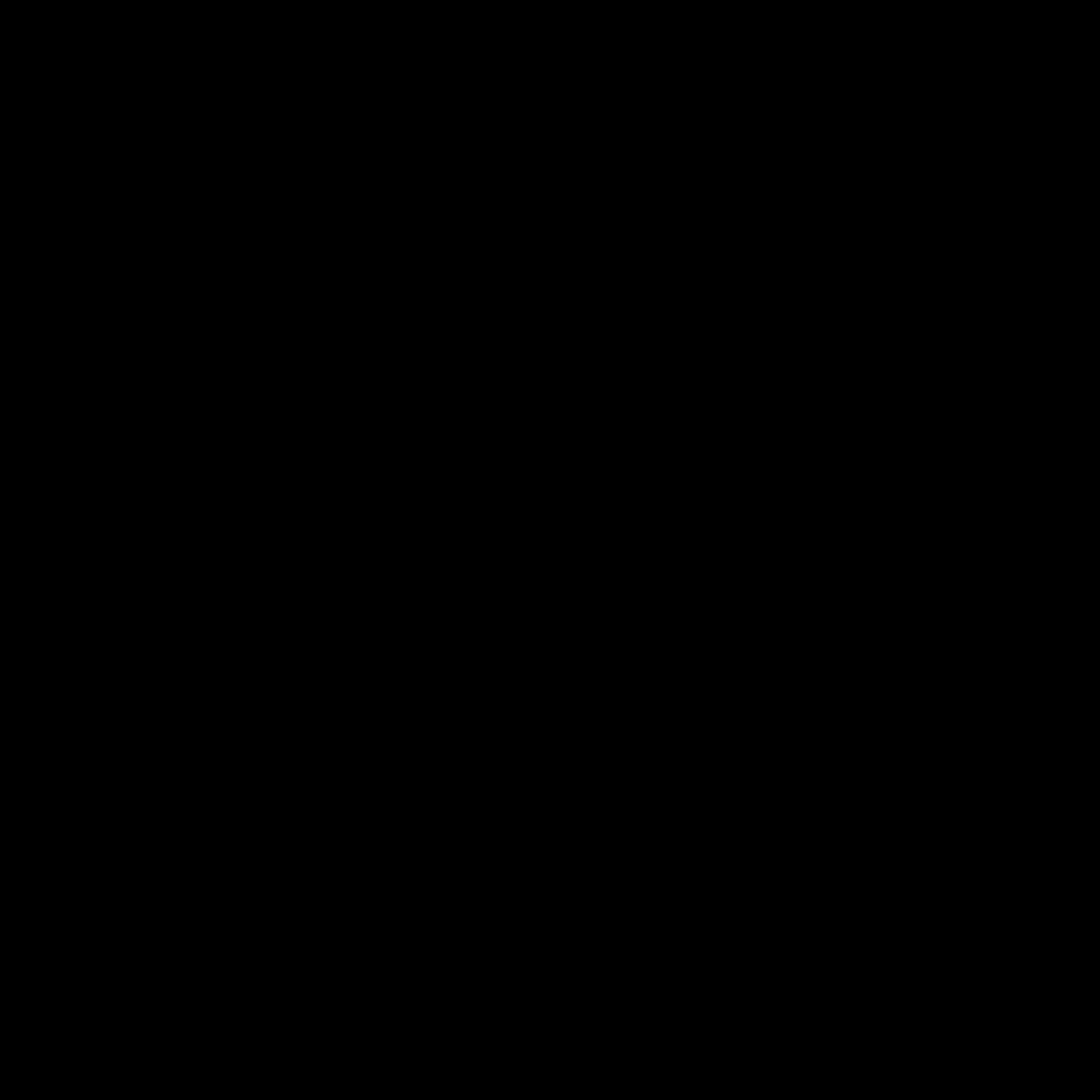}} 
	\hspace{-.15cm}
	\subfigure[L2O(CFNO)-Wafer]{\includegraphics[width=.2\textwidth,trim={5cm 5cm 7cm 7cm},clip]{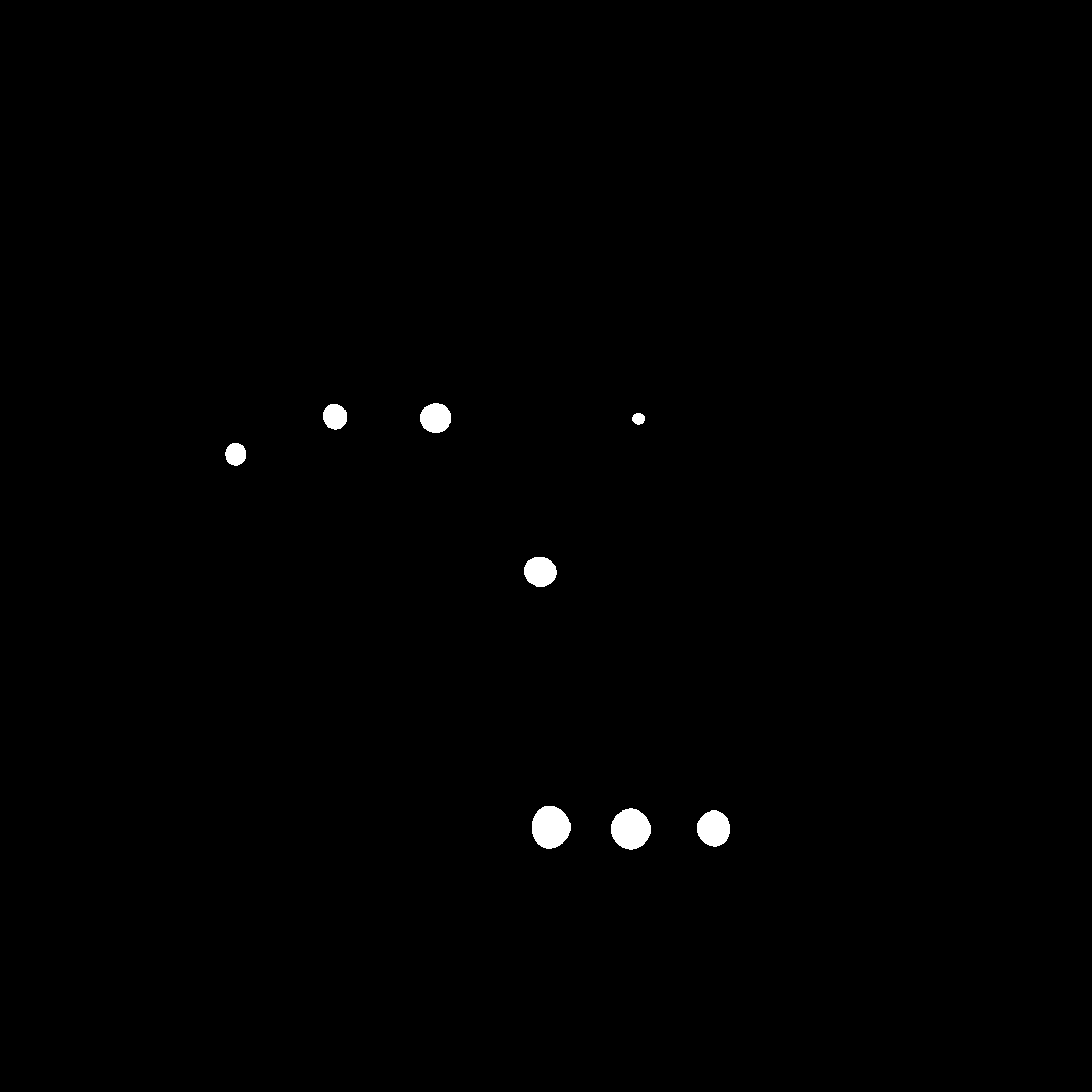}} 
	\caption{Visualization sample of AES chip design for ILILT and L2O with CNN and Neural Operator backbones. L2O with Pix2PixHD backbone even produces meaningless artifacts.}
	\label{fig:additional}
\end{figure}
}

\end{document}